\documentclass[10pt,twocolumn,letterpaper]{article}
\usepackage{times}
\usepackage{epsfig}
\usepackage{graphicx}
\usepackage{amsmath}
\usepackage{amssymb}

\usepackage[tight,scriptsize]{subfigure}
\usepackage{multirow}
\usepackage[usenames,dvipsnames,svgnames,table]{xcolor}
\usepackage[percent]{overpic}
\usepackage{transparent}
\pdfpageattr {/Group << /S /Transparency /I true /CS /DeviceRGB>>}

\usepackage{cvpr}

\newcommand{\be}{\begin{equation}}
\newcommand{\ee}{\end{equation}}
\newcommand{\ba}{\left[ \begin{array}}
\newcommand{\ea}{\end{array} \right]}
\newcommand{\bea}{\begin{eqnarray}}
\newcommand{\eea}{\end{eqnarray}}

\def\real{\mathbb{R}}

\def\cut#1{{}}

\def\cutForReview#1{{}}

\def\eg{{\em e.g.,}}
\def\ie{{\em i.e., }}

\def\newcomment#1{{#1}}
\def\comment#1{{#1}}
\def\jcomment#1{{#1}}

\usepackage[pagebackref=true,breaklinks=true,letterpaper=true,colorlinks,bookmarks=false]{hyperref}

\cvprfinalcopy 

\def\cvprPaperID{858} 

\ifcvprfinal\fi
\begin{document}

\title{\bf Domain-Size Pooling in Local Descriptors: DSP-SIFT} 
\author{Jingming Dong ~~~~~~~~~~~~~~~ Stefano Soatto\\~\\
UCLA Vision Lab, \ University of California, Los Angeles, CA 90095\\
{\tt \small \{dong,soatto\}@cs.ucla.edu}
}

\maketitle
\thispagestyle{empty}

\begin{abstract}
We introduce a simple modification of local image descriptors, such as SIFT, based on pooling gradient orientations across different domain sizes, in addition to spatial locations. The resulting descriptor, which we call DSP-SIFT, outperforms other methods in wide-baseline matching benchmarks, including those based on convolutional neural networks, despite having the same dimension of SIFT and requiring no training. 
\end{abstract}

\section{Introduction}

Local image descriptors, such as SIFT  \cite{lowe04distinctive} and its variants, are designed to reduce variability due to illumination and vantage point while retaining discriminative power. This facilitates finding correspondence between different views of the same underlying scene. In a wide-baseline matching task on the Oxford benchmark \cite{mikolajc03survey,mikolajczyk04comparison}, nearest-neighbor SIFT descriptors achieve a mean average precision (mAP) of \jcomment{$27.50\%$, a $71.85\%$} improvement over direct comparison of normalized grayscale values. Other datasets yield similar results \cite{moreels2007evaluation}. Functions that reduce sensitivity to nuisance variability can also be learned from data \cite{memisevic,ranzato2007unsupervised,susskind,taylor,winder2007learning}. Convolutional neural networks (CNNs) can been trained to ``learn away'' nuisance variability while retaining class labels using large annotated datasets. In particular, \cite{fischer2014descriptor} use\jcomment{s} (patches of) natural images as surrogate classes and add\jcomment{s} transformed versions to train the network to discount nuisance variability. The activation maps in response to image values can be interpreted as a descriptor and used for correspondence. \cite{dosovitskiy2013unsupervised,fischer2014descriptor} show that the CNN outperforms SIFT, albeit with a much larger dimension. Here we show that a simple modification of SIFT, obtained by pooling gradient orientations across different domain sizes (``scales''), in addition to spatial locations, improves it by a considerable margin, also outperforming the best CNN. We call the resulting descriptor ``domain-size pooled'' SIFT, or DSP-SIFT.

\begin{figure}[tb]
\begin{center}
\vspace{-.2cm}
\subfigure[]{\label{fig-sub-a}\includegraphics[width=.21\columnwidth]{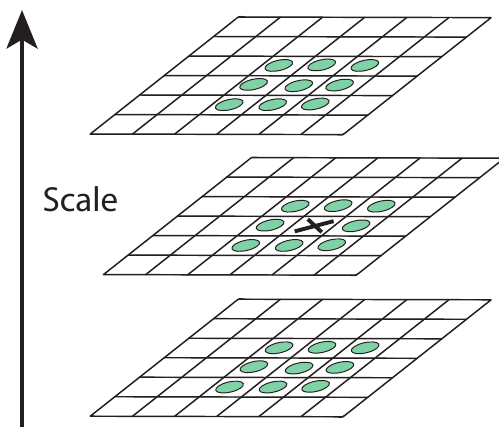}}
\subfigure[]{\label{fig-sub-b}\includegraphics[width=.21\columnwidth]{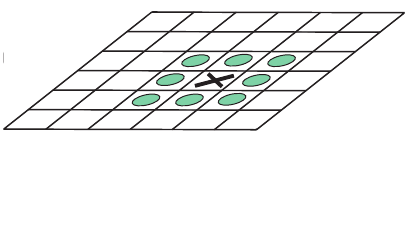}}
\subfigure[]{\label{fig-sub-c}\includegraphics[width=.21\columnwidth]{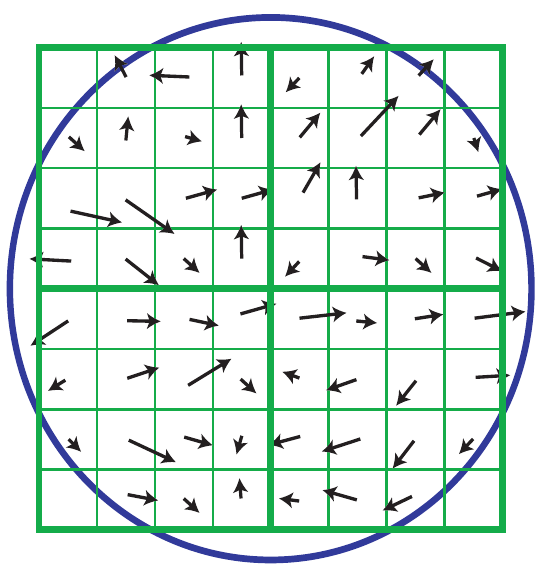}}
\subfigure[]{\label{fig-sub-d}\includegraphics[width=.21\columnwidth]{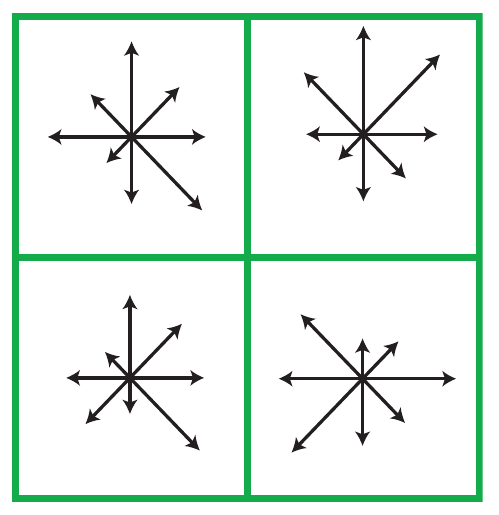}}
\subfigure[]{\label{fig-sub-e}\includegraphics[width=.063\columnwidth]{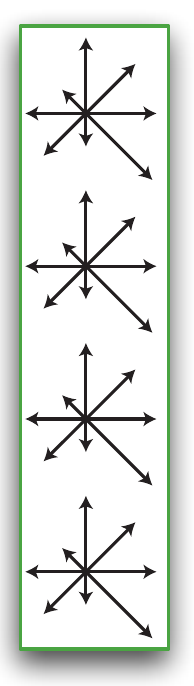}}\vspace{-.2cm}
\subfigure{\includegraphics[width=.21\columnwidth]{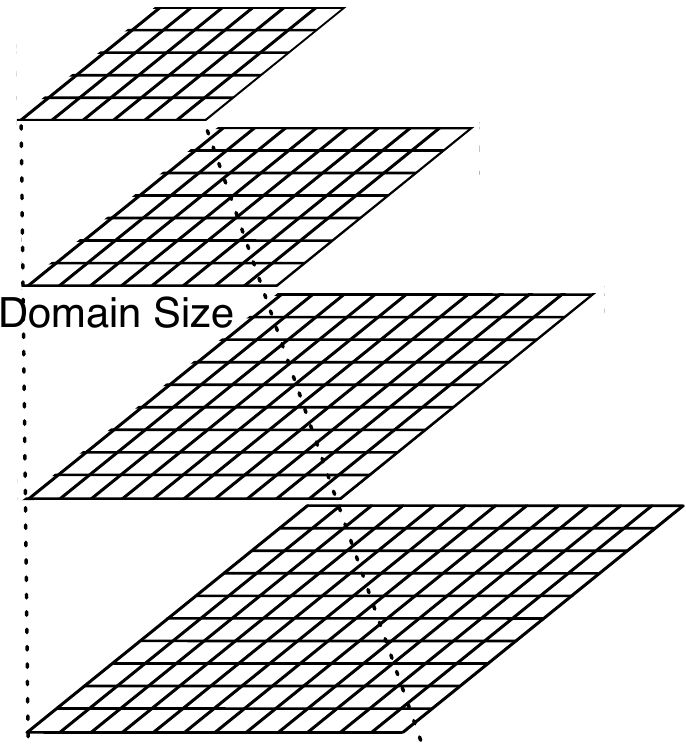}}
\subfigure{\includegraphics[width=.21\columnwidth]{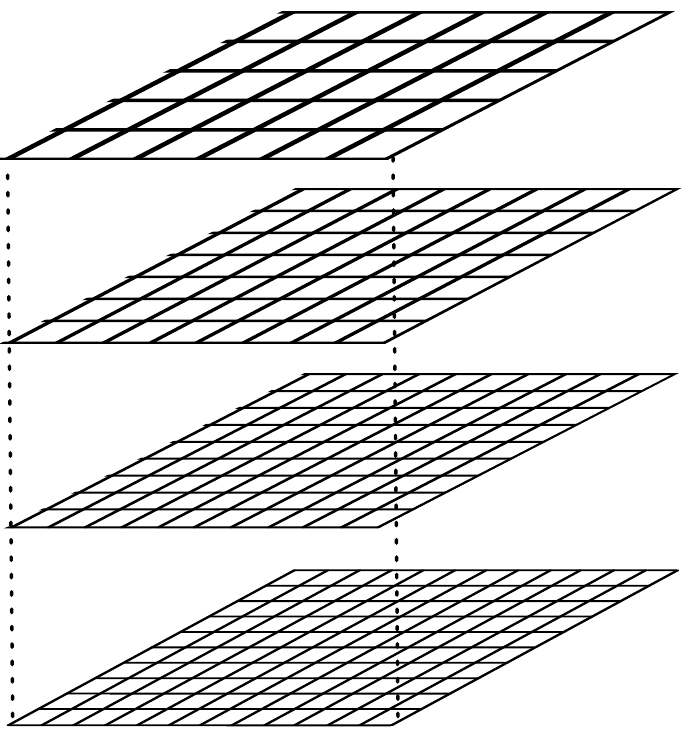}}
\subfigure{\includegraphics[width=.21\columnwidth]{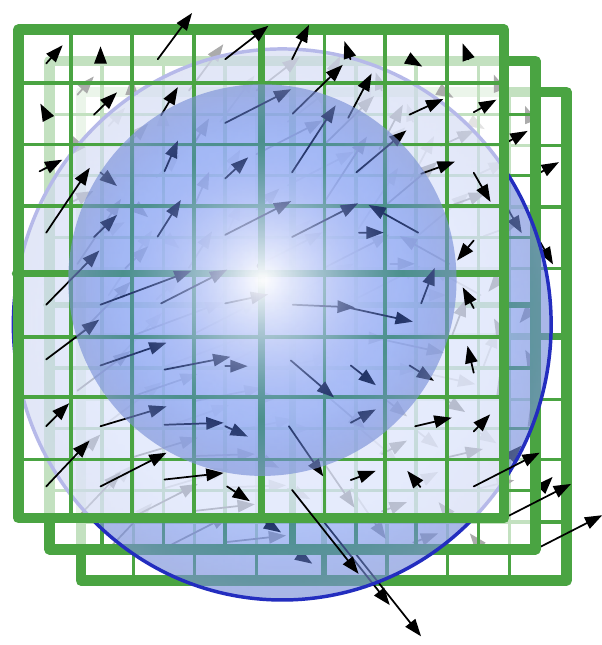}}
\subfigure{\includegraphics[width=.21\columnwidth]{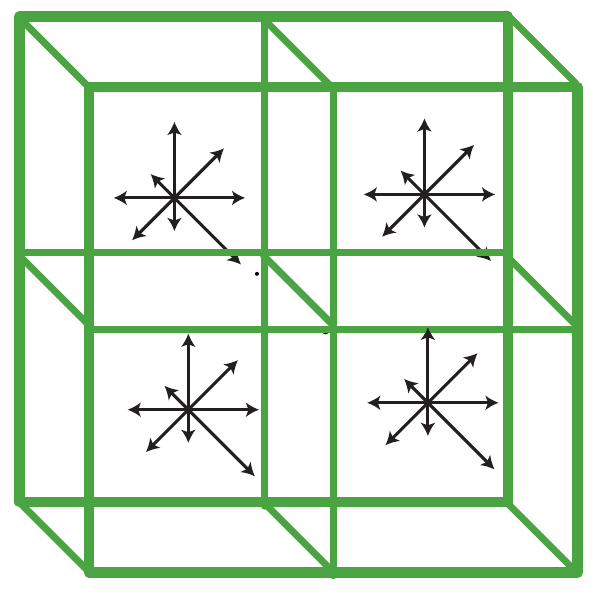}}
\subfigure{\includegraphics[width=.063\columnwidth]{DSP-SIFT-vec.pdf}}\vspace{-.2cm}
\end{center}
\caption{In SIFT (top, recreated according to \cite{lowe04distinctive}) isolated scales are selected \subref{fig-sub-a} and the descriptor constructed from the image at the selected scale \subref{fig-sub-b} by computing  gradient orientations \subref{fig-sub-c} and pooling them in spatial neighborhoods \subref{fig-sub-d} yielding histograms that are \newcomment{concatenated and normalized} to form the descriptor \subref{fig-sub-e}. In DSP-SIFT (bottom), pooling occurs across different domain sizes \subref{fig-sub-a}: Patches of different sizes are re-scaled \subref{fig-sub-b}, gradient orientation computed \subref{fig-sub-c} and pooled across locations {\em and} scales \subref{fig-sub-d}, and concatenated yielding a descriptor \subref{fig-sub-e} of the same dimension of ordinary SIFT.}
\label{fig-visualize}
\end{figure}

Pooling across different domain sizes is implemented in few lines of code, can be applied to any histogram-based method (Sect.~\ref{sect-how}), and yields a descriptor of the same size that outperforms the original essentially uniformly (Fig.~\ref{fig-scat}). Yet combining histograms of images of different sizes is counterintuitive and  seemingly at odds with the teachings of scale-space theory and the resulting established practice of {\em scale selection} \cite{lindeberg98} (Sect.~\ref{sect-related}). It is, however, rooted in classical sampling theory and anti-aliasing. Sect.~\ref{sect-what} describes {\em what} we do, Sect.~\ref{sect-how} {\em how} we do it, and Sect.~\ref{sect-why} {\em why} we do it. \newcomment{Sect.}~\ref{sect-expm} validates our method empirically.

\subsection{Related work} 
\label{sect-related}

A single, un-normalized cell of the ``scale-invariant feature transform'' SIFT \cite{lowe04distinctive} and its variants \cite{surf,chandrasekhar2009chog,dalalT05} can be written compactly as a formula \cite{dongKDHBS15,vlfeat}:
\be 
h_{\rm SIFT} (\theta | I, \hat \sigma)[x] =  
\int \mathcal{N}_{\epsilon}\left(\theta - \angle \nabla I(y) \right) \mathcal{N}_{\hat \sigma}(y-x) d\mu(y)
\label{eq-sift}
\ee 
where $I$ is the image restricted to a square domain, centered at a location $x  \in \Lambda(\hat \sigma)$ with size $\hat \sigma$ in the lattice $\Lambda$ determined by the response to a difference-of-Gaussian (DoG) operator across all locations and scales (SIFT {\em detector}). Here $d\mu(y) \doteq \| \nabla I(y) \| dy$, $\theta$ is the independent variable, ranging from $0$ to $2\pi$, corresponding to an orientation histogram bin of size $\epsilon$, and $\hat \sigma$ is the {\em spatial pooling scale}. The kernel ${\cal N}_\epsilon$ is bilinear of size $\epsilon$ and ${\cal N}_{\hat \sigma}$ separable-bilinear of size $\hat \sigma$ \cite{vlfeat}, although they could be replaced by a Gaussian with standard deviation $\hat \sigma$ and an {\em angular Gaussian}\cut{ \cite{watson1983statistics}} with dispersion parameter $\epsilon$.  The SIFT descriptor is the concatenation of $16$ cells \eqref{eq-sift} computed at locations $x \in \{x_{1}, x_{2}, \dots, x_{16}\}$ on a $4\times 4$ lattice $\Lambda$, and normalized. 

The spatial pooling scale $\hat \sigma$ and the size of the image domain where the SIFT descriptor is computed $\Lambda = \Lambda (\hat \sigma)$ are {\em tied} to the photometric characteristics of the image, since $\hat \sigma$ is derived from the response of a DoG operator on the (single) image.\footnote{Approaches based on ``dense SIFT'' forgo the detector and instead compute descriptors on a regular sampling of locations and scales (Fig. \ref{fig-unc-pri}). However, no existing dense SIFT method performs domain-size pooling.} Such a response depends on the {\em reflectance} properties of the scene and {\em optical characteristics} and {\em resolution} of the sensor, neither of which is related to the size and shape of co-visible (corresponding) regions. Instead, how large a portion of a scene is visible in each corresponding image(s) depends on the {\em shape} of the scene, the {\em pose} of the two cameras, and the resulting visibility ({\em occlusion}) relations. Therefore, we propose to {\em untie} the size of the domain where the descriptor is computed (``scale'') from photometric characteristics of the image, departing from the teachings of scale selection \jcomment{(Fig.~\ref{fig-scale-size}).} Instead, we use basic principles of classical sampling theory and {\em anti-aliasing} to achieve robustness to domain size changes due to occlusions (Sect.~\ref{sect-why}).

Pooling is commonly understood as {\em the combination of responses of feature detectors/descriptors at nearby locations, aimed at transforming the joint feature representation into a more usable one that preserves important information} (intrinsic variability) {\em while discarding irrelevant detail} (nuisance variability) \cite{boureau2010theoretical,jia2012beyond}. However, precisely how pooling trades off these two conflicting aims is unclear and mostly addressed empirically in end-to-end comparisons with numerous confounding factors.
Exceptions include \cite{boureau2010theoretical}, where intrinsic and nuisance variability are combined and abstracted into the variance and distance between the means of scalar random variables in a binary classification task. For more general settings, the goals of reducing nuisance variability while preserving intrinsic variability is elusive as a {\em single image} does not afford the ability to separate the two \cite{dongKDHBS15}. 

\jcomment{An alternate interpretation of pooling as anti-aliasing \cite{soattoC14ICLR} clearly highlights its effects on intrinsic and nuisance variability: Because one cannot know what portion of an object or scene will be visible in a test image, a scale-space (``semi-orbit'') of domain sizes (``receptive fields'') should be marginalized or searched over (``max-out''). Neither can be computed in closed-form, so the semi-orbit has to be sampled. To reduce complexity,  only a small number of samples should be retained, resulting in undersampling and aliasing phenomena that can be mitigated by anti-aliasing, with quantifiable effects on the sensitivity to nuisance variability. For the case of histogram-based descriptors, anti-aliasing planar translations consists of spatial pooling, routinely performed by most descriptors. Anti-aliasing visibility results in {\em domain-size aggregation}, which no current descriptor practices. This interpretation also offers a way to quantify the effects of pooling on discriminative (reconstruction) power directly, using classical results from sampling theory, rather than indirectly through an end-to-end classification experiment that may contain other confounding factors. 
}

Domain-size pooling can be applied to a number of different descriptors or convolutional architectures.  We illustrate its effects on the most popular, SIFT. \jcomment{However, we point out that proper marginalization requires the availability of multiple images of the same scene, and therefore cannot be performed in a single image. While most local image descriptors are computed from a single image, exceptions include \cite{dongKDHBS15, LeeS10}. Of course, multiple images can be ``hallucinated'' from one, but the resulting pooling operation can only achieve invariance to modeled transformations. }

In neural network architectures, there is evidence that abstracting spatial pooling hierarchically, \ie aggregating nearby responses in feature maps, is beneficial \cite{boureau2010theoretical}. This process could be extended by aggregating across different neighborhood sizes in feature space. To the best of our knowledge, the only architecture that performs some kind of pooling across scales is \cite{serre2007feedforward}, although the justification provided in \cite{rosasco} only concerns translation within each scale. The same goes for \cite{mallatB11}, where pooling (low-pass filtering) is only performed within each scale, and not across scales. Other works learn the regions for spatial pooling, for instance \cite{jia2012beyond,simonyan2014learning}, but still restrict pooling to within-scale, similar to \cite{lecun2012learning}, rather than across scales as we advocate. 

We distinguish {\em multi-scale methods} that concatenate descriptors computed {\em independently at each scale}, from {\em cross-scale pooling}, where statistics of the image at different scales are combined directly in the descriptor. Examples of the former include \cite{hassner2012sifts}, where ordinary SIFT descriptors computed on domains of different size are assumed to belong to a linear subspace, and \cite{simonyan2014learning}, where Fisher vectors are computed for multiple sizes and aspect ratios and spatial pooling occurs within each level. 
\jcomment{Also bag-of-word (BoW) methods \cite{sivic2003video}, as mid-level representations, aggregate different low level descriptors by counting their frequency after discretization. Typically, vector quantization or other clustering technique is used, each descriptor is associated with a cluster center (``word''), and the frequency of each word is recorded in lieu of the descriptors themselves. This can be done for domain size, by computing different descriptors at the same location, for different domain sizes, and then counting frequencies relative to a dictionary learned from a large training dataset (Sect.~\ref{sect-bow}).}

Aggregation across time, which may include changes of domain size, is advocated in \cite{hamel2011temporal}, but in the absence of formulas it is unclear how this approach relates to our work. In \cite{farabet2012scene}, weights are shared across scales, which is not equivalent to pooling, but still establishes some dependencies across scales. MTD \cite{leeS11} appears to be the first instance of pooling across scales, although the aggregation is global in scale-space with consequent loss of discriminative power. Most recently, \cite{gong2014multi} advocates the same but in practice space-pooled VLAD descriptors obtained at different scales are simply concatenated. Also \cite{berg01geometric} can be thought of as a form of pooling, but the resulting descriptor only captures the mean of the resulting distribution. \jcomment{In addition, \cite{tau2014dense} exploits the possibility of estimating the proper scales for nearby features via scale propagation but still no pooling is performed across scales. Additional details in related prior work are discussed in Appendix \ref{sect-relation}.} 

\section{Domain-Size Pooling}
\label{sect-what}

If SIFT is written as \eqref{eq-sift}, then DSP-SIFT is given by 
\be
h_{\rm DSP}(\theta | I)[x] =  \int h_{\rm SIFT} (\theta | I, \sigma)[x]  {\cal E}_{s}(\sigma) d\sigma  ~~~ x\in \Lambda
\label{eq-new-sift}
\ee
where $s>0$ is the size-pooling scale and $\cal E$ is an exponential or other unilateral density function.\cut{This is our main contribution.} The process is visualized in Fig.~\ref{fig-visualize}. Unlike SIFT, that is computed on a scale-selected lattice $\Lambda(\hat\sigma)$, DSP-SIFT is computed on a {\em regularly sampled} lattice $\Lambda$. Computed on a different lattice, the above can be considered as a recipe for DSP-HOG \cite{dalalT05}. Computed on a tree, it can be used to extend deformable-parts models (DPM) \cite{felzenswalb} to DSP-DPM. Replacing $h_{\rm SIFT}$ with other  histogram-based descriptor ``X'' (for instance, SURF \cite{surf}), the above yields DSP-X. Applied to a hidden layer of a convolutional network, it yields a DSP-CNN, or DSP-Deep-Fisher-Network \cite{simonyan2013deep}. The details of the implementation are in Sect.~\ref{sect-how}.

While the implementation of DSP is straightforward, its justification is less so. We report the summary in Sect. \ref{sect-why} and the detailed derivation in Appendix \ref{sect-derivation}, that provides a theoretical justification and conditions under which the resulting descriptors are valid. In Sect.~\ref{sect-expm} we compare DSP-SIFT to alternate approaches. Motivated by the experiments of \cite{mikolajczyk04comparison,moreels2007evaluation} that compare local descriptors, we choose SIFT as a paragon and compare it to DSP-SIFT on the standard benchmark \cite{mikolajczyk04comparison}. Motivated by \cite{fischer2014descriptor} that compares SIFT to both supervised and unsupervised CNNs trained on Imagenet and Flickr respectively on the same benchmark \cite{mikolajczyk04comparison}, we submit DSP-SIFT to the same protocol. We also run the test on the new synthetic dataset introduced by \cite{fischer2014descriptor}, that yields the same qualitative assessment.\cut{ It should be noted that the comparison is unfair in favor of the CNNs, due to its increased dimension compared to SIFT and DSP-SIFT.  Moreover, the best performance  of a CNN is obtained using its fourth layer  responses, that contain $8,192$ coefficients, a $64$-fold complexity increase, even without accounting for the cost of learning, which is none for DSP-SIFT.}

Clearly, domain-size pooling of under-sampled semi-orbits cannot outperform fine sampling, so if we were to retain all the scale samples instead of aggregating them, performance would further improve. However, computing and matching a large collection of SIFT descriptors across different scales would incur significantly increased computational and storage costs. To contain the latter, \cite{hassner2012sifts} assumes that descriptors at different scales populate a linear subspace and fit a high-dimensional hyperplane. \jcomment{The resulting Scale-less SIFT (SLS)} outperforms ordinary SIFT as shown in Fig.~\ref{fig-tradeoff}. However, the linear subspace assumption breaks when considering large scale changes, so SLS is outperformed by DSP-SIFT despite the considerable difference in (memory and time) complexity.

\vspace{.3cm}
\section{Implementation and Parameters}
\label{sect-how}

\begin{figure}[t]
\begin{center}
\subfigure{\label{fig-params-range}\includegraphics[width=.44\columnwidth]{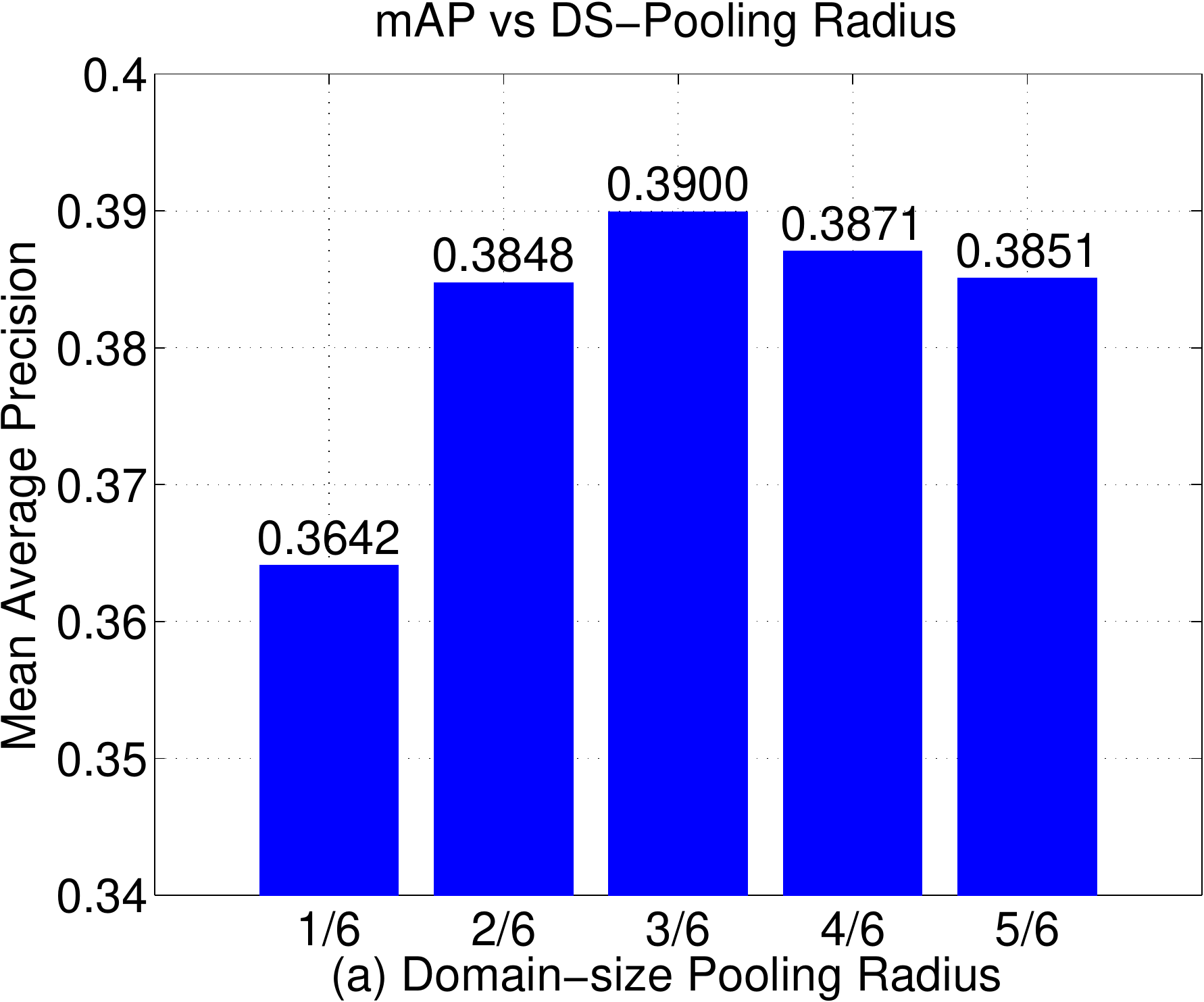}}\hspace{.2cm}
\subfigure{\label{fig-params-ns}\includegraphics[width=.44\columnwidth]{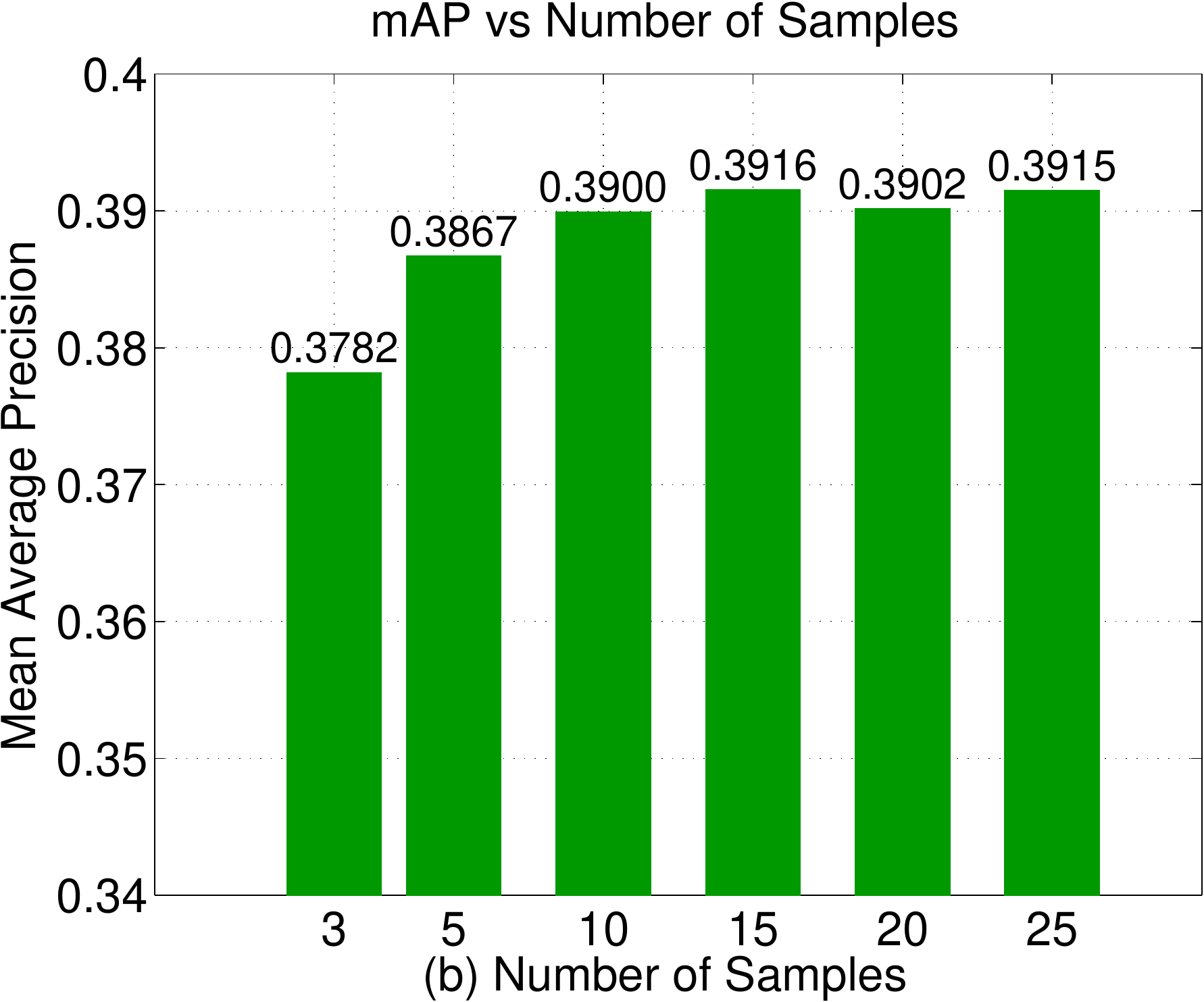}}\vspace{-.2cm}
\end{center}
   \caption{{\sl Mean Average Precision for different parameters.} \subref{fig-params-range} shows that mAP changes with the radius $s$ of DS pooling. The best mAP is achieved at $\hat s$ $ = \hat \sigma /2$; \subref{fig-params-ns} shows mAP as a function of the number of samples used within the best range ($\hat \sigma - \hat s, \hat \sigma + \hat s$).}
\label{fig-params}
\end{figure}

Following other evaluation protocols, we use {\em Maximally Stable Extremal Regions} (MSER) \cite{matas03robust} to detect candidate regions, affine-normalize, re-scale and align them to the dominant orientation. For a detected scale $\hat \sigma$, {DSP-SIFT} samples $N_{\hat \sigma}$ scales within a neighborhood $(\lambda_{1} \hat \sigma, \lambda_{2} \hat \sigma)$ around it. For each scale-sampled patch, a single-scale un-normalized SIFT descriptor \eqref{eq-sift} is computed on the SIFT scale-space octave corresponding\footnote{This is an updated version of the protocol described in \cite{fischer2014descriptor}, as discussed in detail in Appendix \ref{sect-ds}.} to the sampled scale $\sigma$. By choosing ${\cal E}_s$ to be a uniform density, these raw histograms of gradient orientations at different scales are accumulated and normalized\footnote{We follow the practice of SIFT \cite{lowe04distinctive} to normalize, clamp and re-normalize the histograms, with the clamping threshold set to 0.067 empirically.} \eqref{eq-new-sift}. Fig.~\ref{fig-params-range} shows the mean average precision (defined in Sect.~\ref{sect-metrics}) for different domain size pooling ranges. Improvements are observed as soon as more than one scale is used, with diminishing return: Performance decreases with domain size pooling radius exceeding $\hat \sigma / 2$. Fig.~\ref{fig-params-ns} shows the effect of the number of size samples used to construct DSP-SIFT. Although the more samples the merrier, three size samples are sufficient to outperform ordinary SIFT, and improvement beyond $10$ samples is minimal. Additional samples do not further increase the mean average precision, but incur more computational cost. In the evaluation in Sect.~\ref{sect-expm}, we use $\lambda_1 = 1/6, \lambda_2 = 4/3$ and $N_{\hat \sigma} = 15$. These parameters are empirically selected on the Oxford dataset \cite{mikolajc03survey, mikolajczyk04comparison}.

\section{Validation}
\label{sect-expm}

As a baseline, the RAW-PATCH descriptor (named following \cite{fischer2014descriptor}) is the unit-norm grayscale intensity of the affine-rectified and resized patch of a fixed size ($91 \times 91$). 

The standard SIFT, which is widely accepted as a paragon \cite{mikolajc03survey,moreels2007evaluation}, is computed using the VLFeat library \cite{vlfeat}. Both SIFT and DSP-SIFT are computed on the SIFT scale-space corresponding to the detected scales. Instead of mapping all patches to an arbitrarily user-defined size, we use the area of each selected and rectified MSER region to determine the octave level in the scale-space where SIFT (as well as DSP-SIFT) is to be computed. 

Scale-less SIFT (SLS) is computed using the source code provided by the authors \cite{hassner2012sifts}: For each selected and rectified patch, the standard SIFT descriptors are computed at $20$ scales from a scale range of $(0.5, 12)$, and the standard PCA subspace dimension is set to $8$, yielding a final descriptor of dimension $8256$ after a subspace-to-vector mapping. 

\newcomment{To compare DSP-SIFT to a convolutional neural network}, we use the top-performer in \cite{fischer2014descriptor}, \newcomment{an unsupervised model} pre-trained on $16000$ natural images undergoing $150$ transformations each (total $2.4$M). The responses at the intermediate layers $3$ ({CNN-L3}) and $4$ ({CNN-L4}) are used for comparison, following \cite{fischer2014descriptor}. Since the network requires input patches of fixed size, we tested and report the results on both $69\times69$ (PS69) and $91\times91$ (PS91) as in \cite{fischer2014descriptor}.

Although no direct comparison with Multiscale Template Descriptors (MTD) \cite{leeS11} is performed, SLS can be considered as dominating it since it uses all scales without collapsing them into a single histogram. The derivation in Sect.~\ref{sect-why} suggests, and empirical evidence in Fig.~\ref{fig-params-range} confirms, that aggregating the histogram across {\em all} scales significantly reduces discriminative power. Sect.~\ref{sect-bow} compares DSP-SIFT to a BoW which pools SIFT descriptors computed at different sizes at the same location. 

\subsection{Datasets}
The Oxford dataset \cite{mikolajc03survey,mikolajczyk04comparison} comprises $40$ pairs of images of mostly planar scenes seen under different pose, distance, blurring, compression and lighting. They are organized into $8$ categories undergoing increasing magnitude of transformations. While routinely used to evaluate descriptors, this dataset has limitations in terms of size and restriction to mostly planar scenes, modest scale changes, and no occlusions. Fischer {\em et al.} \cite{fischer2014descriptor} recently introduced a dataset of $400$ pairs of images with more extreme transformations including zooming, blurring, lighting change, rotation, perspective and nonlinear transformations. 

\subsection{Metrics}
\label{sect-metrics}
Following \cite{mikolajc03survey}, we use precision-recall (PR) curves to evaluate descriptors. A {\em match} between two descriptors is called if their Euclidean distance is less than a threshold $\tau_d$. It is then labeled as a {\em true positive} if the area of intersection over union (IoU) of their corresponding MSER-detected regions is larger than $50\%$. Both datasets provide ground truth mapping between images, so the overlapping is computed by warping the first MSER region into the second image and then computing the overlap with the second MSER region. {\em Recall} is the fraction of true positives over the total number of correspondences. {\em Precision} is the percentage of true matches within the total number of matches. By varying the distance threshold $\tau_d$, a PR curve can be generated and {\em average precision} (AP, {\em a.k.a} {\em area  under the curve}, AUC) can be estimated. The average of APs provides the {\em mean average precision} (mAP) scores used for comparison. 

\begin{figure*}[t]
\begin{center}
\subfigure{\includegraphics[width=.19\textwidth]{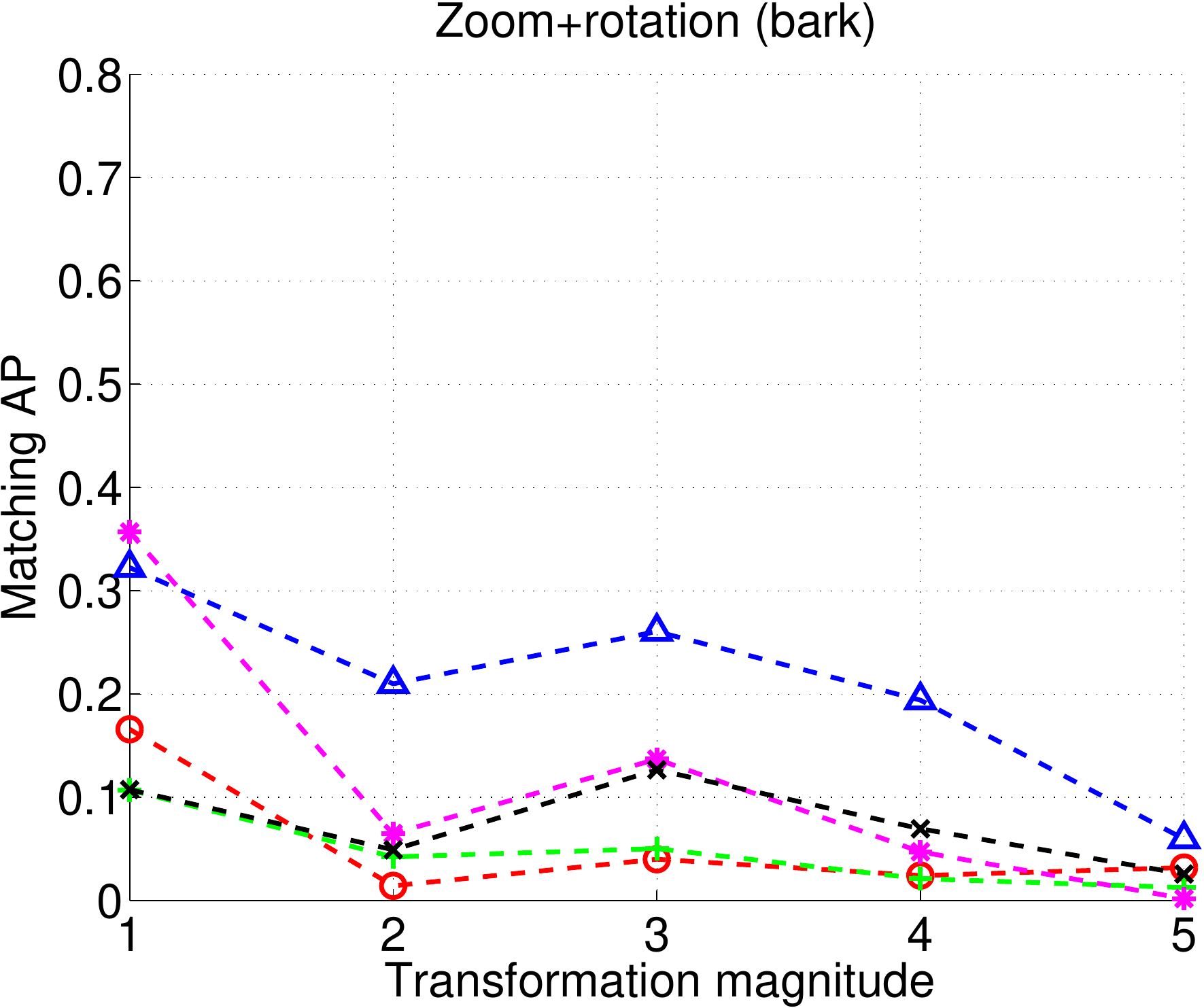}}
\subfigure{\includegraphics[width=.19\textwidth]{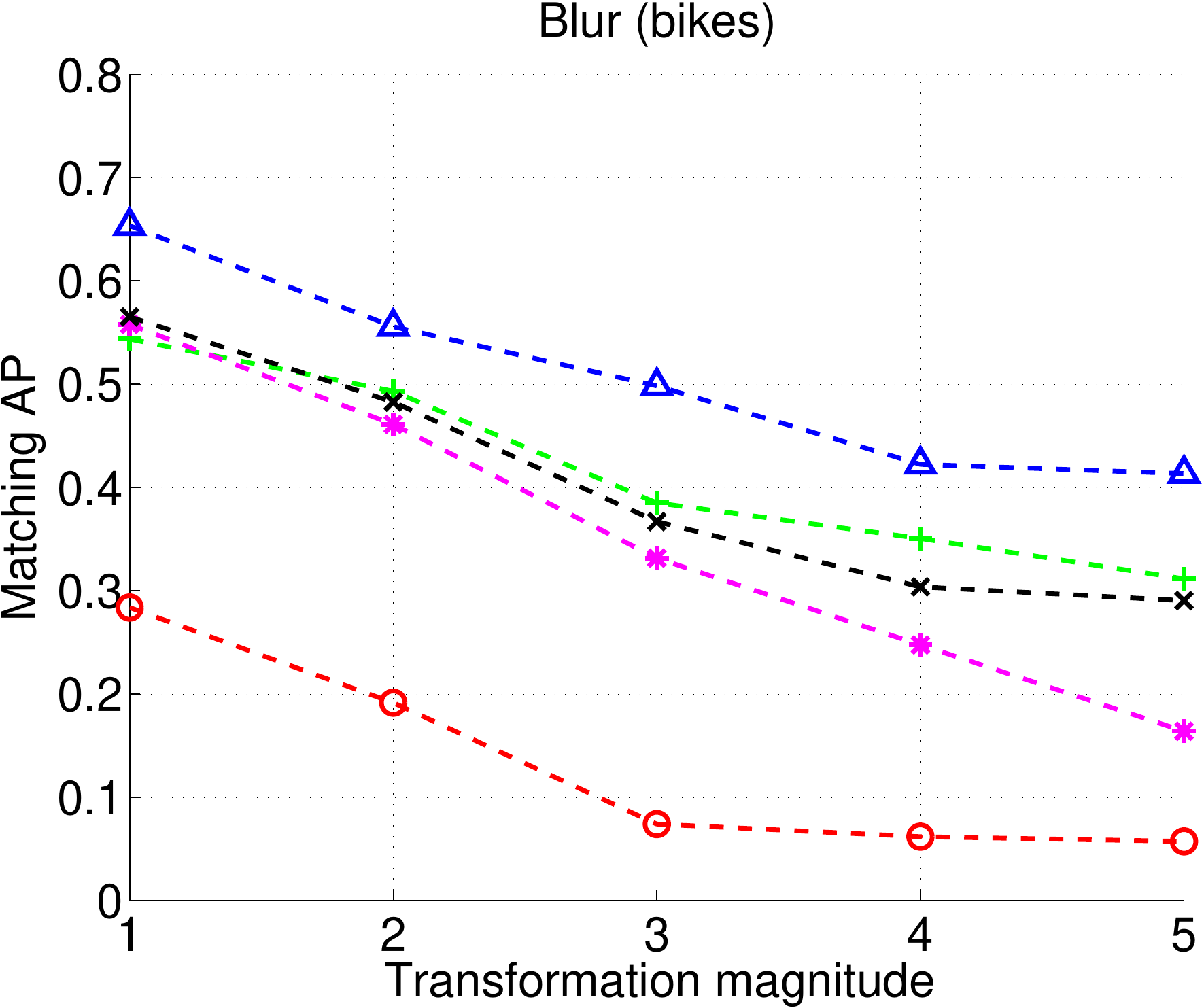}}
\subfigure{\includegraphics[width=.19\textwidth]{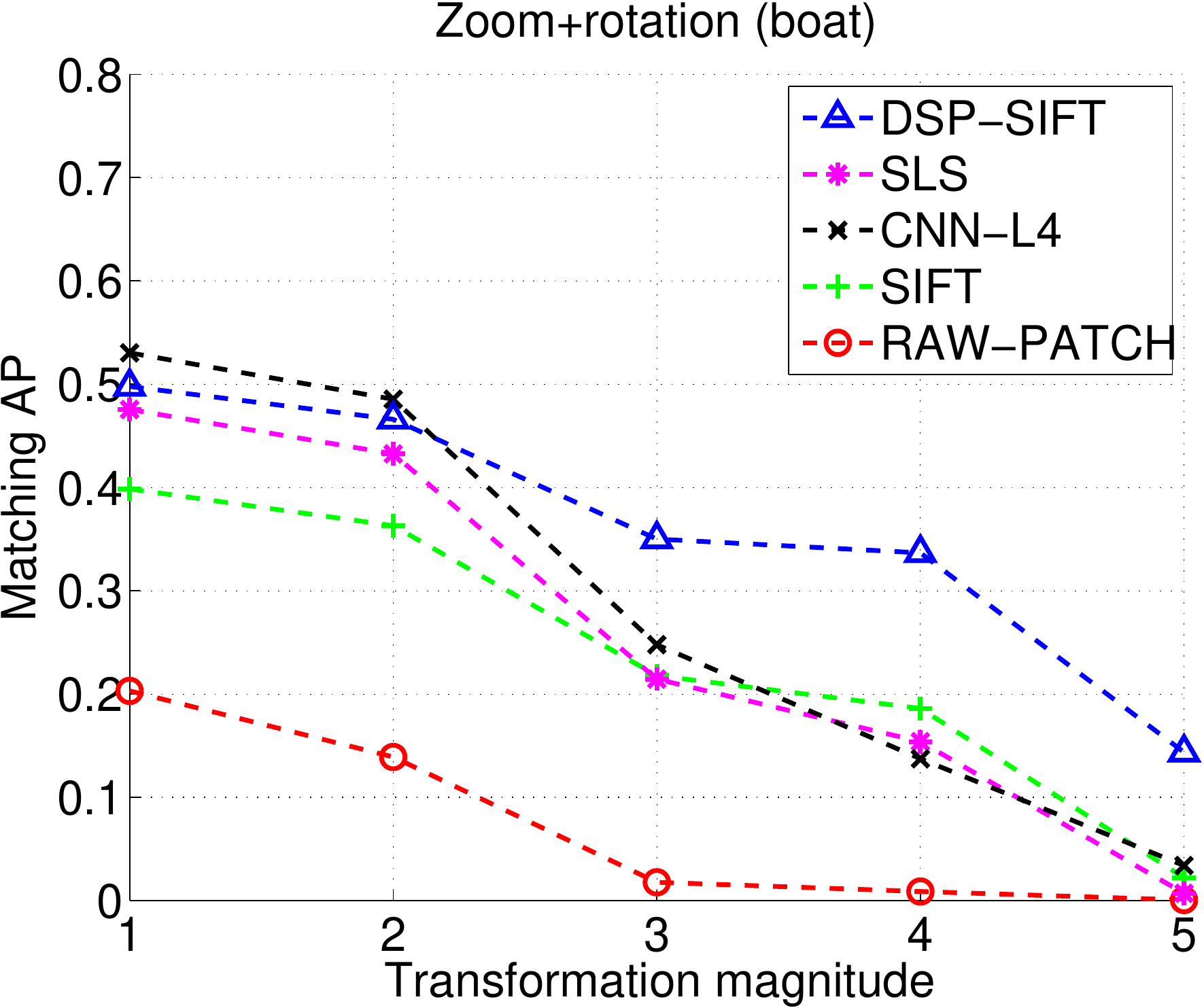}}\hspace{.2cm}
\subfigure{\includegraphics[width=.00185\textwidth]{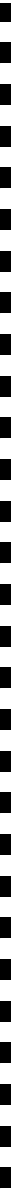}} \hspace{.1cm}
\subfigure{\includegraphics[width=.188\textwidth]{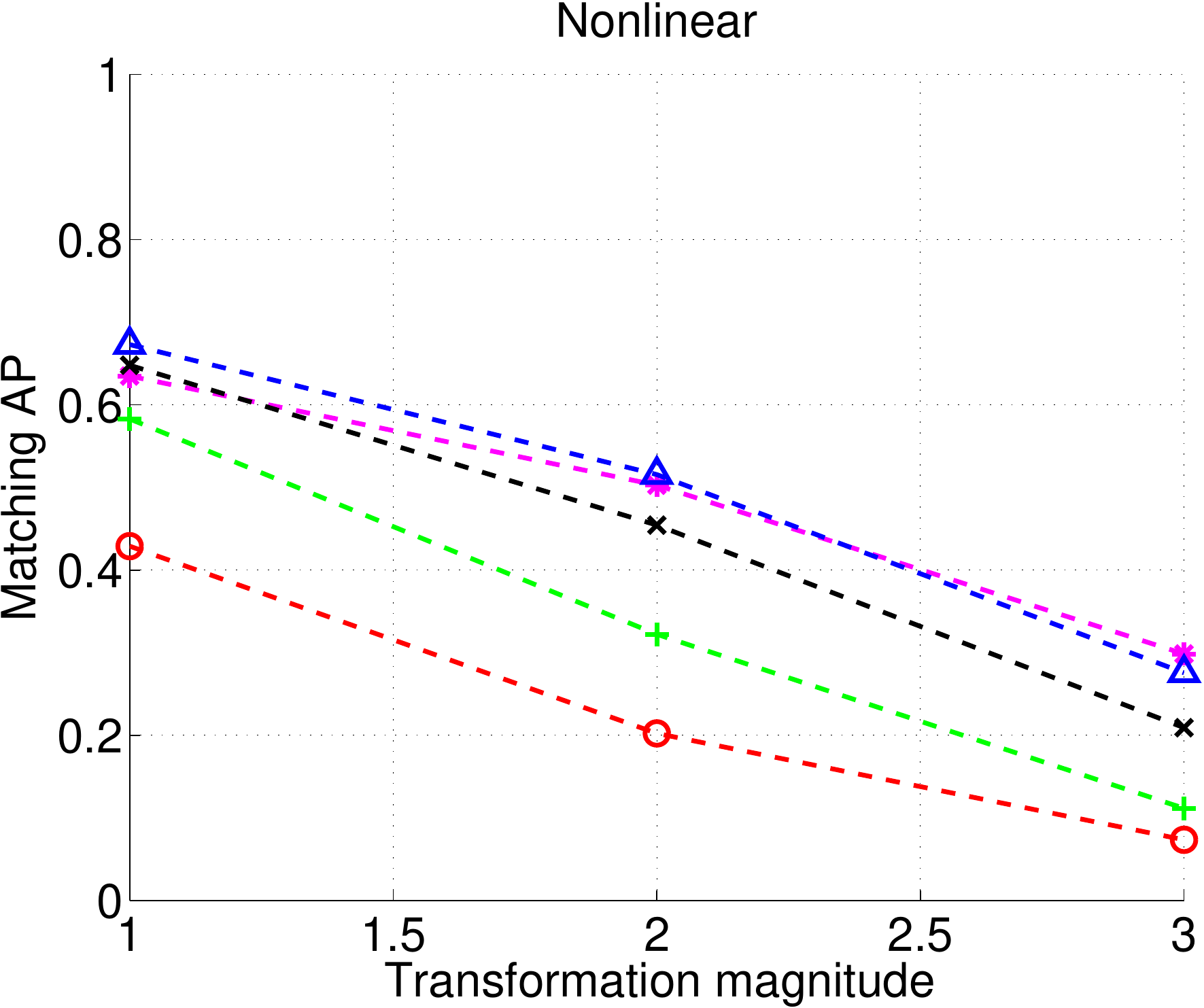}}\vspace{-.3cm}
\subfigure{\includegraphics[width=.187\textwidth]{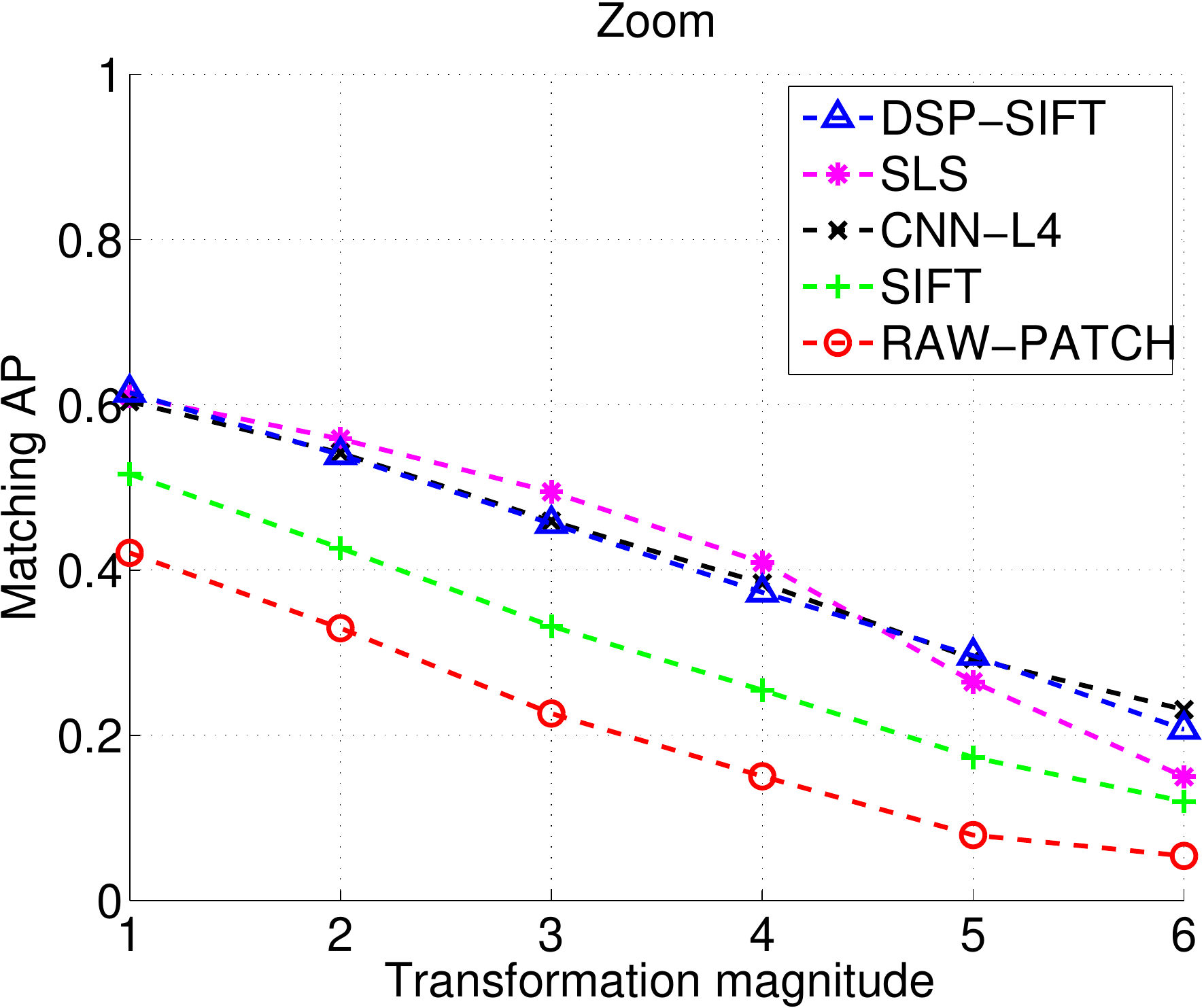}}
\subfigure{\includegraphics[width=.19\textwidth]{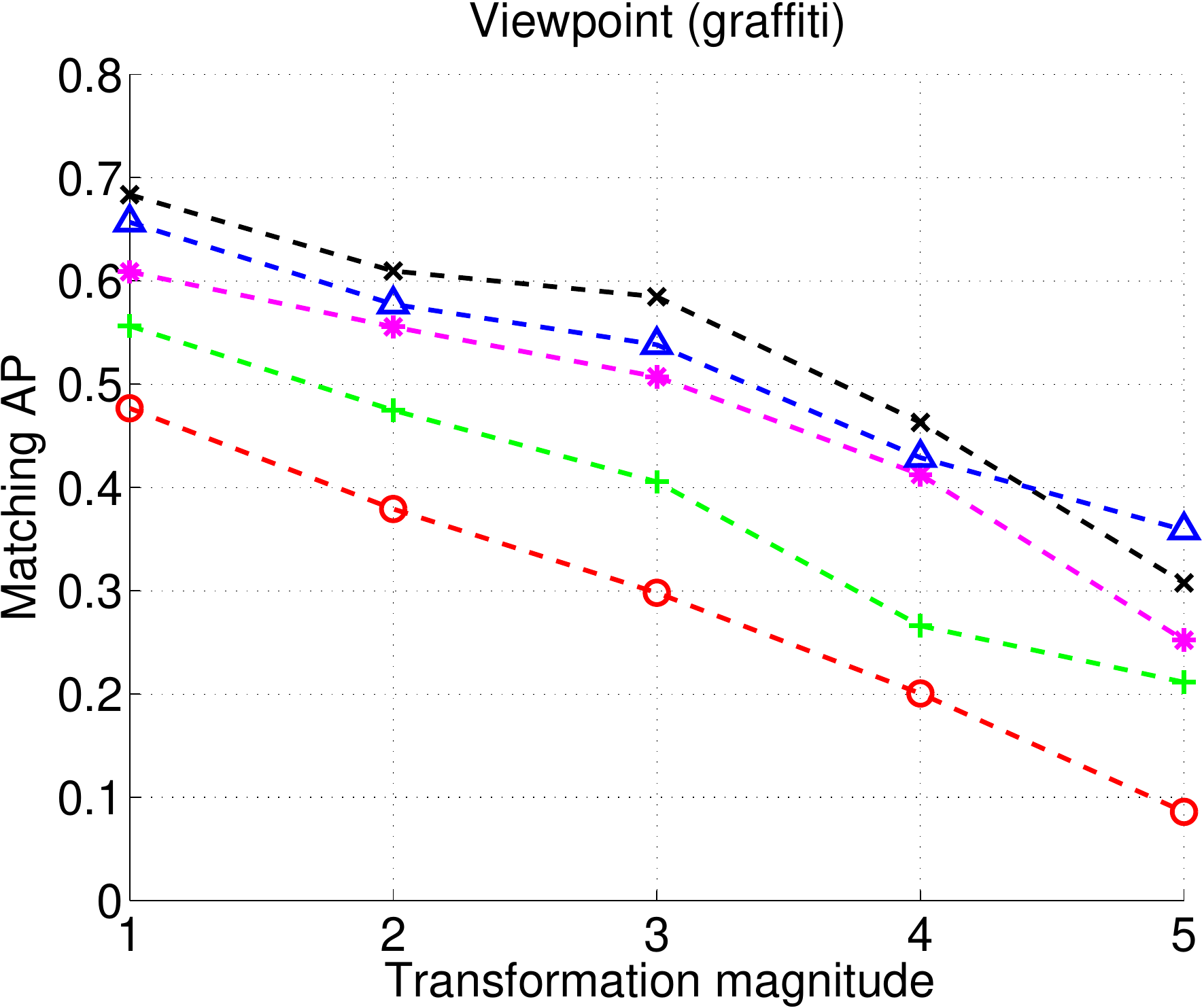}}
\subfigure{\includegraphics[width=.19\textwidth]{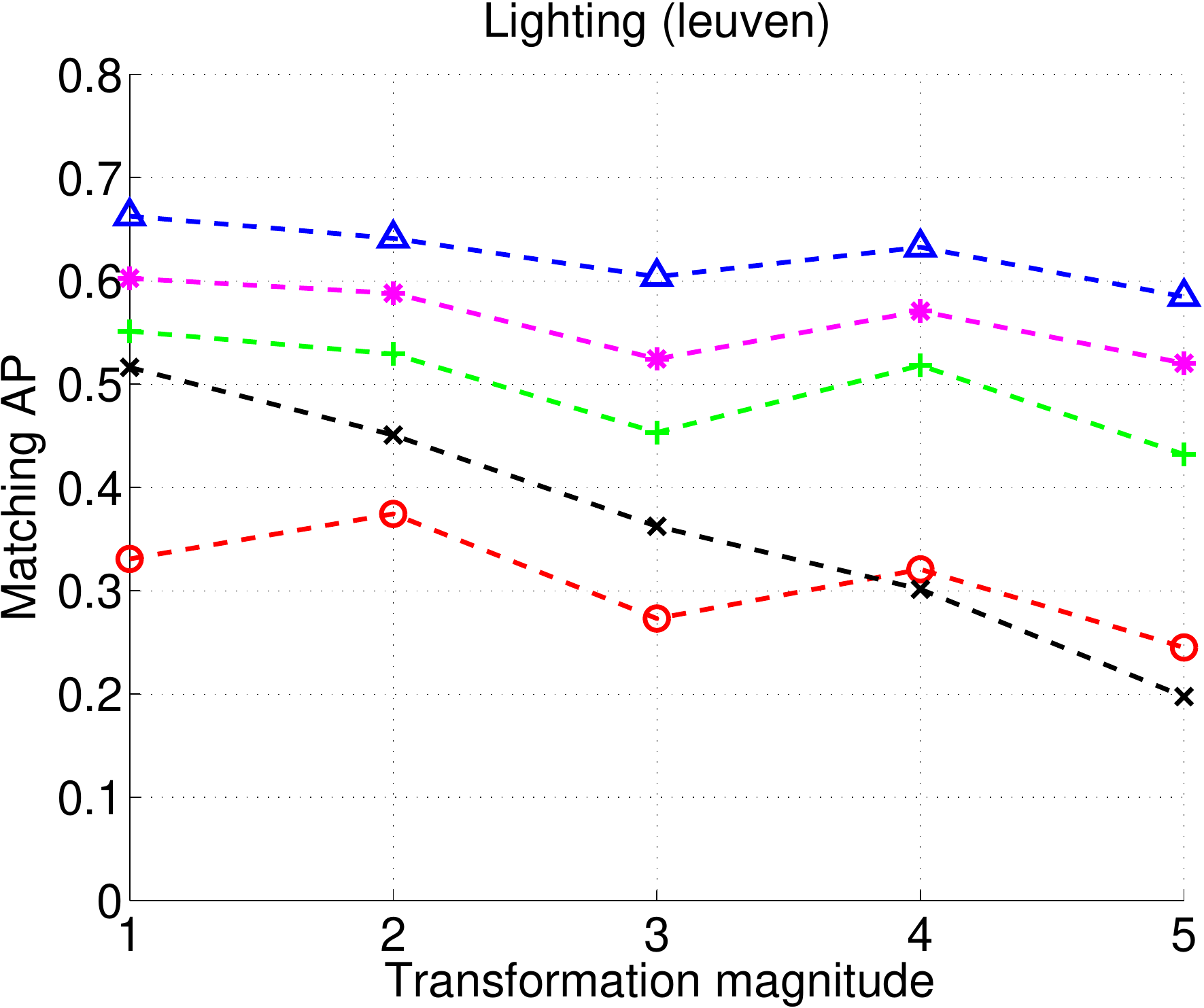}}
\subfigure{\includegraphics[width=.19\textwidth]{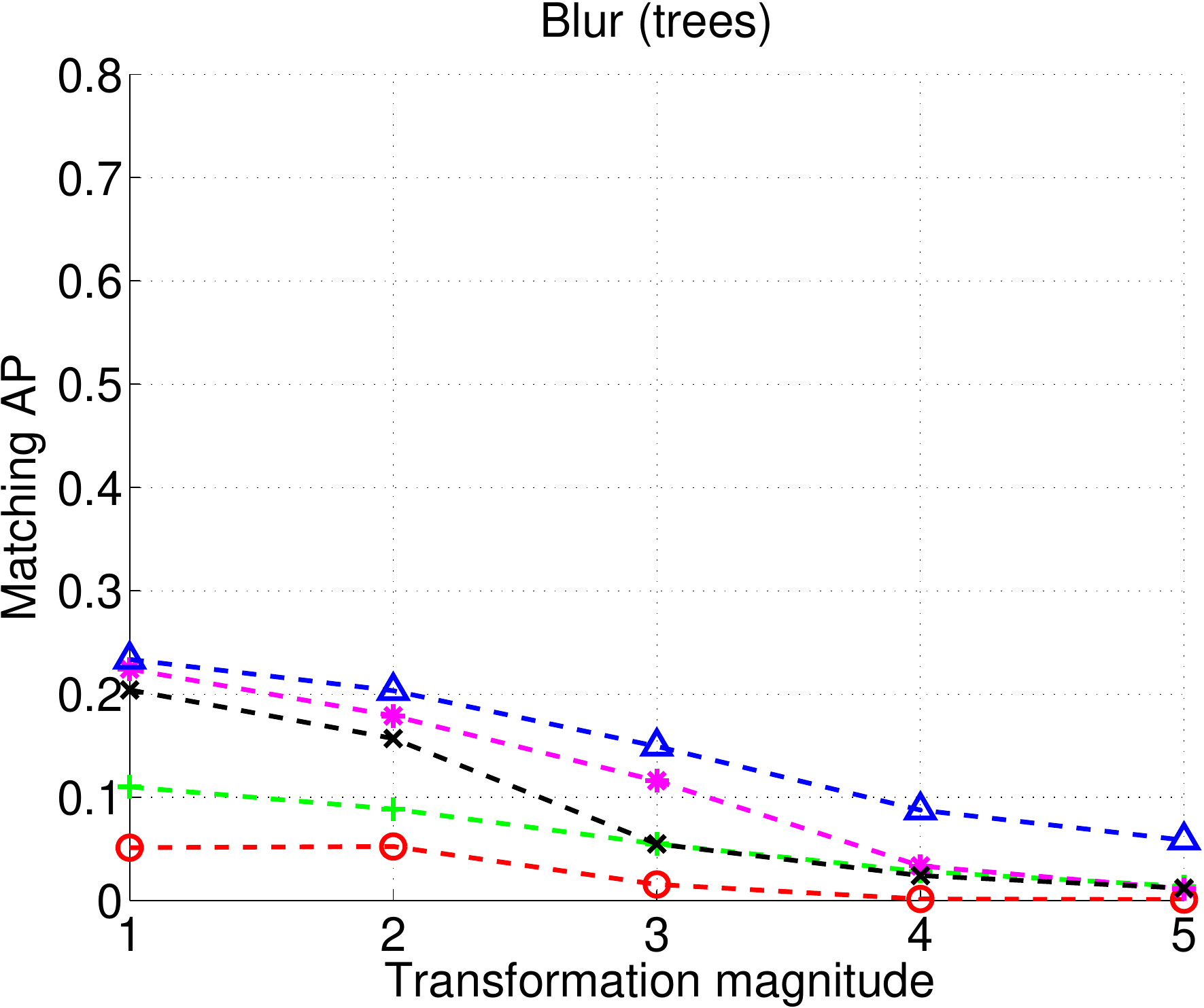}}\hspace{.2cm}
\subfigure{\includegraphics[width=.00185\textwidth]{./vsep.pdf}}\hspace{.1cm}
\subfigure{\includegraphics[width=.19\textwidth]{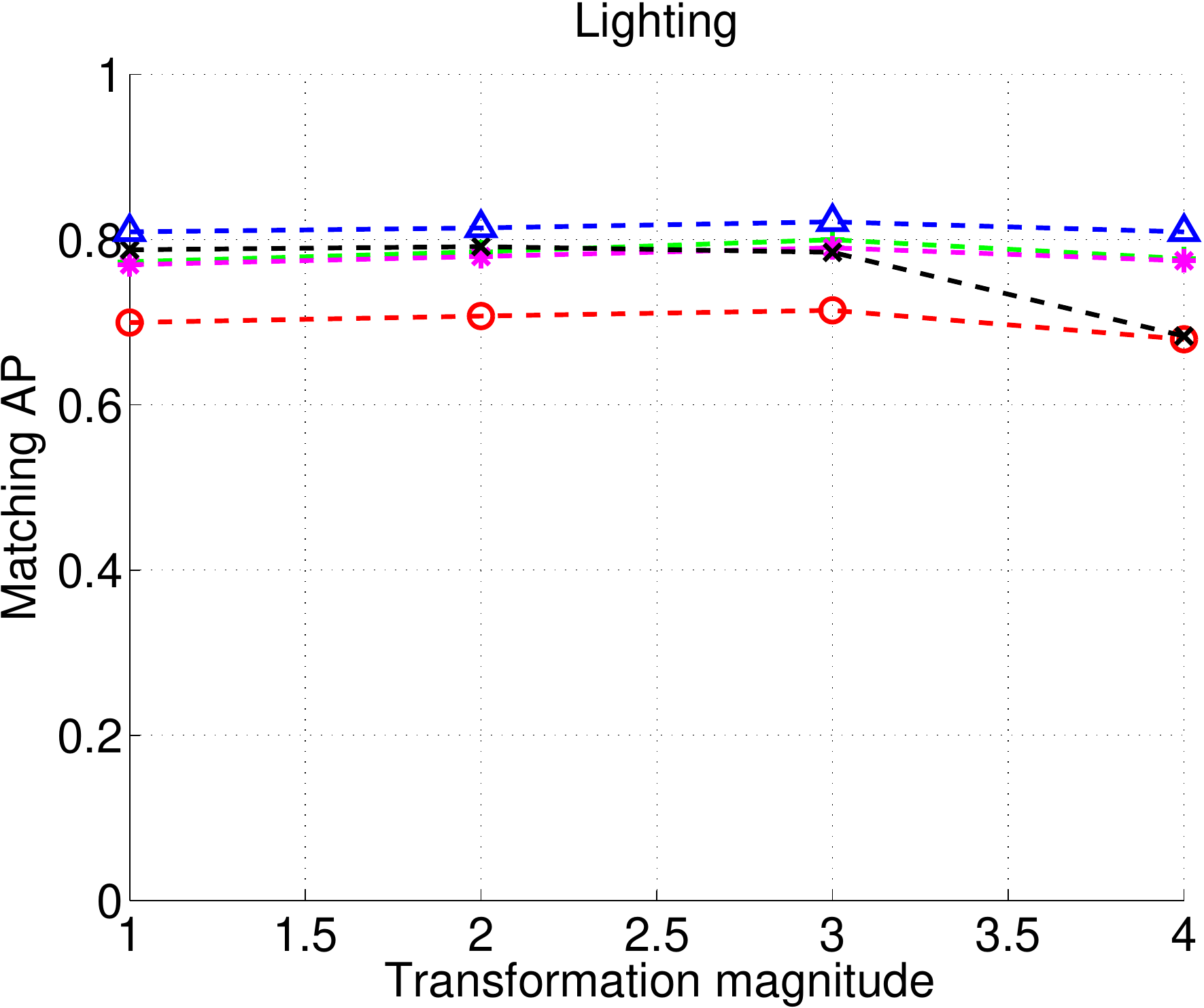}}\vspace{-.3cm}
\subfigure{\includegraphics[width=.19\textwidth]{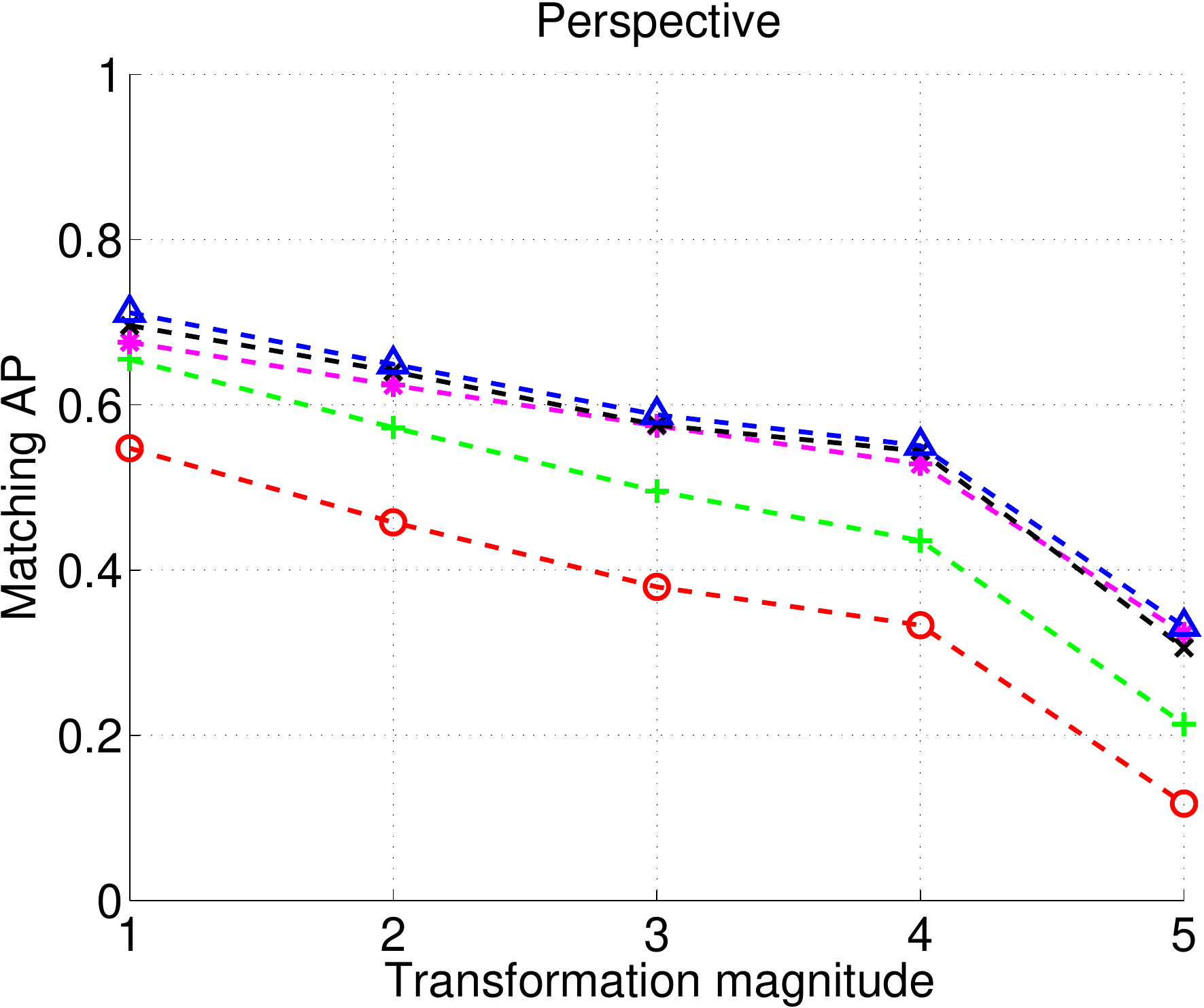}}
\subfigure{\includegraphics[width=.19\textwidth]{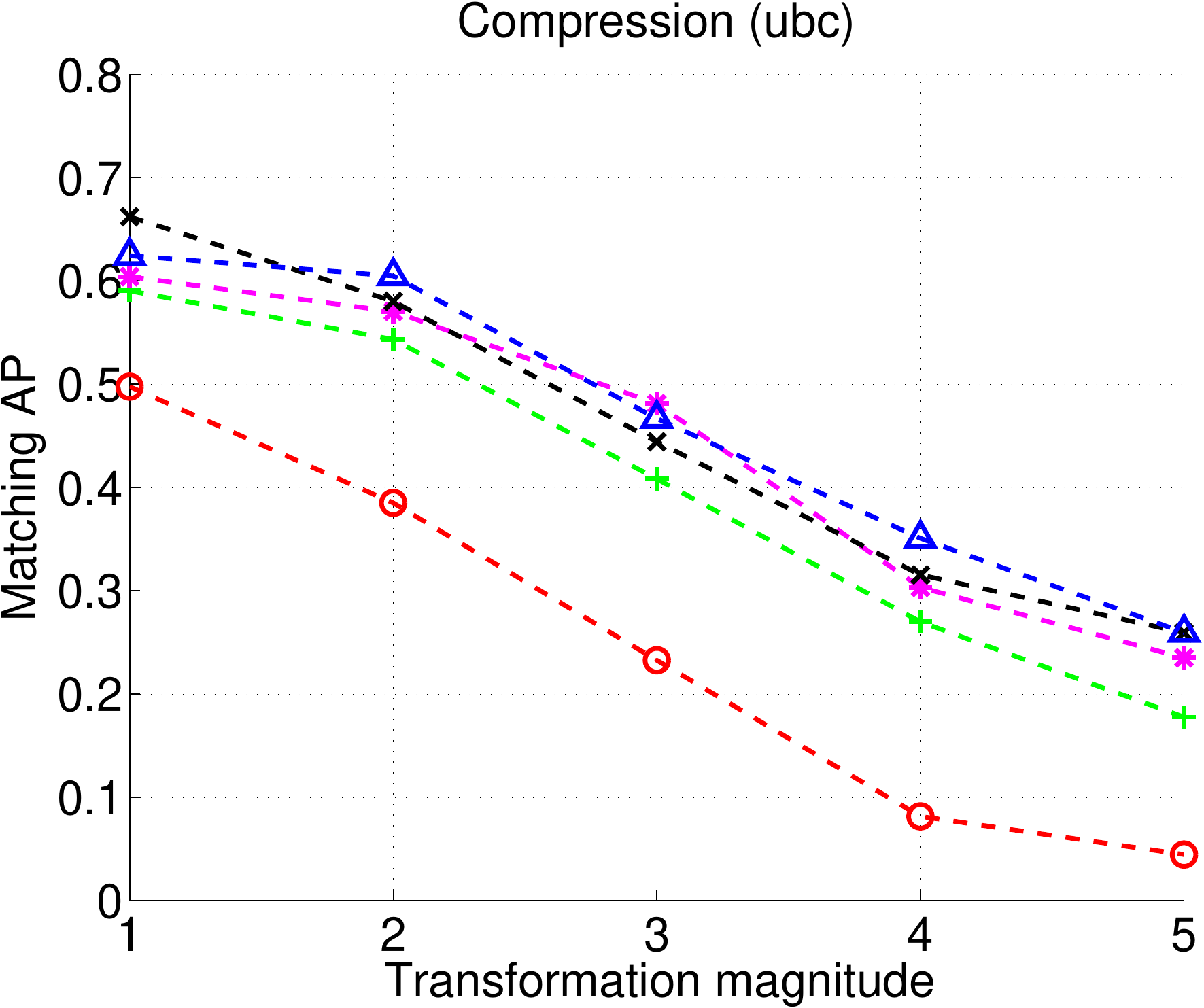}}
\subfigure{\includegraphics[width=.19\textwidth]{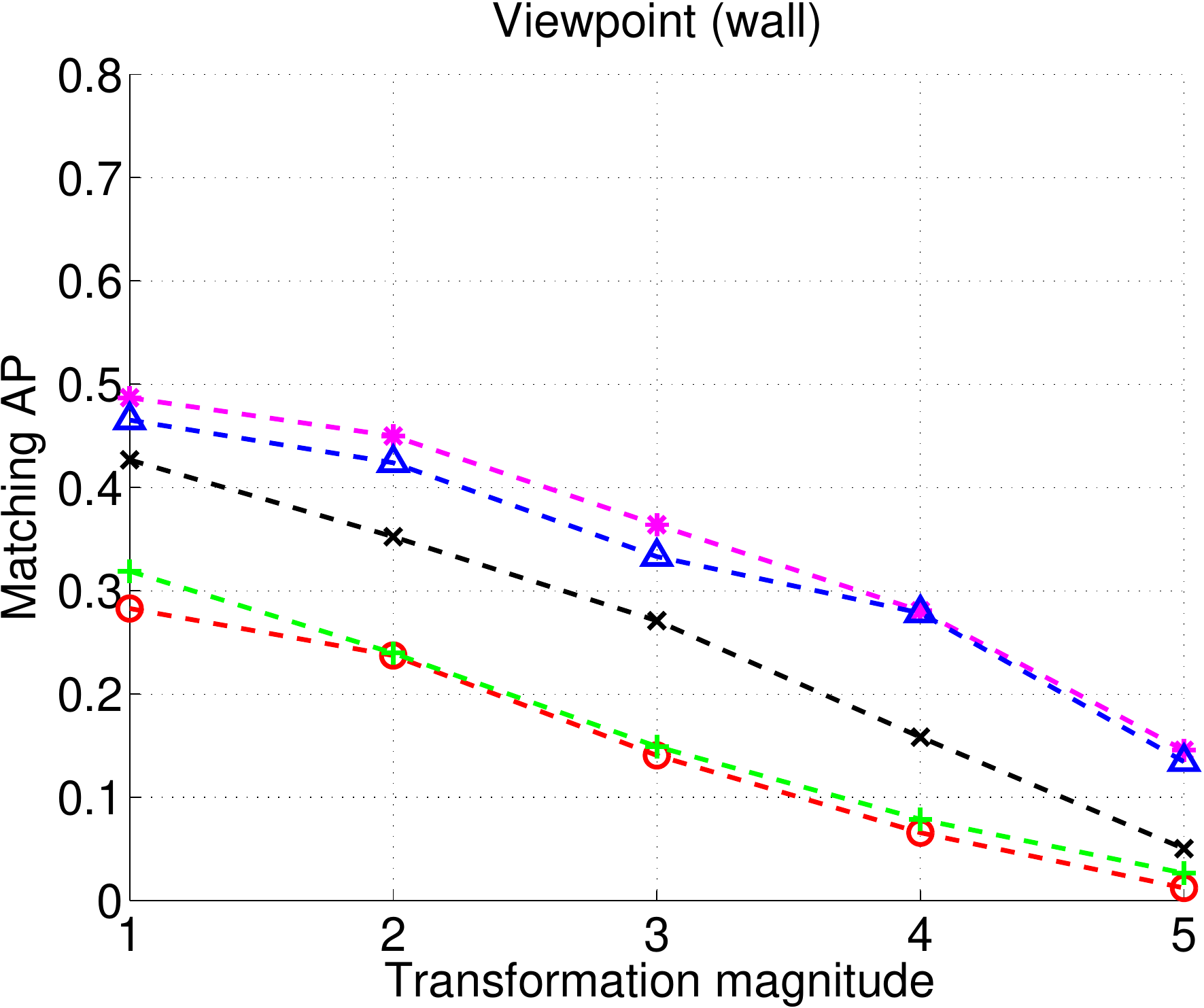}}
\subfigure{\includegraphics[width=.19\textwidth]{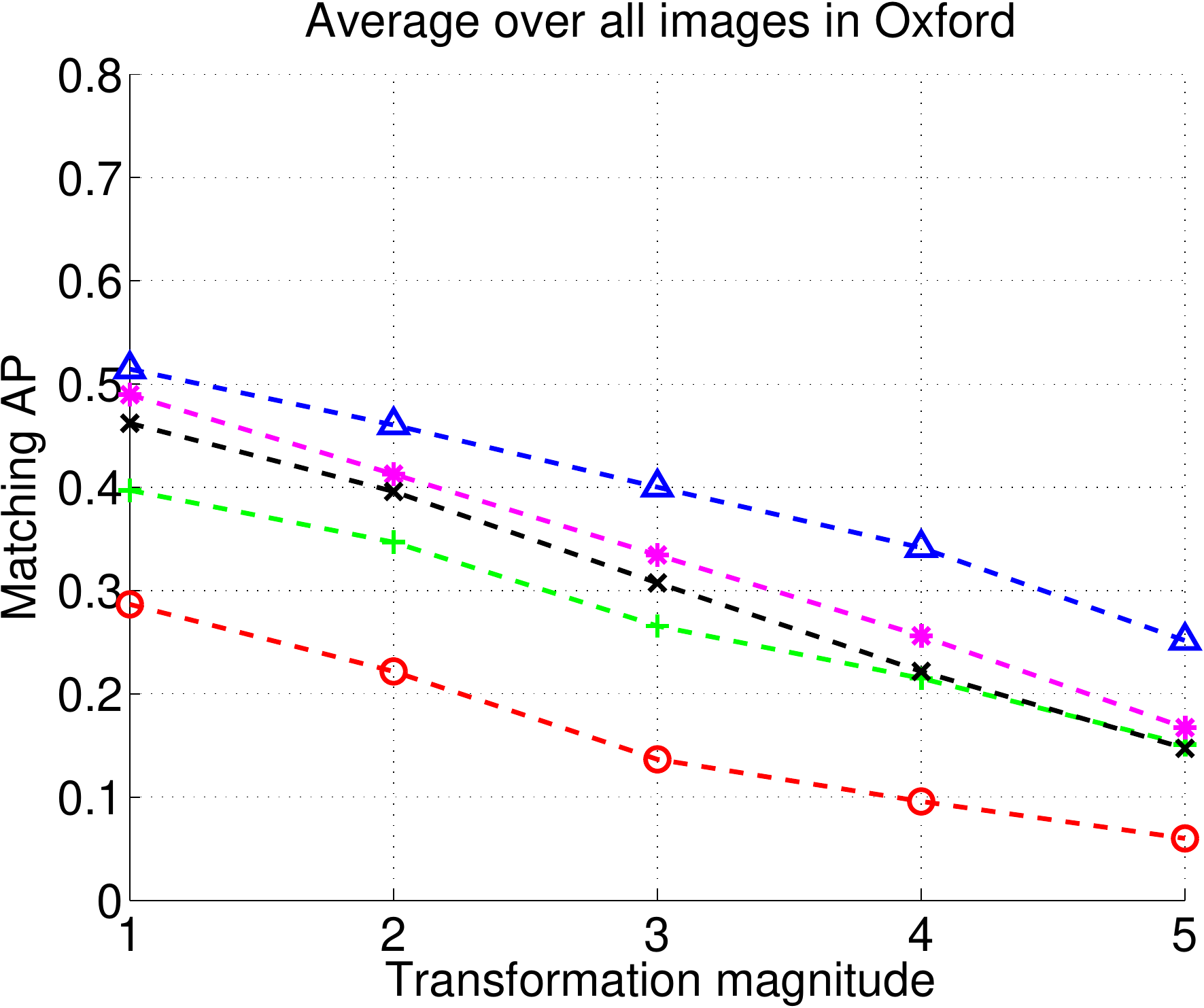}}\hspace{.2cm}
\subfigure{\includegraphics[width=.00185\textwidth]{./vsep.pdf}}\hspace{.1cm}
\subfigure{\includegraphics[width=.19\textwidth]{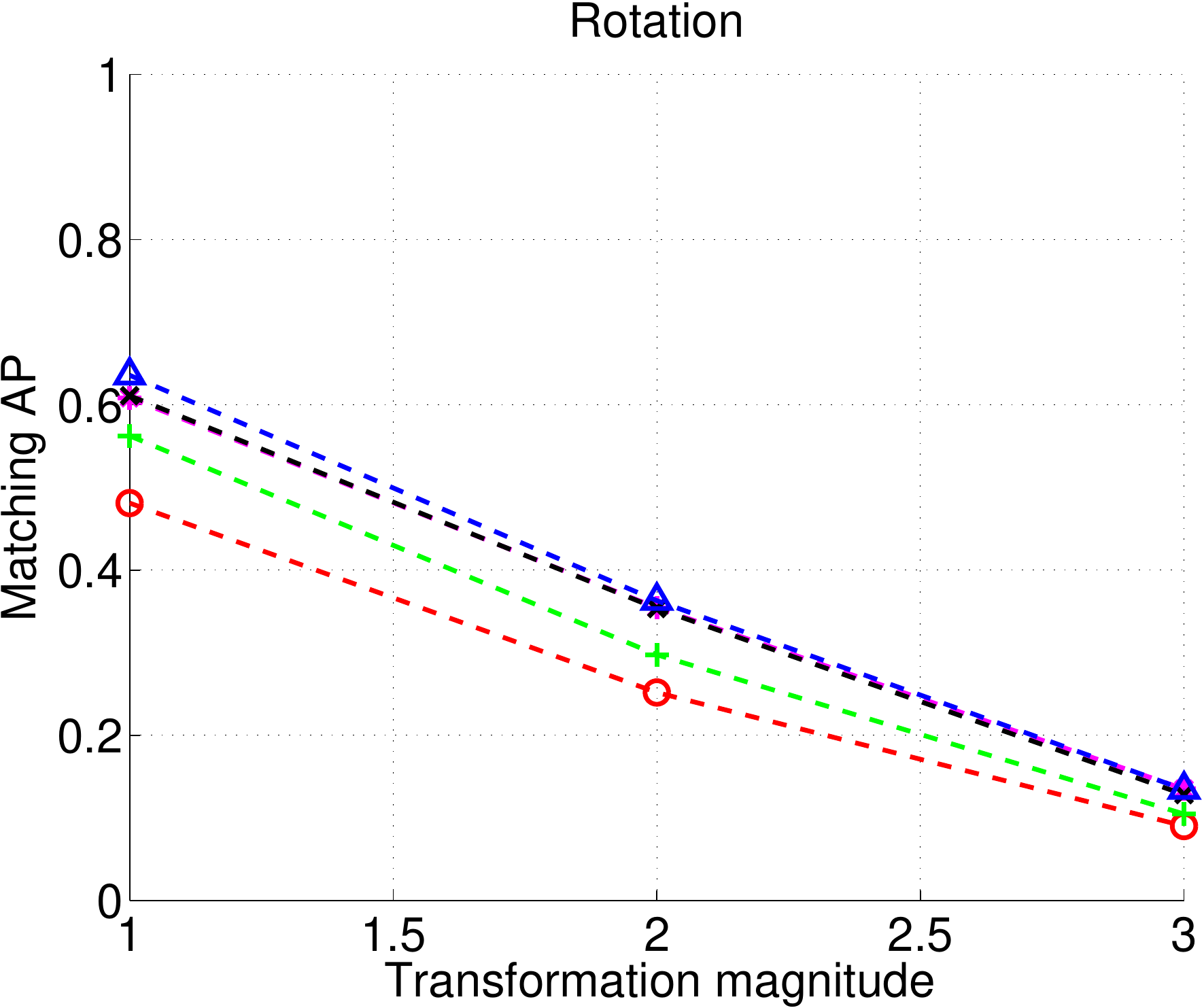}}
\subfigure{\includegraphics[width=.19\textwidth]{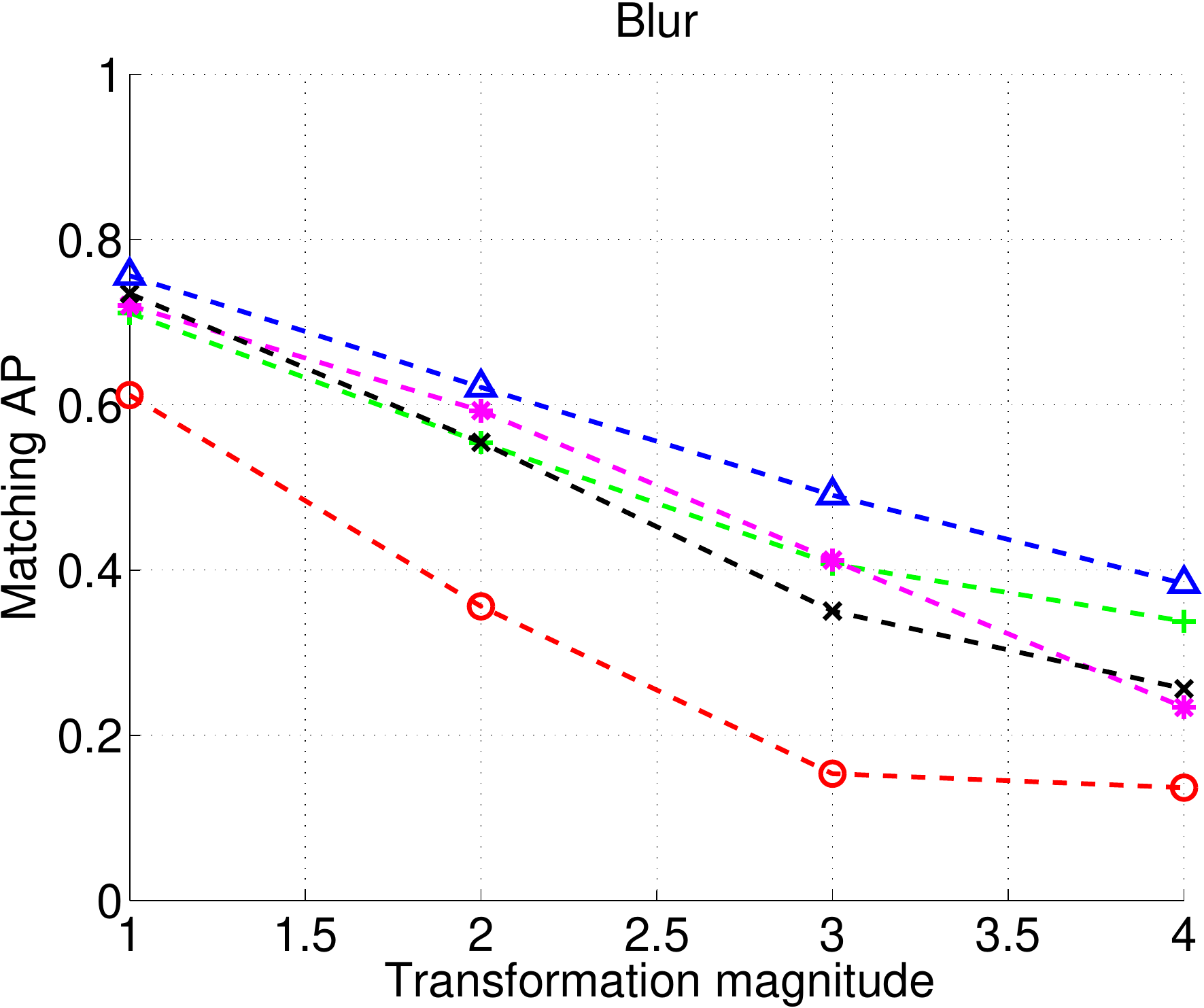}}\vspace{-.2cm}
\end{center}
   \caption{{\sl Average Precision for different magnitude of transformations.} The left $9$ panels show (AP) for increasing magnitude of the $8$ transformations in the Oxford dataset \cite{mikolajc03survey}. \jcomment{The mean AP over all pairs with corresponding amount of transformation are shown in the middle of the third row.} The right $6$ panels show the same for Fischer's dataset \cite{fischer2014descriptor}.}
\label{fig-curv}
\end{figure*}

\begin{figure*}[t]
\begin{center}
\subfigure{\includegraphics[width=.19\textwidth]{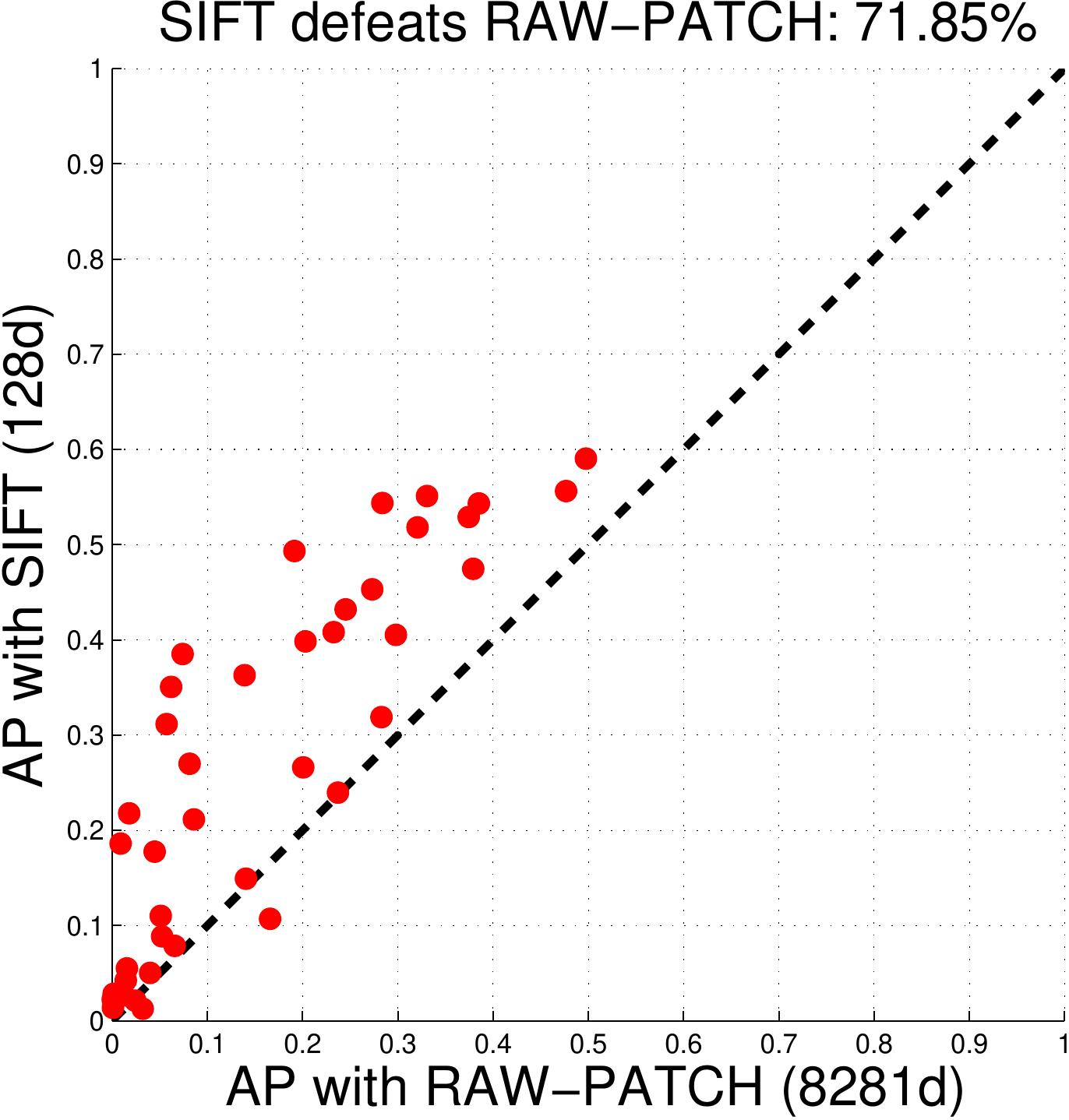}}
\subfigure{\includegraphics[width=.19\textwidth]{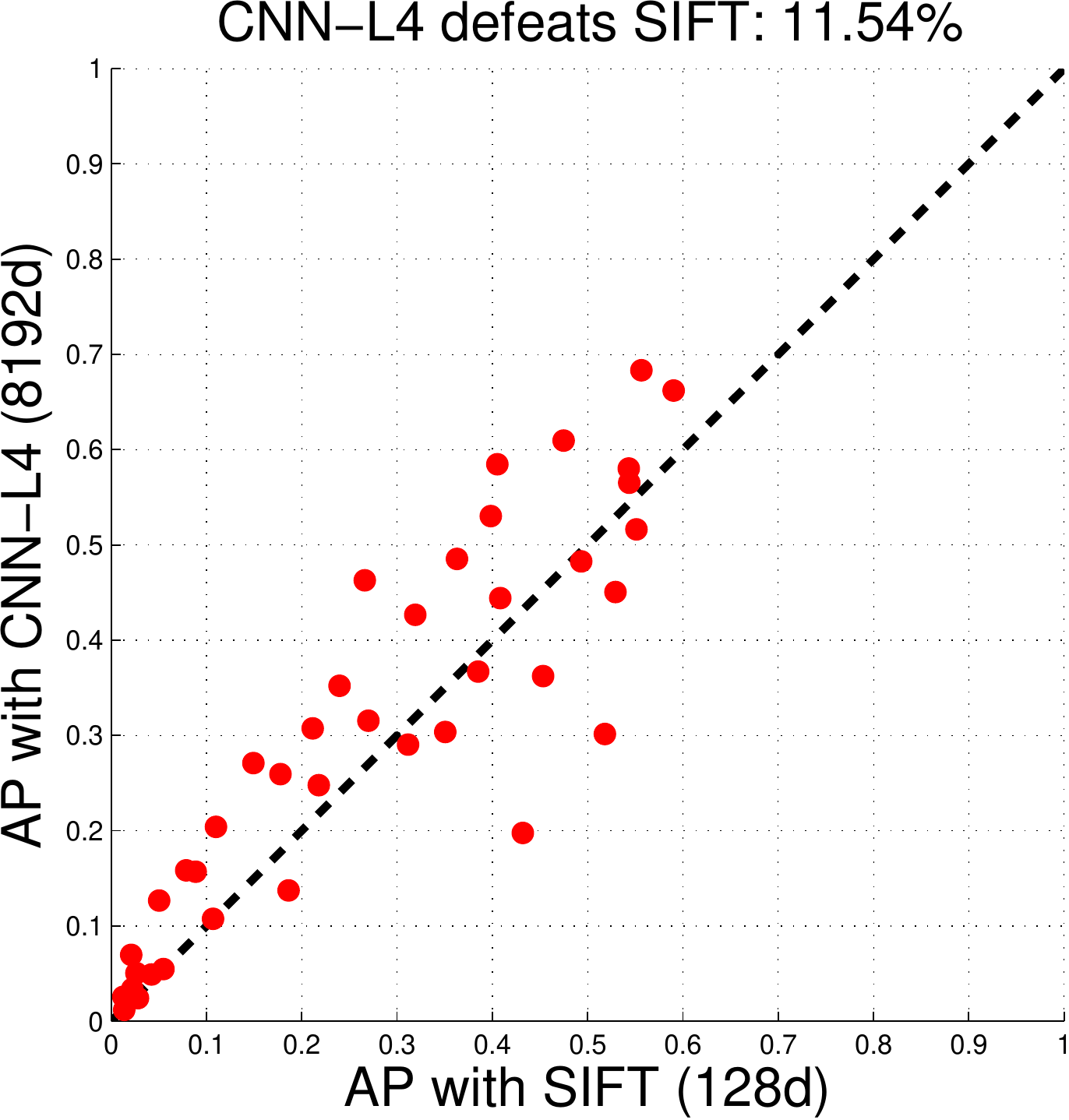}}
\subfigure{\includegraphics[width=.19\textwidth]{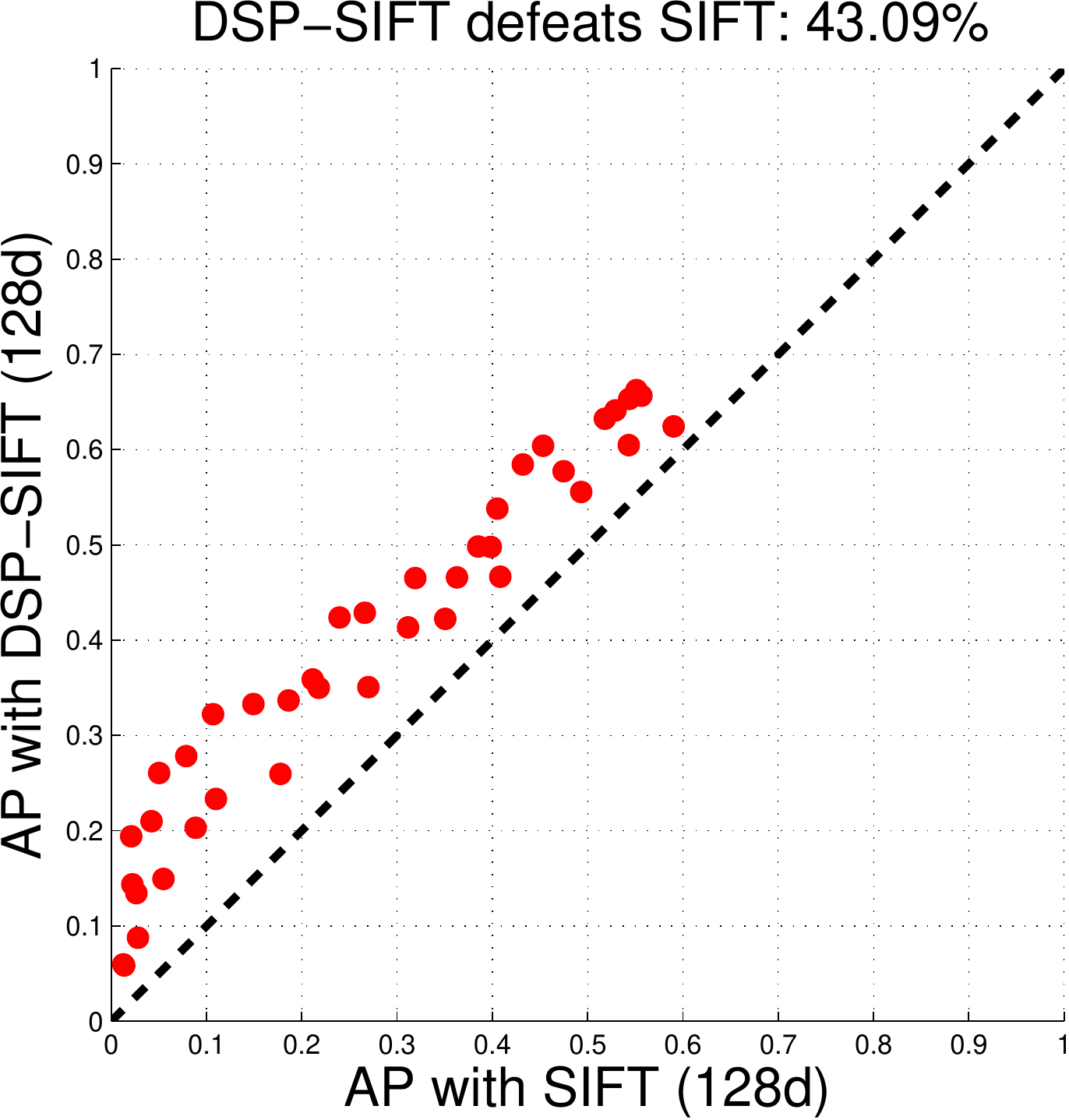}}
\subfigure{\includegraphics[width=.19\textwidth]{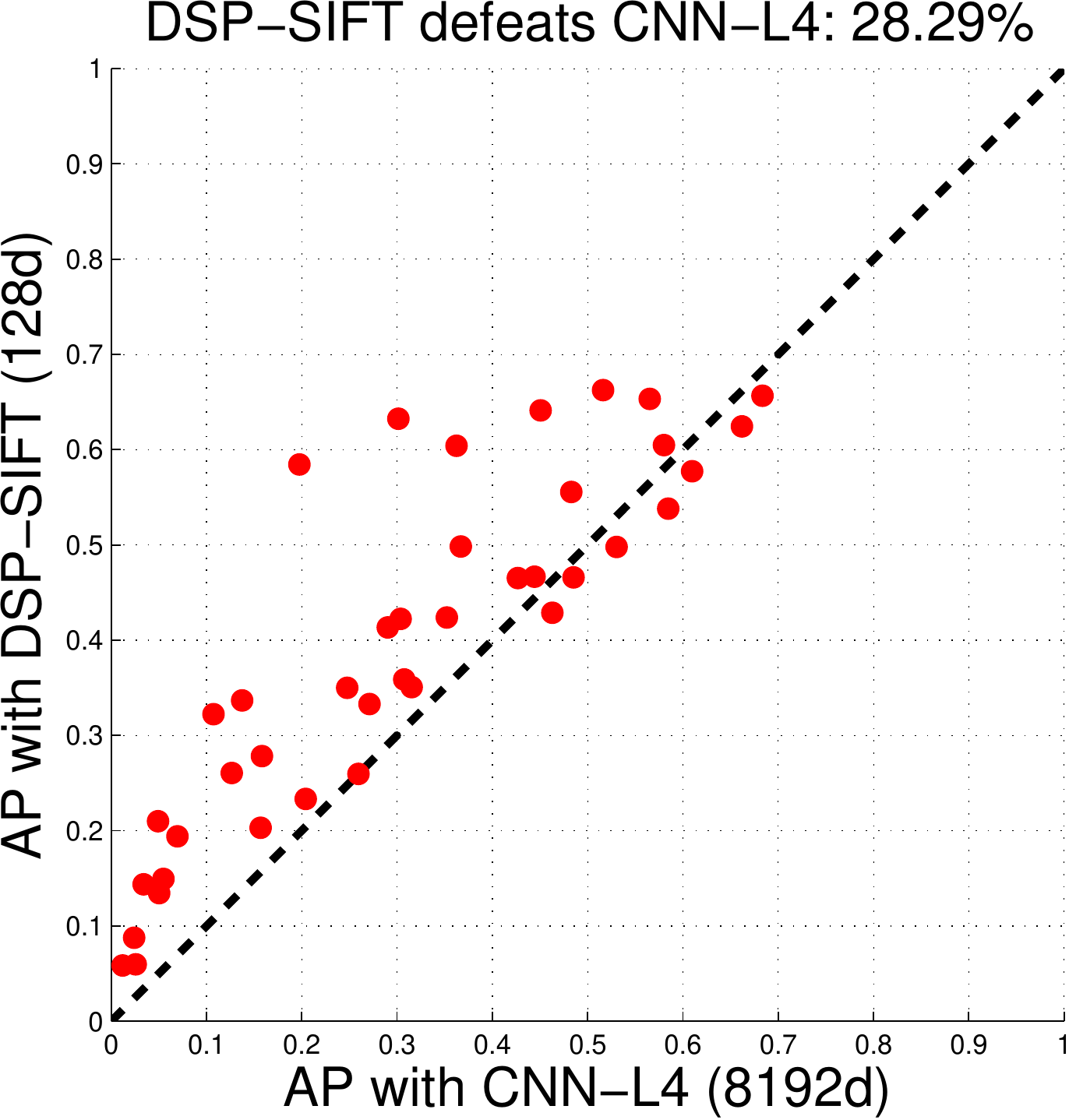}}
\subfigure{\includegraphics[width=.19\textwidth]{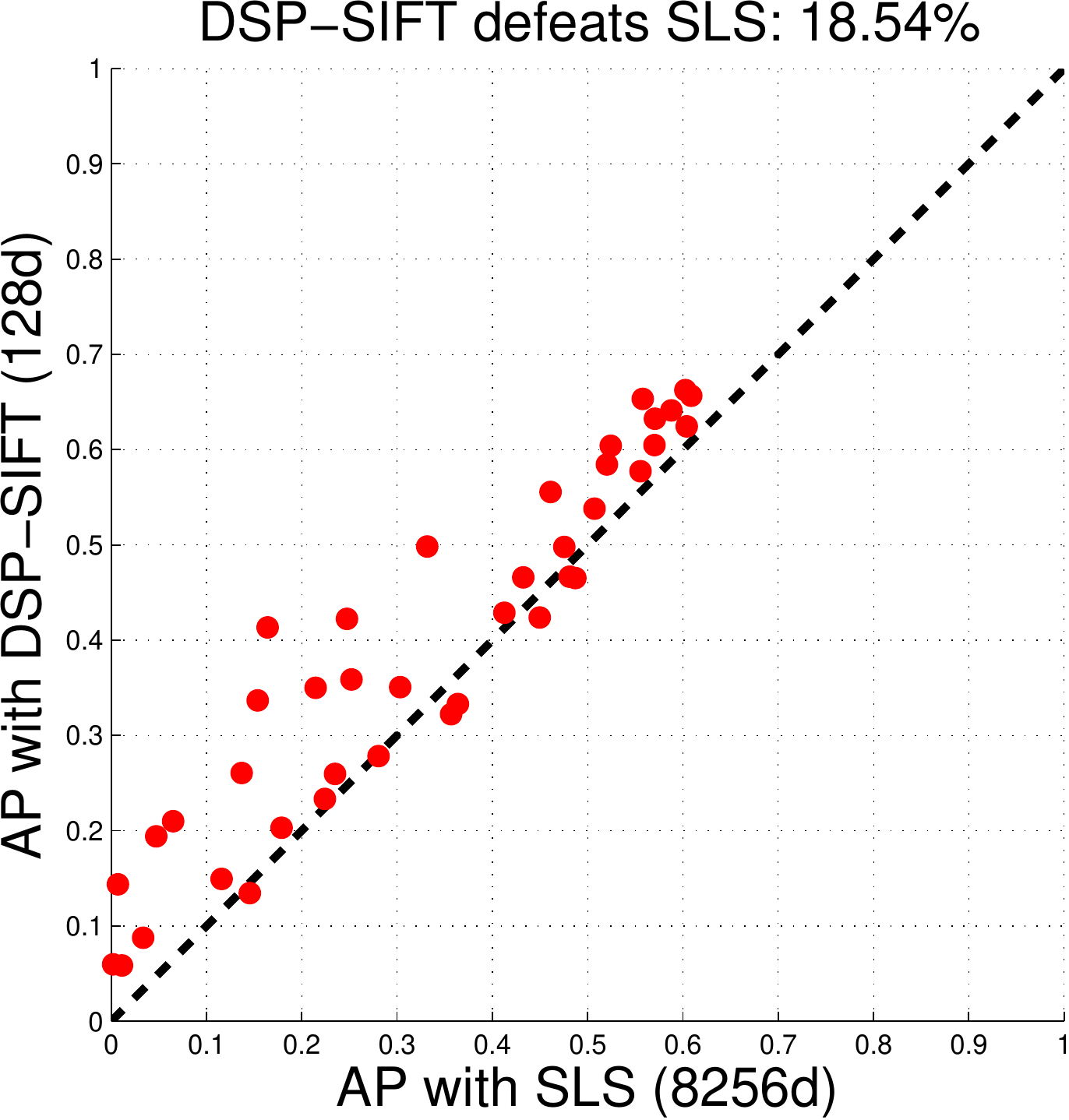}}\vspace{-.3cm}
\subfigure{\includegraphics[width=.19\textwidth]{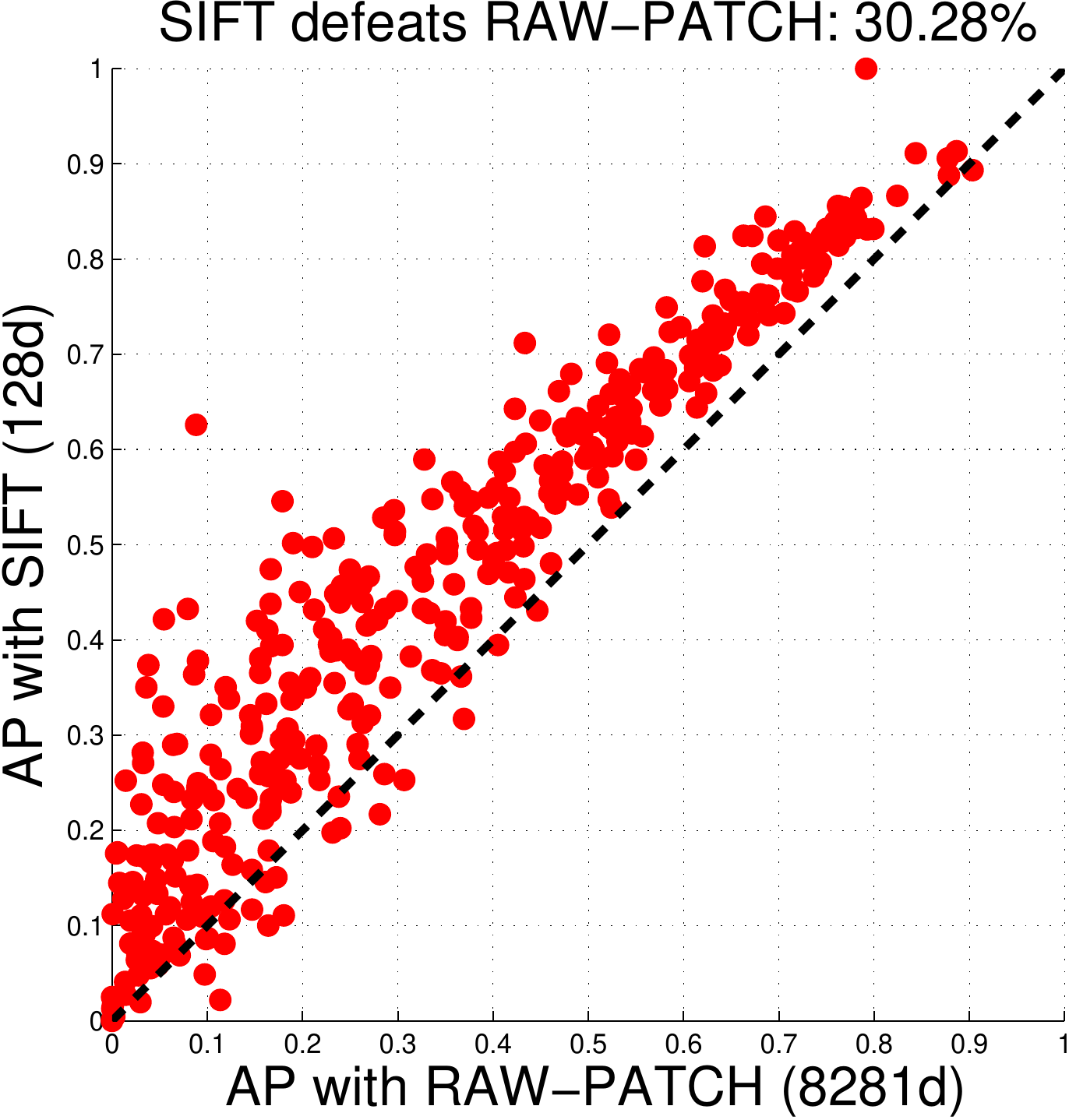}}
\subfigure{\includegraphics[width=.19\textwidth]{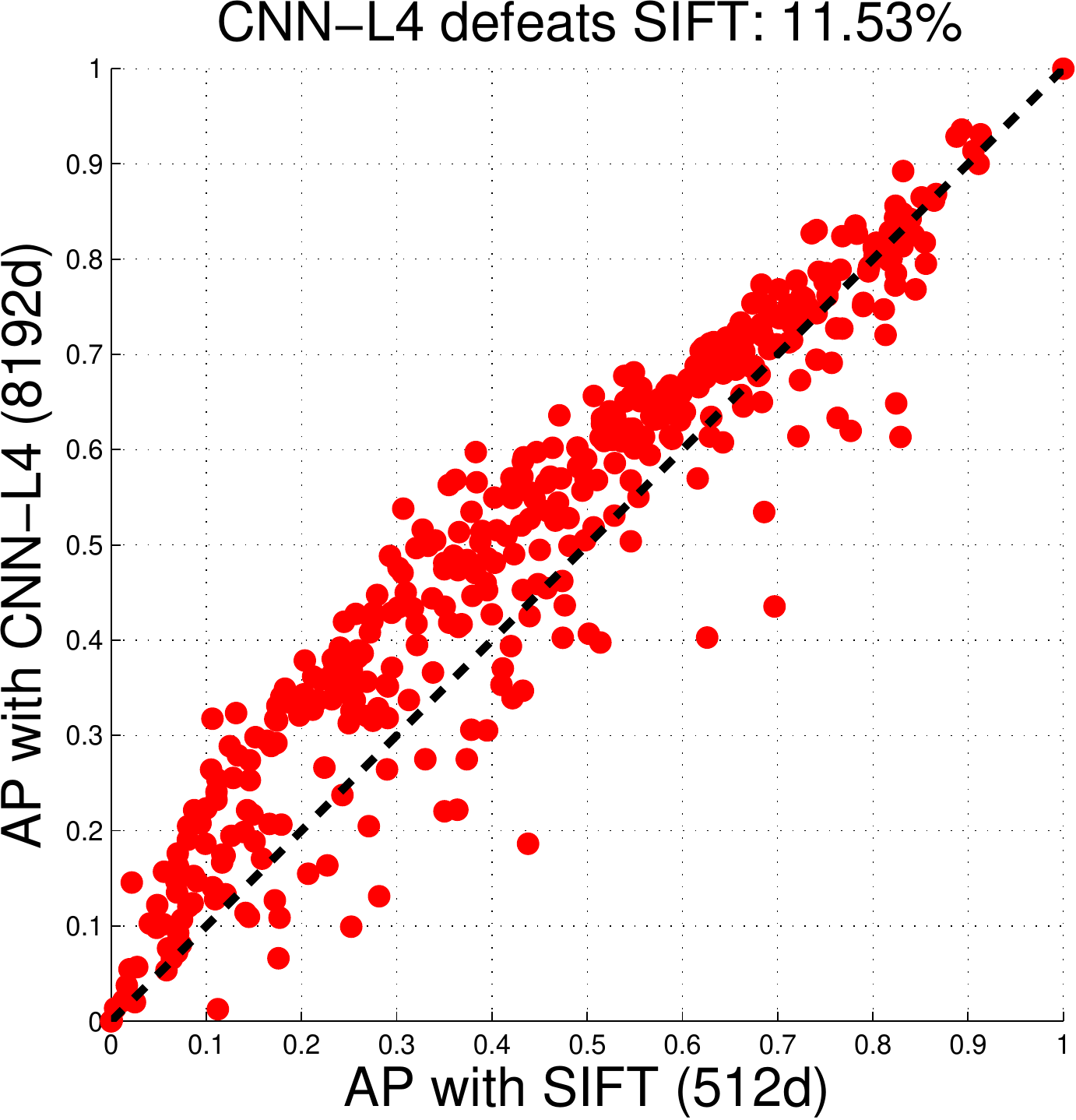}}
\subfigure{\includegraphics[width=.19\textwidth]{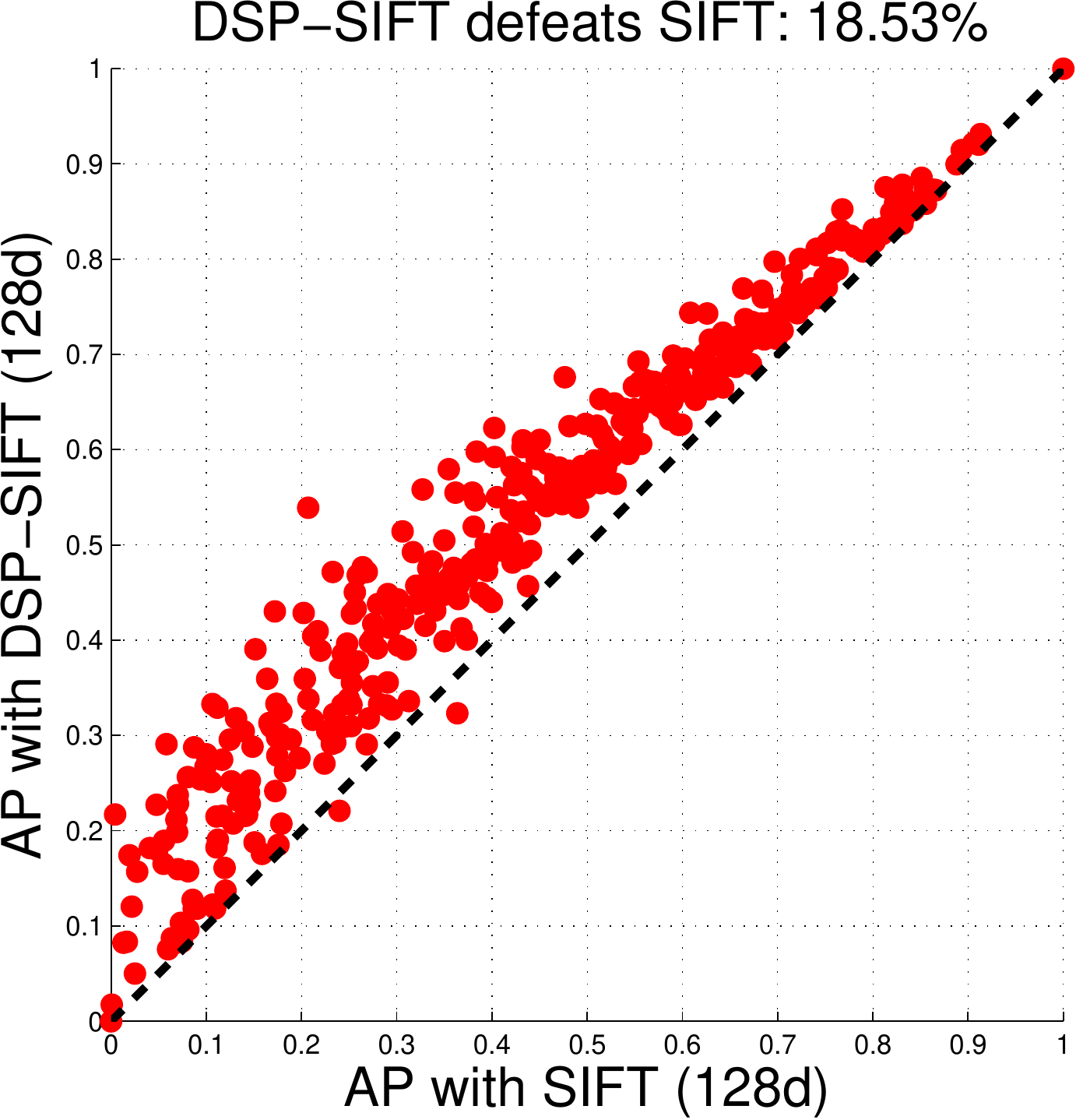}}
\subfigure{\includegraphics[width=.19\textwidth]{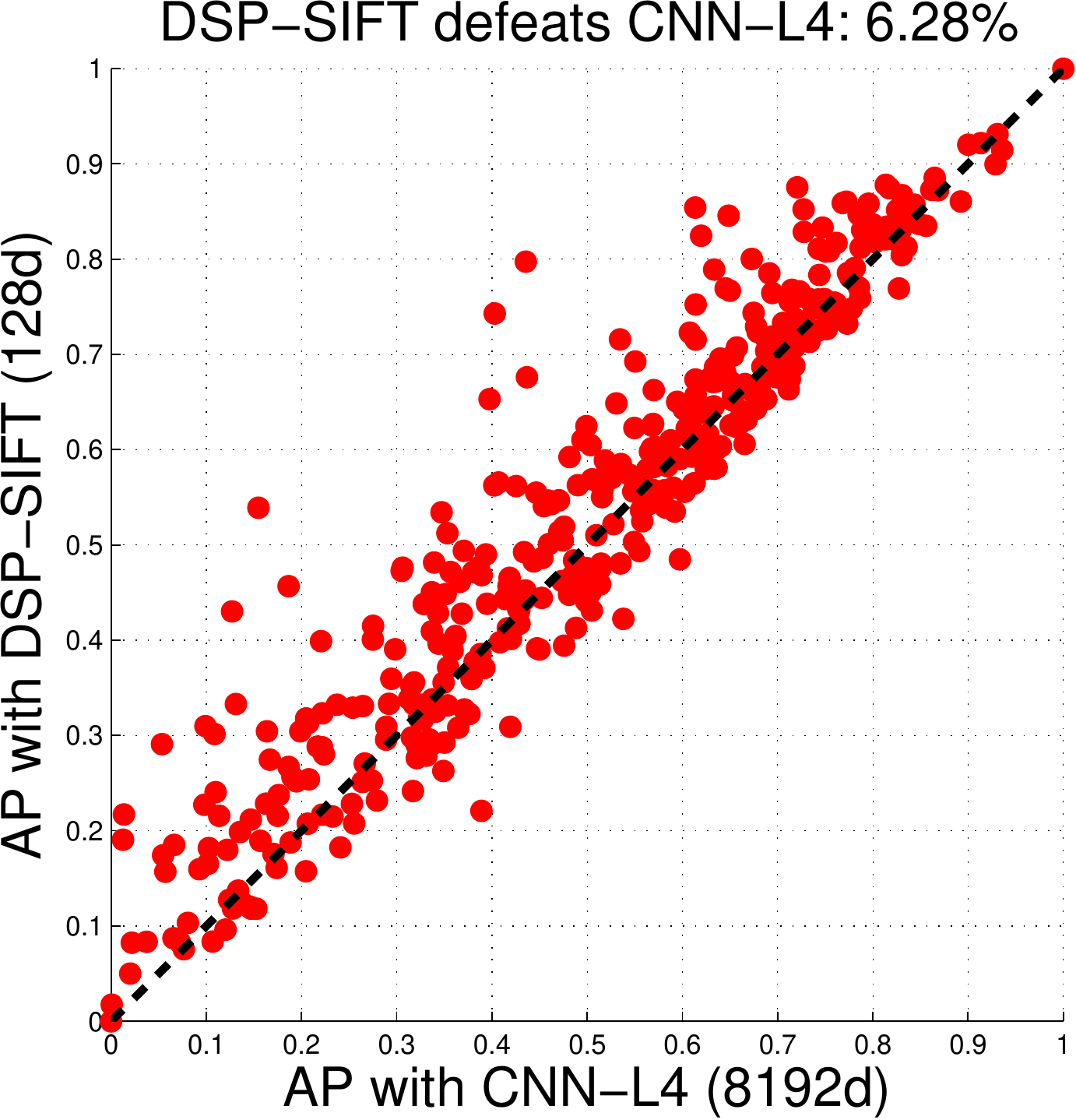}}
\subfigure{\includegraphics[width=.19\textwidth]{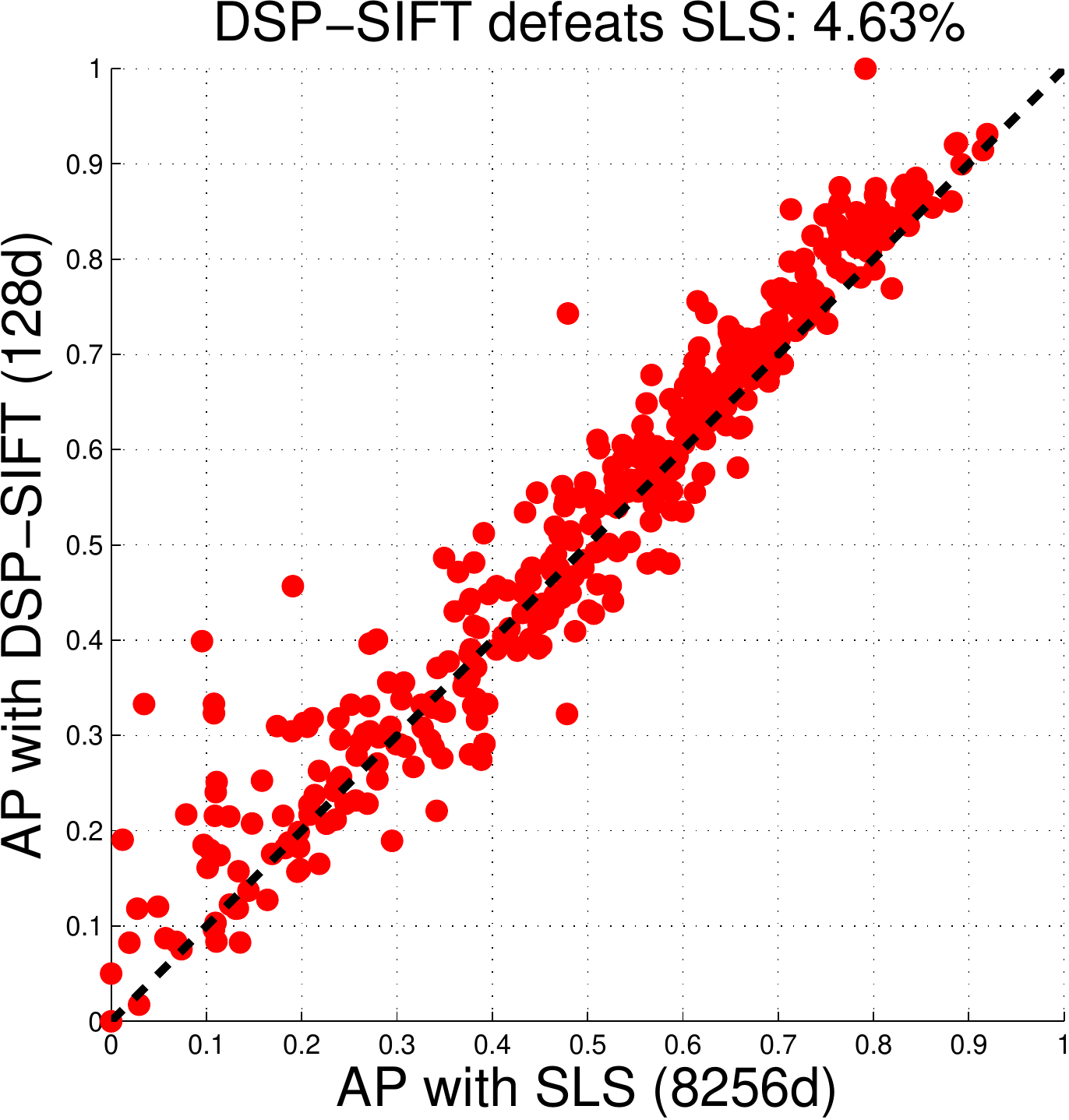}}\vspace{-.2cm}
\end{center}
   \caption{{\sl Head-to-head comparisons.} Similarly to \cite{fischer2014descriptor}, each point represents one pair of images in the Oxford (top) and Fischer (bottom) datasets. The coordinates indicate average precision for each of the two methods under comparison. SIFT is superior to RAW-PATCH, but is outperformed by DSP-SIFT and CNN-L4. The right \newcomment{two} columns show that DSP-SIFT is better than SLS and CNN-L4 despite the difference in dimensions (shown in the axes). The relative performance improvement of the winner is shown in the title of each panel.}
\label{fig-scat}
\end{figure*}

\subsection{Comparison}

Fig.~\ref{fig-curv} shows the behavior of each descriptor for varying degree of severity of each transformation. DSP-SIFT consistently outperforms other methods when there are large scale changes (zoom). It is also more robust to other transformations such as blur, lighting and compression in the Oxford dataset \cite{mikolajczyk04comparison}, and to nonlinear, perspective, lighting, blur and rotation in Fischer's \cite{fischer2014descriptor}. DSP-SIFT is not at the top of the list of all compared descriptors in viewpoint change cases, although ``viewpoint'' is a misnomer as MSER-based rectification accounts for most of the viewpoint variability, and the residual variability is mostly due to interpolation and rectification artifacts. The fact that DSP-SIFT outperforms CNN in nearly all cases in Fischer's dataset is surprising, considering that the neural network is trained by augmenting the dataset using similar types of transformations. 

\jcomment{
Fig.~\ref{fig-scat} shows head-to-head comparisons between these methods, in the same format of \cite{fischer2014descriptor}. 
DSP-SIFT outperforms SIFT by \jcomment{$43.09\%$ and $18.54\%$} on Oxford and Fischer respectively. Only on two out of $400$ pairs of images in Fischer dataset does domain-size pooling negatively affect the performance of SIFT, but the decrease is rather small. DSP-SIFT improves SIFT on every pair of images in the Oxford dataset. The improvement of DSP-SIFT comes without increase in dimension. In comparison, CNN-L4 \jcomment{achieves $11.54\%$ and $11.53\%$ improvements over SIFT} by increasing dimension $64$-fold. On both datasets, DSP-SIFT also consistently outperforms CNN-L4 and SLS despite its lower dimension.
}

\begin{figure}[t]
\begin{center}
\subfigure{\includegraphics[width=.49\columnwidth]{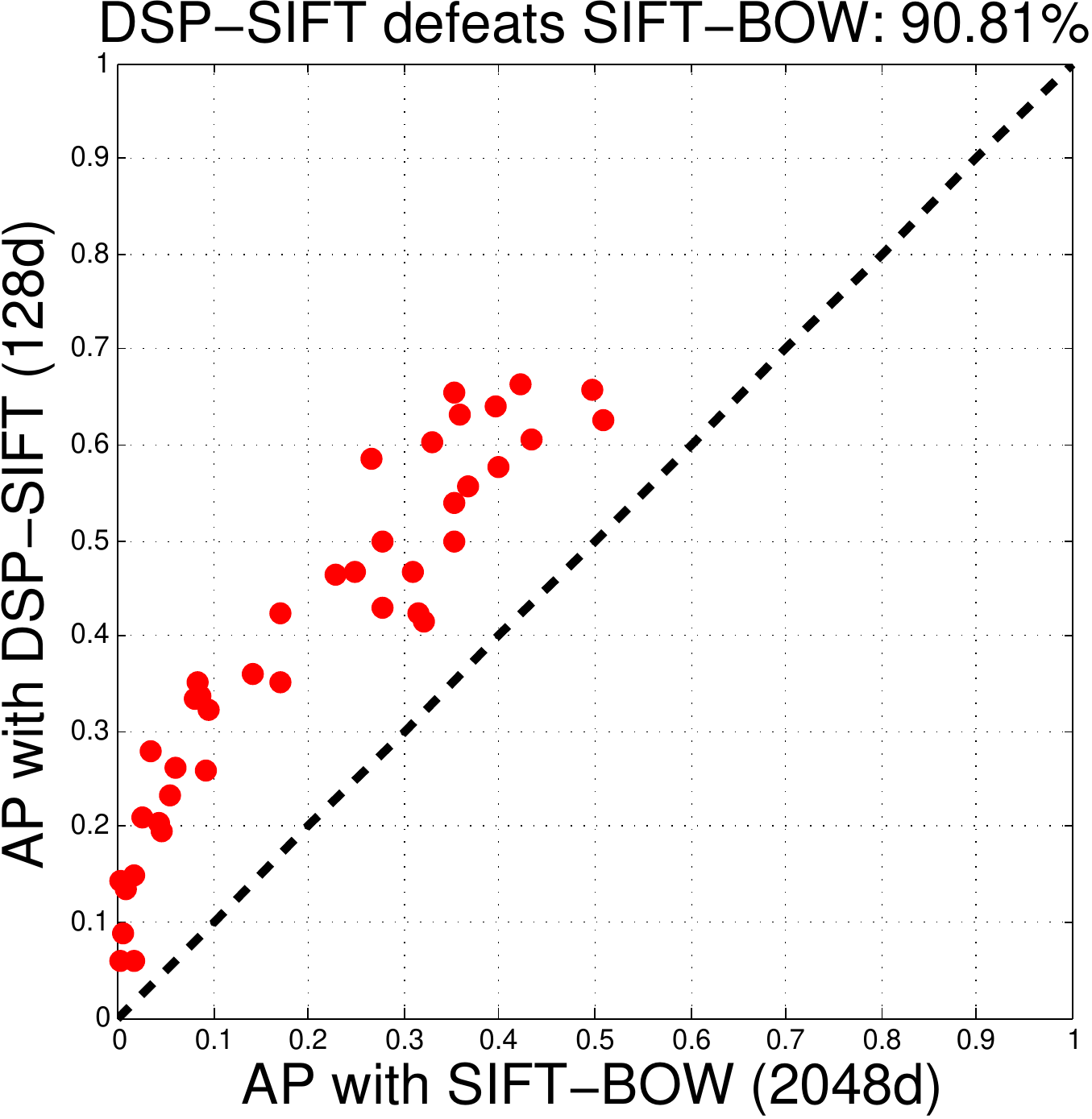}}
\subfigure{\includegraphics[width=.49\columnwidth]{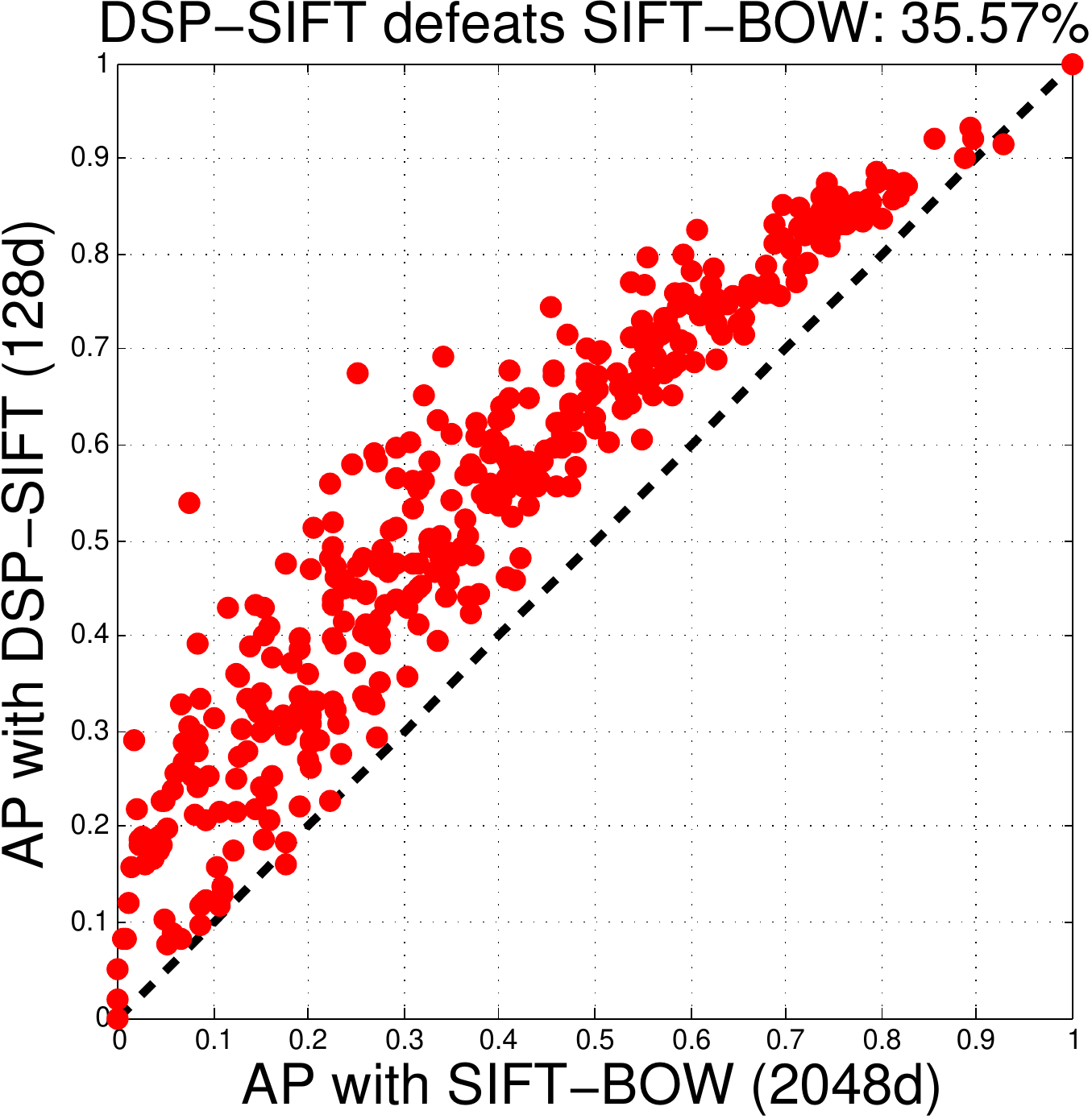}}\vspace{-.2cm}
   \end{center}
   \caption{{\sl DSP-SIFT vs. SIFT-BOW.} Similarly to Fig.~\ref{fig-scat}, each point represents one pair of images in the Oxford (left) and Fischer (right) datasets. The coordinates indicate average precision for each of the two methods under comparison. The relative performance improvement of the winner is shown in the title of each panel. DSP-SIFT outperforms SIFT-BOW by a wide margin on both datasets.}
\label{fig-bow}
\end{figure}

\jcomment{
\subsection{Comparison with Bag-of-Words}
\label{sect-bow}
To compare DSP-SIFT to BoW we computed SIFT at $15$ scales on concentric regions with dictionary sizes ranging from $512$ to $2048$, trained on over $100$K SIFT descriptors computed on samples from ILSVRC-2013 \cite{imagenet_cvpr09}. To make the comparison fair, the same $15$ scales are used to compute DSP-SIFT. By doing so, the only difference between these two methods is {\em how} to pool across scales rather than {\em what} or {\em where} to pool. In SIFT-BOW, pooling is performed by encoding SIFTs from nearby scales using the quantized visual dictionary, while DSP-SIFT combines the histograms of gradient orientations across scales directly. To compute similarity between SIFT-BOWs, we tested both the intersection kernel and $\ell_1$ norm, and achieved a best performance with the latter at $20.62\%$ mAP on Oxford and $39.63\%$ on Fischer. Fig.~\ref{fig-bow} shows the direct comparison between DSP-SIFT and SIFT-BOW with the former being a clear winner.
}

\subsection{Complexity and Performance Tradeoff}

\begin{figure}[t]
\begin{center}
\vspace{.2cm}
{\includegraphics[width=.99\columnwidth]{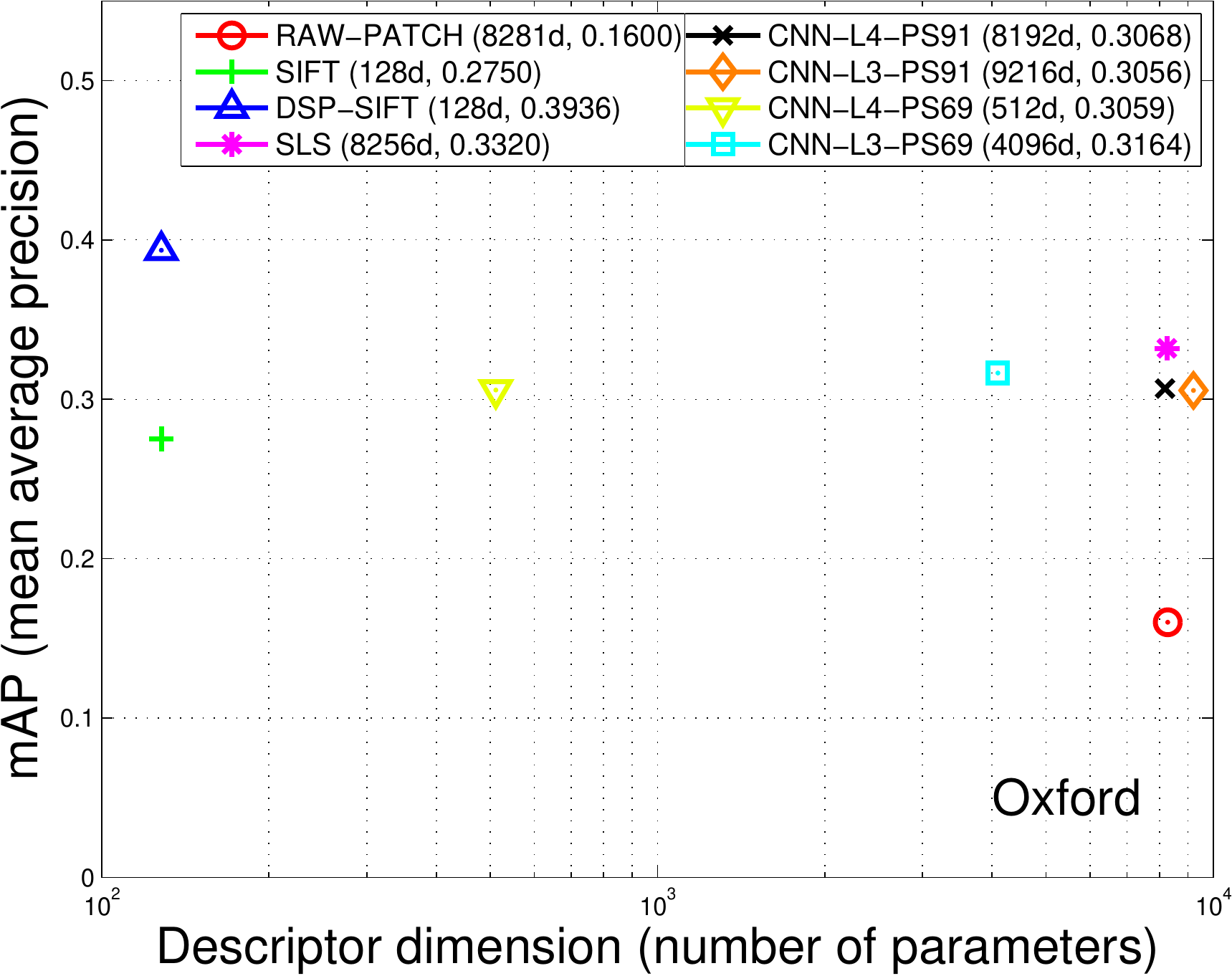}}\vspace{.2cm}
{\includegraphics[width=.99\columnwidth]{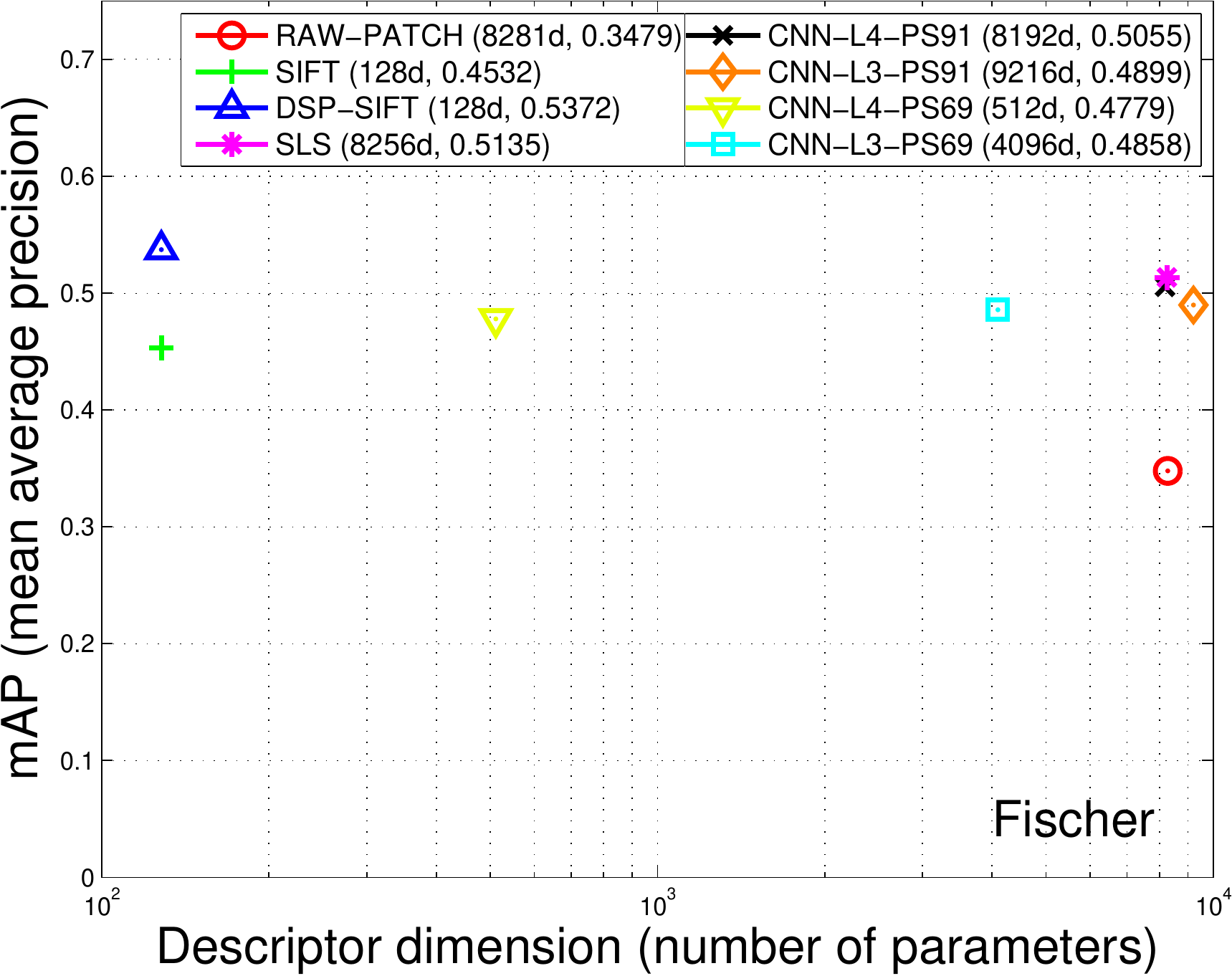}}\vspace{-.2cm}
\end{center}
   \caption{{\sl Complexity-Performance Tradeoff.} The abscissa is the descriptor dimension shown in log-scale, the ordinate shows the mean average precision. 
}
\label{fig-tradeoff}
\end{figure}

Fig.~\ref{fig-tradeoff} shows the complexity (descriptor dimension) and performance (mAP) tradeoff. Table \ref{tbl-summ} summarizes the results. In Fig.~\ref{fig-tradeoff}, an ``ideal'' descriptor would achieve mAP $=1$ by using the smallest possible number of bits and land at the top-left corner of the graph. DSP-SIFT has the same lowest complexity as SIFT and is the best in mAP among all the descriptors. Looking horizontally in the graph, DSP-SIFT outperforms all the other methods at a fraction of complexity. SLS achieves the second best performance but at the cost of a $64$-fold increase in dimension. In general, the performance of CNN descriptors is worse than DSP-SIFT but, interestingly, their mAPs do not change significantly if the network responses are computed on a resampled patch of size $69\times 69$ to obtain lower dimensional descriptors.

\subsection{Comparison with SIFT on Larger Domain Sizes}

\begin{figure}[t]
\begin{center}
\subfigure{\label{fig-large-a}\includegraphics[width=.49\columnwidth]{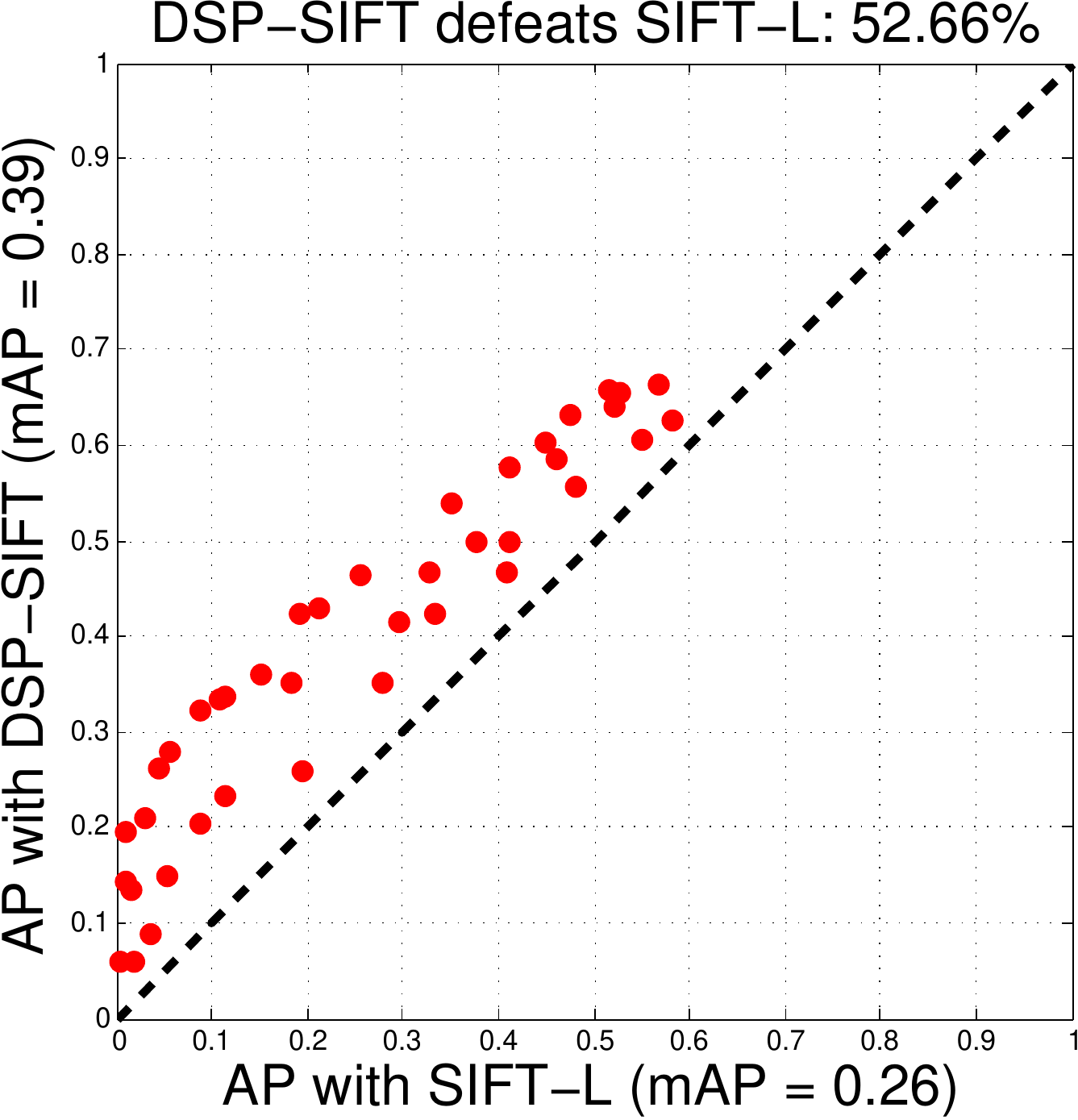}}
\subfigure{\label{fig-large-b}\includegraphics[width=.49\columnwidth]{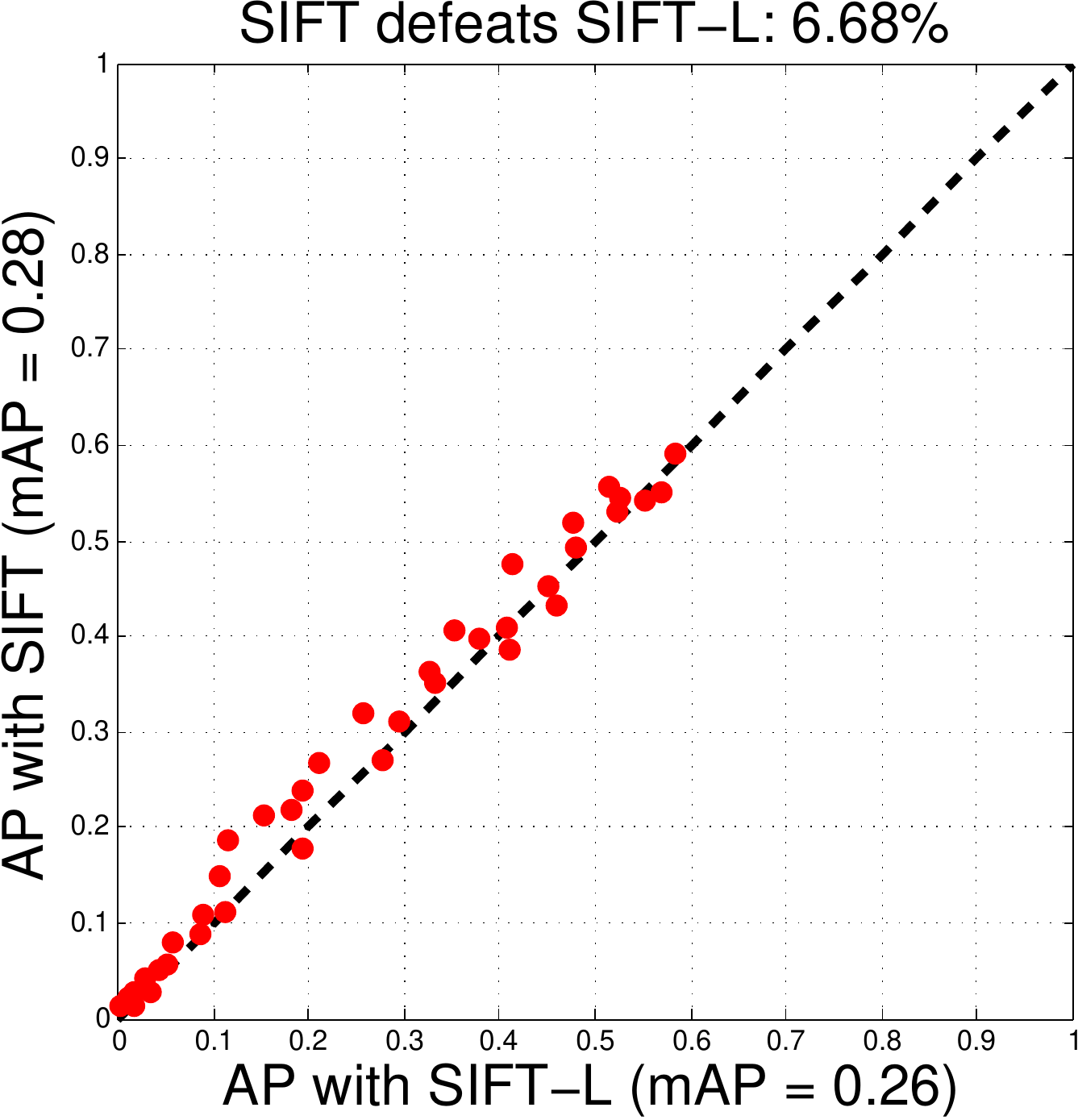}}\vspace{-.2cm}
\end{center}
   \caption{{\sl DSP-SIFT vs. SIFT-L.} Similarly to Fig.~\ref{fig-scat}, each point represents one pair of images in the Oxford dataset. The coordinates indicate average precision for each of the two methods under comparison. The relative performance improvement of the winner is shown in the title of each panel. \ref{fig-large-a} shows that DSP-SIFT outperforms SIFT computed at the largest domain size. This shows that the improvement of DSP-SIFT comes from the pooling across domain sizes rather than choosing a larger domain size. \ref{fig-large-b} shows that choosing a larger domain size actually decreases the performance on the Oxford dataset.}
\label{fig-large}
\end{figure}

Descriptors computed on larger domain sizes are usually more discriminative, up to the point where the domain straddles occluding boundaries (Fig.~\ref{fig-performance-vs-base-size}). When using a detector, the size of the domain is usually chosen to be a factor of the detected scale, which affects performance in a way that depends on the dataset and the incidence of occlusions. In our experiments, this parameter (dilation factor) is set at 3, following \cite{mikolajc03survey}, and we note that DSP-SIFT is less sensitive than ordinary SIFT to this parameter. Since DSP-SIFT aggregates domains of various sizes (smaller and larger) around the nominal size, it is important to ascertain whether the improvement in DSP-SIFT comes from size pooling, or simply from including larger domains. To this end, we compare DSP-SIFT by pooling domain sizes from $1/6$th through $4/3$rd of the scale determined by the detector, to a single-size descriptor computed at the largest size (SIFT-L). This establishes that the increase in performance of DSP-SIFT over ordinary SIFT comes from pooling across domain sizes, not just by picking larger domain sizes. In the example in Fig.~\ref{fig-large}, the largest domain size yields an even worse performance than the detection scale (Fig.~\ref{fig-large-b}). In a more complex scene where the test images exhibit occlusion, this will be even more pronounced as there is a tradeoff between discriminative power (calling for a larger size) and the probability of straddling an occlusion (calling for a smaller size).

\begin{table}[h]
\begin{center}
{\small{
\begin{tabular}{|l|c|c|c|}
\hline
\multirow{2}{*}{Method} & \multirow{2}{*}{Dim.} & \multicolumn{2}{c|}{mAP} \\ 
			&			& ~~~Oxford~~~  & ~~~Fischer~~~ \\
\hline
SIFT 		& {\bf128}	& .2750 		& .4532 		\\
DSP-SIFT 	& {\bf128} 	& {\bf.3936}	& {\bf.5372}	\\
CNN-L4-PS69 & 512 		& .3059 		& .4779 		\\
SIFT-BOW	& 2048		& .2062			& .3963 		\\
CNN-L3-PS69 & 4096 		& .3164 		& .4858 		\\
CNN-L4-PS91 & 8192 		& .3068 		& .5055 		\\
SLS 		& 8256 		& .3320 		& .5135 		\\
RAW-PATCH 	& 8281 		& .1600 		& .3479 		\\
CNN-L3-PS91 & 9216 		& .3056 		& .4899 		\\
\hline
\end{tabular}}}
\end{center}
\caption{{\sl Summary of complexity (dimension) and performance (mAP) for all descriptors} sorted in order of increasing complexity. The lowest complexities and the best performances are highlighted in bold. We also report mAP for CNN descriptors computed on $69\times69$ patches as in \cite{fischer2014descriptor}. The fourth row shows comparison with a bag-of-words of SIFT descriptors computed at the same location but different domain sizes, described in detail in Sect.~\ref{sect-bow}.}
\label{tbl-summ}
\end{table}

\section{Derivation} 
\label{sect-why}

In this section we \jcomment{describe the} trace of the derivation of DSP-SIFT, which is \jcomment{reported in Appendix \ref{sect-derivation}.} Crucial to the derivation is the interpretation of a descriptor as a likelihood function \cite{soattoC14ICLR}.

\noindent {1.}~~The likelihood function of the scene given images is a minimal sufficient statistic of the latter for the purpose of answering questions on the former \cite{bahadur1954sufficiency}. Invariance to nuisance transformations induced by (semi-)group actions on the data can be achieved by representing orbits, which are maximal invariants \cite{shao98}. The planar translation-scale group can be used as a crude first-order approximation of the action of the translation group in space (viewpoint changes) including scale change-inducing translations along the optical axis. This draconian assumption is implicit in most single-view descriptors.

\noindent {2.}~~Comparing (semi-)orbits entails a continuous search (non-convex optimization) that has to be discretized for implementation purposes. The orbits can be sampled adaptively, through the use of a co-variant detector and the associated invariant descriptor, or regularly -- as customary in classical sampling theory. 

\noindent {3.}~~In adaptive sampling, the {\em detector} should exhibit high sensitivity to nuisance transformations (\eg~small changes in scale should cause a large change in the response to the detector, thus providing accurate scale localization) and the {\em descriptor} should exhibit small sensitivity (so small errors in scale localization cause a small change in the descriptor). Unfortunately, for the case of SIFT (DoG detector and gradient orientation histogram descriptor), the converse is true. 

\noindent {4.}~~Because correspondence entails search over samples of each orbit,  time complexity increases with the number of samples. Undersampling introduces structural artifacts, or ``aliases,'' corresponding to topological changes in the response of the detector. 
These can be reduced by ``anti-aliasing,'' an averaging operation. 
For the case of (approximations of) the likelihood function, such as SIFT and its variants, anti-aliasing corresponds to {\em pooling}. While spatial pooling is common practice, and reduces sensitivity to translation parallel to the image plane, scale pooling -- which would provide insensitivity to translation orthogonal to the image plane -- and domain-\newcomment{size} pooling -- which would provide insensitivity to small changes of visibility, are not. This motivates the introduction of DSP-SIFT, and the rich theory on sampling and anti-aliasing could provide guidelines on what and how to pool, as well as bounds on the loss of discriminative power coming from undersampling and anti-aliasing operations.

\begin{figure}[tb]
\begin{center}
\includegraphics[width=.25\textwidth]{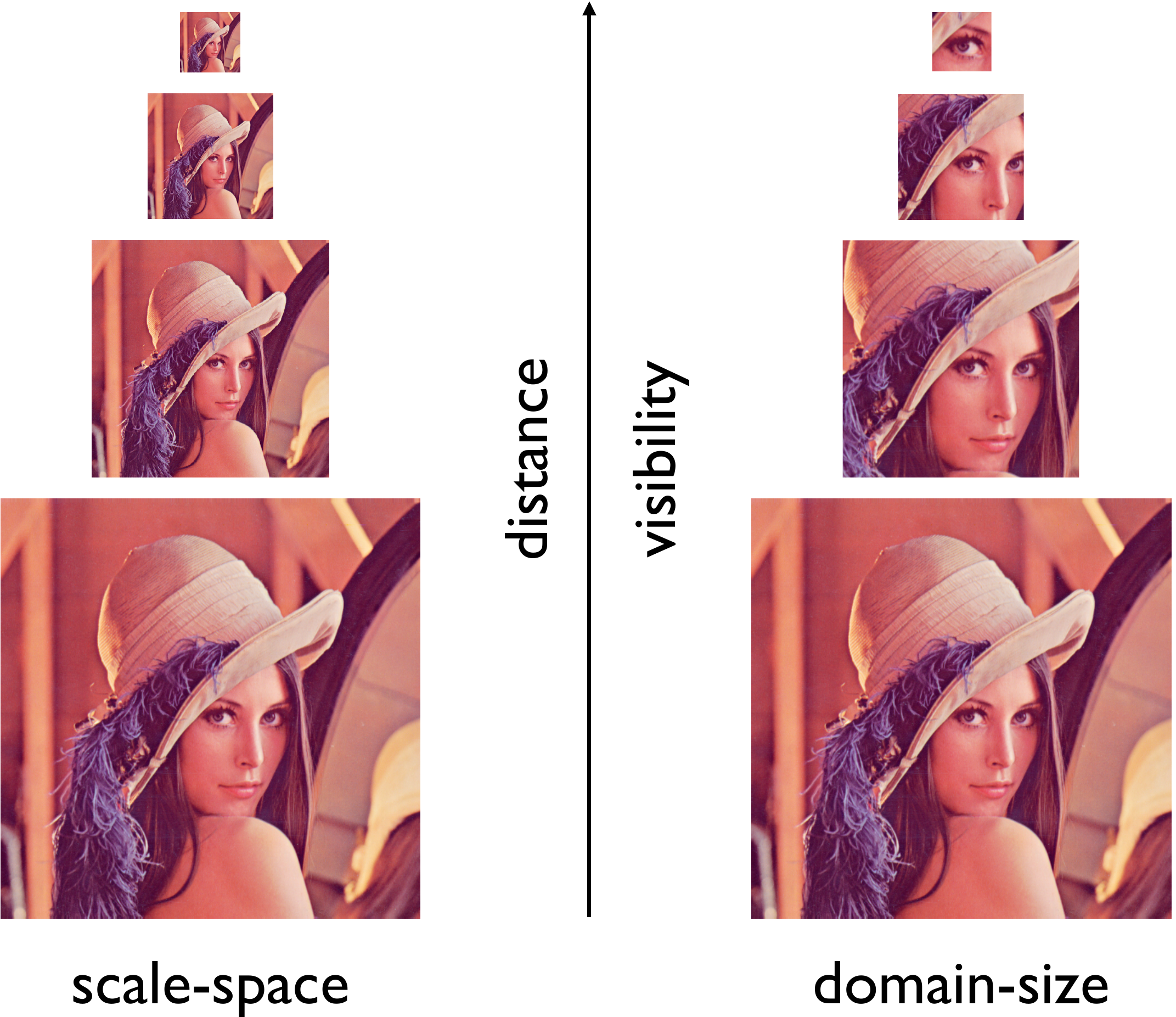} ~~~~\includegraphics[width=.16\textwidth]{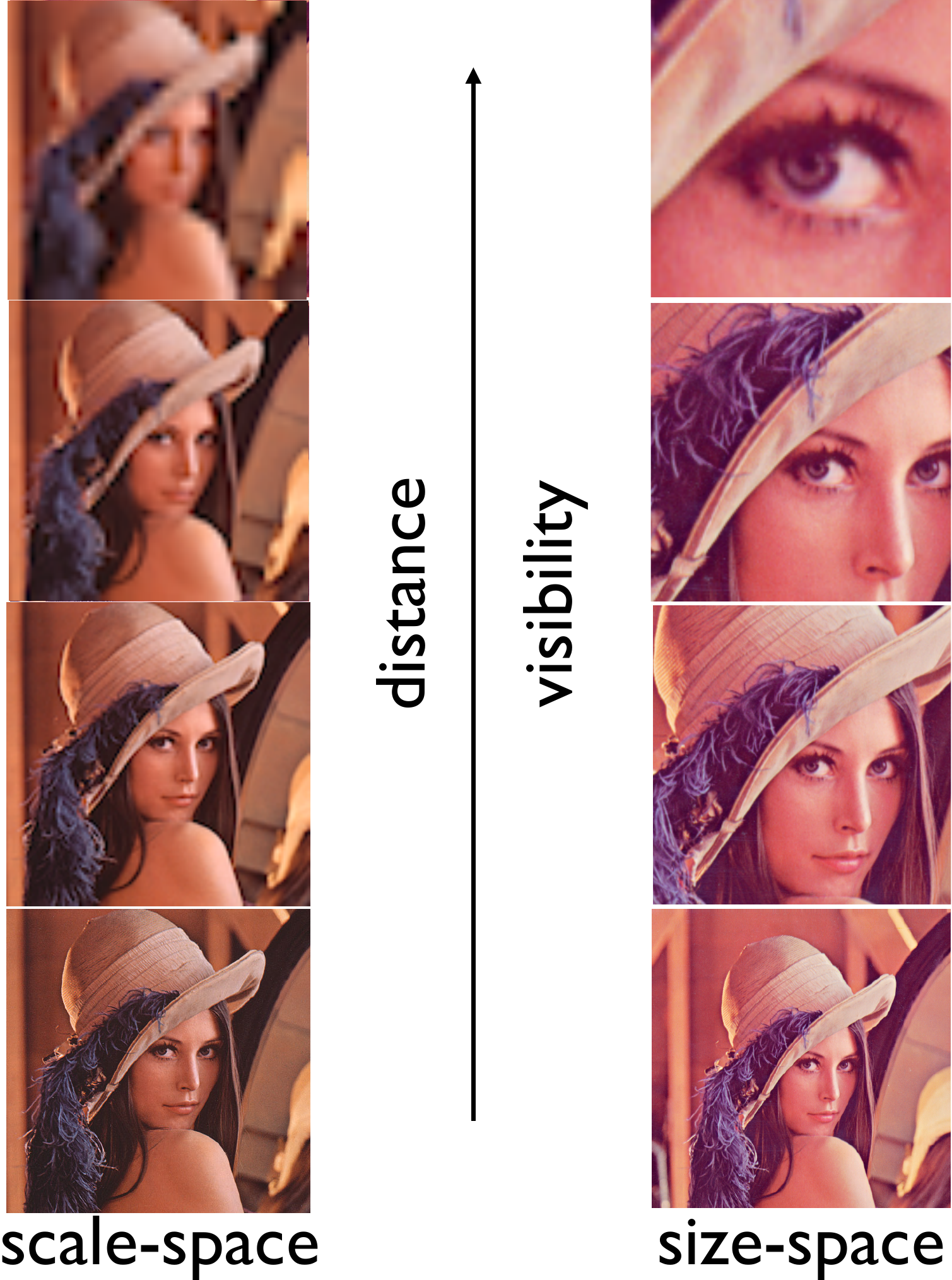}
\end{center}
\caption{{\sl Scale-space vs. Size-space.} Scale-space refers to a continuum of images obtained by smoothing and downsampling a base image. It is relevant to searching for correspondence when the distance to the scene changes. Size-space refers to a scale-space obtained by maintaining the same scale of the base image, but considering subsets of it of variable size. It is relevant to searching for correspondence in the presence of occlusions, so the size (and shape) of co-visible domains are not known.} 
\label{fig-scale-size}
\end{figure}

\section{Discussion}
\label{sect-discussion}

Image matching under changes of viewpoint, illumination and partial occlusions is framed as a hypothesis testing problem, which results in a
non-convex optimization over continuous nuisance parameters. The need for efficient test-time performance has spawned an industry of engineered descriptors, which are computed locally so the effects of occlusions can be reduced to a binary classification (co-visible, or not). The best known is SIFT, which has been shown to work well in a number of independent empirical assessments \cite{mikolajc03survey,moreels2007evaluation}, that however come with little analysis on {\em why} it works, or indications on how to improve it. We have made a step in that direction, by showing that SIFT can be derived from sampling considerations, where spatial binning and pooling are the result of anti-aliasing operations. However, SIFT and its variants only perform such operations for planar translations, whereas our interpretation calls for anti-aliasing domain-size as well. Doing so can be accomplished in few lines of code and yields significant performance improvements.  Such improvements even place the resulting DSP-SIFT descriptor above a convolutional neural network (CNN), that had been recently reported as a top performer in the Oxford image matching benchmark \cite{fischer2014descriptor}. Of course, we are not advocating replacing large neural networks with local descriptors.
Indeed, there are interesting relations between DSP-SIFT and convolutional architectures, explored in \cite{soattoC14ICLR,soattoDK15}.

\begin{figure}[tb]
\begin{center}
\includegraphics[width=.38\textwidth]{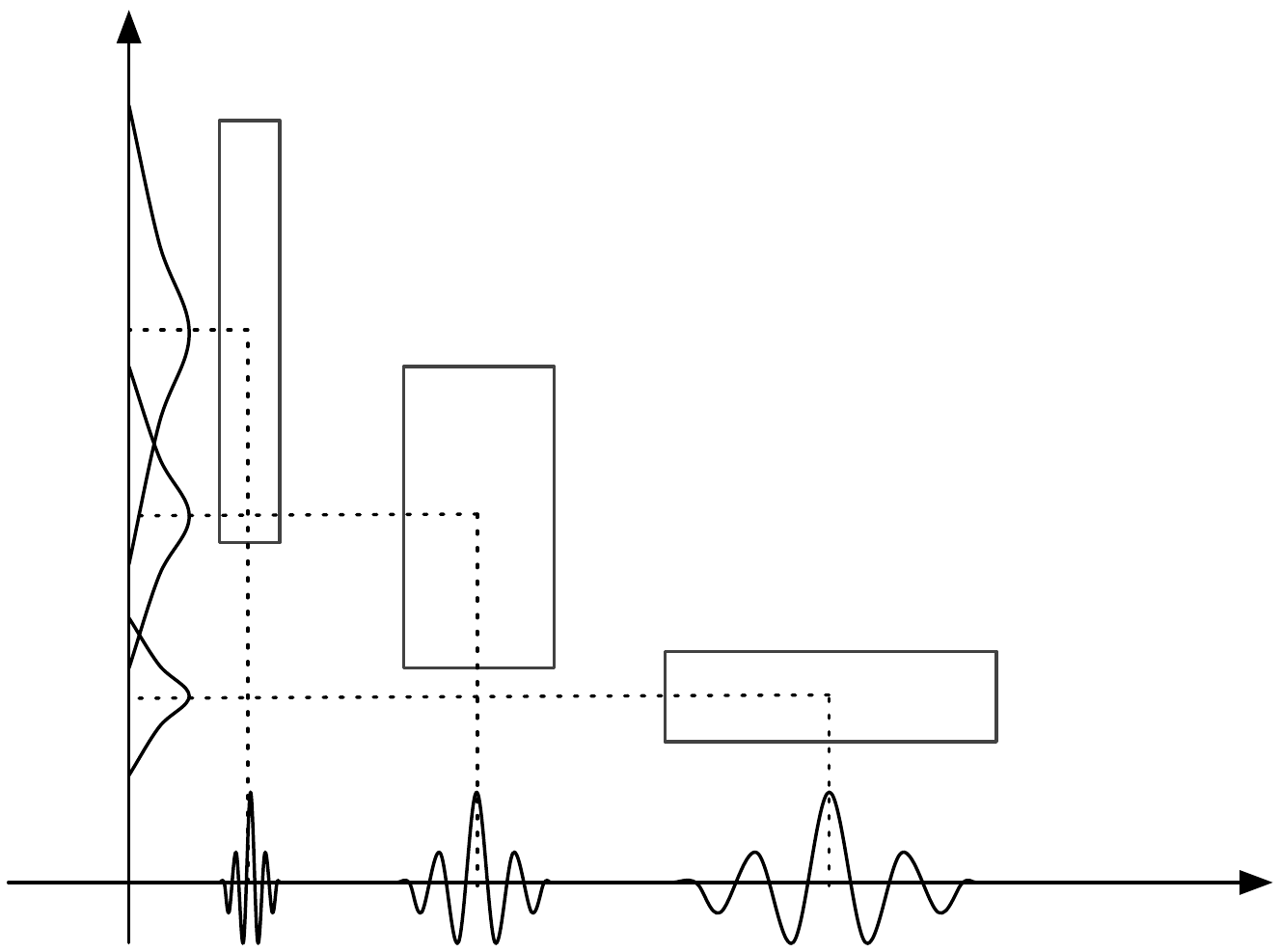}
\end{center}
\caption{The ``uncertainty principle'' links the size of the domain of a filter (ordinate) to its spatial frequency (abscissa): As the data is analyzed for the purpose of compression, regions with high spatial frequency must be modeled at small scale, while regions with smaller spatial frequency can be encoded at large scale. When the task is correspondence, however, the size of the co-visible domain is independent of the spatial frequency of the scene within. While approaches using ``dense SIFT'' forgo the detector and compute descriptors at regularly sampled locations and scales, they perform spatial pooling by virtue of the descriptor, but fail to perform pooling across scales, as we propose.}
\label{fig-unc-pri}
\end{figure}

Domain-size pooling, and regular sampling of scale ``unhinged'' from the spatial frequencies of the signal is divorced from scale selection principles, rooted in scale-space theory, wavelets and harmonic analysis. There, the goal is to reconstruct a signal, with the focus on photometric nuisances (additive noise). In our case, the size of the domain where images correspond depends on the three-dimensional shape of the underlying scene, and visibility (occlusion) relations, and has little to do with the spatial frequencies or ``appearance'' of the scene. Thus, we do away with the linking of domain size and spatial frequency (``uncertainty principle'', Fig.~\ref{fig-unc-pri}).

DSP can be easily extended to other descriptors, such as HOG, SURF, CHOG, including those supported on structured domains such as DPMs \cite{felzenswalb}, and to \jcomment{network architectures such as convolutional neural networks and scattering networks \cite{mallatB11}}, opening the door to multiple extensions of the present work. In addition, a number of interesting open theoretical questions can now be addressed using the tools of classical sampling theory, given the novel interpretation of SIFT and its variants introduced in this paper.

\begin{figure}[htb]
\centering
\includegraphics[width=.42\textwidth]{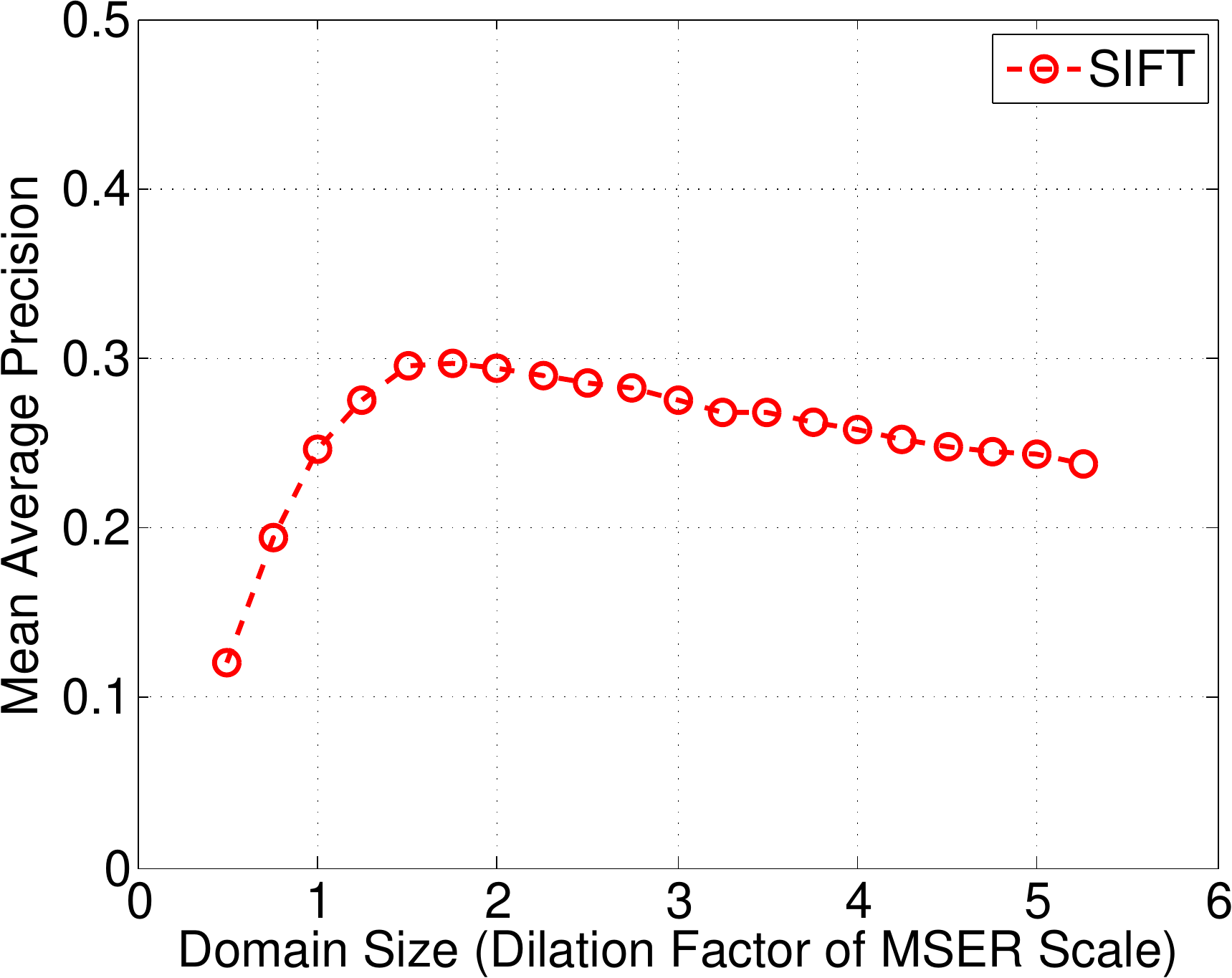}
\caption{The discriminative power of a descriptor (\eg~mAP of SIFT) increases with the size of the domain, but so does the probability of straddling an occlusion and the approximation error of the imaging model implicit in the detector/descriptor. This effect, which also depends on the base size, is most pronounced when occlusions are present, but is present even on the Oxford dataset, shown above.}
\label{fig-performance-vs-base-size}
\end{figure}

\section*{Acknowledgments}
We are thankful to Nikolaos Karianakis for conducting the comparison with various forms of CNNs, and to Philipp Fischer, Alexey Dosovitskiy and Thomas Brox for sharing their dataset, evaluation protocol and comments. Research sponsored in part by NGA HM02101310004, leveraging on theoretical work conducted under the aegis of ONR N000141110863, NSF RI-1422669, ARO W911NF-11-1-0391, and FA8650-11-1-7156. 

\bibliographystyle{ieee}

\begin{thebibliography}{10}\itemsep-1pt

\bibitem{bahadur1954sufficiency}
R.~R. Bahadur.
\newblock Sufficiency and statistical decision functions.
\newblock {\em Annals of Mathematical Statistics}, 25(3), pages 423--462, 1954.

\bibitem{surf}
H.~Bay, T.~Tuytelaars, and L.~V. Gool.
\newblock {Surf: Speeded up robust features}.
\newblock In {\em Proc. of the European Conference on Computer Vision (ECCV)}, pages 404--417, Springer, 2006.

\bibitem{berg01geometric}
A.~Berg and J.~Malik.
\newblock Geometric blur for template matching.
\newblock In {\em Proc. of the Conference on Computer Vision and Pattern Recognition (CVPR)}, IEEE, 2001.

\bibitem{boureau2010theoretical}
Y.~L. Boureau, J.~Ponce, and Y.~LeCun.
\newblock A theoretical analysis of feature pooling in visual recognition.
\newblock In {\em Proc. of the International Conference on Machine Learning (ICML)}, pages 111--118, 2010.

\bibitem{rosasco}
J.~V. Bouvrie, L.~Rosasco, and T.~Poggio.
\newblock On invariance in hierarchical models.
\newblock In {\em Advances in Neural Information Processing Systems (NIPS)}, pages 162--170, 2009.

\bibitem{mallatB11}
J.~Bruna and S.~Mallat.
\newblock Classification with scattering operators.
\newblock In {\em Proc. of the Conference on Computer Vision and Pattern Recognition (CVPR)}, IEEE, 2011.

\bibitem{chanE05}
T.~Chan and S.~Esedoglu.
\newblock {Aspects of Total Variation Regularized L1 Function Approximation}.
\newblock {\em SIAM Journal on Applied Mathematics}, 65(5), page 1817, 2005.

\bibitem{chandrasekhar2009chog}
V.~Chandrasekhar, G.~Takacs, D.~Chen, S.~Tsai, R.~Grzeszczuk, and B.~Girod.
\newblock Chog: Compressed histogram of gradients a low bit-rate feature descriptor.
\newblock In {\em Proc. of the Conference on Computer Vision and Pattern Recognition (CVPR)}, pages 2504--2511, IEEE, 2009.

\bibitem{chen2011diffusion}
C.~Chen and H.~Edelsbrunner.
\newblock Diffusion runs low on persistence fast.
\newblock In {\em Proc. of the International Conference on Computer Vision (ICCV)}, pages 423--430, IEEE, 2011.

\bibitem{dalalT05}
N.~Dalal and B.~Triggs.
\newblock Histograms of oriented gradients for human detection.
\newblock In {\em Proc. of the Conference on Computer Vision and Pattern Recognition (CVPR)}, IEEE, 2005.

\bibitem{imagenet_cvpr09}
J.~Deng, W.~Dong, R.~Socher, L.-J. Li, K.~Li, and L.~Fei-Fei.
\newblock Imagenet: A large-scale hierarchical image database.
\newblock In {\em Proc. of the Conference on Computer Vision and Pattern Recognition (CVPR)}, pages 248--255, IEEE, 2009.

\bibitem{dongKDHBS15}
J.~Dong, N.~Karianakis, D.~Davis, J.~Hernandez, J.~Balzer, and S.~Soatto.
\newblock Multi-view feature engineering and learning.
\newblock In {\em Proc. of the Conference on Computer Vision and Pattern Recognition (CVPR)}, IEEE, 2015.

\bibitem{dosovitskiy2013unsupervised}
A.~Dosovitskiy, J.~T. Springenberg, and T.~Brox.
\newblock Unsupervised feature learning by augmenting single images.
\newblock {\em ArXiv preprint:1312.5242}, 2013.

\bibitem{farabet2012scene}
C.~Farabet, C.~Couprie, L.~Najman, and Y.~LeCun.
\newblock Scene parsing with multiscale feature learning, purity trees, and optimal covers.
\newblock {\em ArXiv preprint:1202.2160}, 2012.

\bibitem{felzenswalb}
P.~Felzenszwalb, D.~McAllester, and D.~Ramanan.
\newblock A discriminatively trained, multiscale, deformable part model.
\newblock In {\em Proc. of the Conference on Computer Vision and Pattern Recognition (CVPR)}, pages 1--8, IEEE, 2008.

\bibitem{fischer2014descriptor}
P.~Fischer, A.~Dosovitskiy, and T.~Brox.
\newblock Descriptor matching with convolutional neural networks: a comparison to sift.
\newblock {\em ArXiv preprint:1405.5769}, 2014.

\bibitem{fragoso2013evsac}
V.~Fragoso, P.~Sen, S.~Rodriguez, and M.~Turk.
\newblock Evsac: Accelerating hypotheses generation by modeling matching scores with extreme value theory.
\newblock In {\em Proc. of the International Conference on Computer Vision (ICCV)}, pages 2472--2479, IEEE, 2013.

\bibitem{gong2014multi}
Y.~Gong, L.~Wang, R.~Guo, and S.~Lazebnik.
\newblock Multi-scale orderless pooling of deep convolutional activation features.
\newblock {\em ArXiv preprint:1403.1840}, 2014.

\bibitem{guilleminP74}
V.~Guillemin and A.~Pollack.
\newblock {\em Differential Topology}. Prentice-Hall, 1974.

\bibitem{hamel2011temporal}
P.~Hamel, S.~Lemieux, Y.~Bengio, and D.~Eck.
\newblock Temporal pooling and multiscale learning for automatic annotation and ranking of music audio.
\newblock In {\em Proc. of the International Society of Music Information Retrieval}, pages 729--734, 2011.

\bibitem{hassner2012sifts}
V.~Hassne, T.and~Mayzels and L.~Zelnik-Manor.
\newblock On sifts and their scales.
\newblock In {\em Proc. of the Conference on Computer Vision and Pattern Recognition (CVPR)}, pages 1522--1528, IEEE, 2012.

\bibitem{jia2012beyond}
Y.~Jia, C.~Huang, and T.~Darrell.
\newblock Beyond spatial pyramids: Receptive field learning for pooled image features.
\newblock In {\em Proc. of the Conference on Computer Vision and Pattern Recognition (CVPR)}, pages 3370--3377, IEEE, 2012.

\bibitem{lecun2012learning}
Y.~LeCun.
\newblock Learning invariant feature hierarchies.
\newblock In {\em Proc. of the European Conference on Computer Vision (ECCV)}, pages 496--505, Springer, 2012.

\bibitem{leeS11}
T.~Lee and S.~Soatto.
\newblock Learning and matching multiscale template descriptors for real-time detection, localization and tracking.
\newblock In {\em Proc. of the Conference on Computer Vision and Pattern Recognition (CVPR)}, pages 1457--1464, IEEE, 2011.

\bibitem{LeeS10}
T.~Lee and S.~Soatto.
\newblock Video-based descriptors for object recognition.
\newblock In {\em Image and Vision Computing}, 29(10):639--652, 2011.

\bibitem{lindeberg98}
T.~Lindeberg.
\newblock Principles for automatic scale selection.
\newblock {\em Technical Report}, KTH, Stockholm, CVAP, 1998.

\bibitem{lowe04distinctive}
D.~G. Lowe.
\newblock Distinctive image features from scale-invariant keypoints.
\newblock In {\em International Journal of Computer Vision}, 2(60), pages 91--110, Springer, 2004.

\bibitem{matas03robust}
J.~Matas, O.~Chum, M.Urban, and T.~Pajdla.
\newblock Robust wide baseline stereo from maximally stable extremal regions.
\newblock In {\em Proc. of the British Machine Vision Conference (BMVC)}, 2002.

\bibitem{memisevic}
R.~Memisevic.
\newblock Learning to relate images.
\newblock In {\em IEEE Trans. on Pattern Analysis and Machine Intelligence.}, 35(8):1829--1846, 2013.

\bibitem{mikolajc03survey}
K.~Mikolajczyk and C.~Schmid.
\newblock A performance evaluation of local descriptors.
\newblock In {\em IEEE Trans. on Pattern Analysis and Machine Intelligence.}, pages 1615--1630, 2005.

\bibitem{mikolajczyk04comparison}
K.~Mikolajczyk, T.~Tuytelaars, C.~Schmid, A.~Zisserman, J.~Matas, F.~Schaffalitzky, T.~Kadir, and L.~V. Gool.
\newblock A comparison of affine region detectors.
\newblock In {\em International Journal of Computer Vision}, 1(60):63--86, Springer, 2004.

\bibitem{moreels2007evaluation}
P.~Moreels and P.~Perona.
\newblock Evaluation of features detectors and descriptors based on 3d objects.
\newblock In {\em International Journal of Computer Vision}, 73(3):263--284, Springer, 2007.

\bibitem{ranzato2007unsupervised}
M.~Ranzato, F.~J. Huang, Y.-L. Boureau, and Y.~LeCun.
\newblock Unsupervised learning of invariant feature hierarchies with applications to object recognition.
\newblock In {\em Proc. of the Conference on Computer Vision and Pattern Recognition (CVPR)}, pages 1--8, IEEE, 2007.

\bibitem{serre2007feedforward}
T.~Serre, A.~Oliva, and T.~Poggio.
\newblock A feedforward architecture accounts for rapid categorization.
\newblock {\em Proceedings of the National Academy of Sciences}, 104(15), pages 6424--6429, 2007.

\bibitem{shao98}
J.~Shao.
\newblock {\em Mathematical Statistics}. Springer Verlag, 1998.

\bibitem{simonyan2013deep}
K.~Simonyan, A.~Vedaldi, and A.~Zisserman.
\newblock Deep fisher networks for large-scale image classification.
\newblock In {\em Advances in Neural Information Processing Systems (NIPS)}, pages 163--171, 2013.

\bibitem{simonyan2014learning}
K.~Simonyan, A.~Vedaldi, and A.~Zisserman.
\newblock Learning local feature descriptors using convex optimisation.
\newblock In {\em IEEE Trans. on Pattern Analysis and Machine Intelligence.}, 2(4), 2014.

\bibitem{sivic2003video}
J.~Sivic and A.~Zisserman.
\newblock Video google: A text retrieval approach to object matching in videos.
\newblock In {\em Proc. of the Conference on Computer Vision and Pattern Recognition (CVPR)}, pages 1470--1477. IEEE, 2003.

\bibitem{soatto10}
S.~Soatto.
\newblock Steps towards a theory of visual information: Active perception, signal-to-symbol conversion and the interplay between sensing and control.
\newblock {\em ArXiv preprint: 1110.2053}, 2010.

\bibitem{soattoC14ICLR}
S.~Soatto and A.~Chiuso.
\newblock Visual scene representations: Sufficiency, minimality, invariance and deep approximation.
\newblock {\em ArXiv preprint: 1411.7676}, 2014.

\bibitem{soattoDK15}
S.~Soatto, J.~Dong, and N.~Karianakis.
\newblock Visual scene representations: Contrast, scaling and occlusion.
\newblock {\em ArXiv preprint: 1412.6607}, 2014.

\bibitem{sundaramoorthiPVS09}
G.~Sundaramoorthi, P.~Petersen, V.~S. Varadarajan, and S.~Soatto.
\newblock On the set of images modulo viewpoint and contrast changes.
\newblock In {\em Proc. of the Conference on Computer Vision and Pattern Recognition (CVPR)}, IEEE, 2009.

\bibitem{susskind}
J.~Susskind, R.~Memisevic, G.~E. Hinton, and M.~Pollefeys.
\newblock Modeling the joint density of two images under a variety of transformations.
\newblock In {\em Proc. of the Conference on Computer Vision and Pattern Recognition (CVPR)}, pages 2793--2800, IEEE, 2011.

\bibitem{tau2014dense}
M.~Tau and T.~Hassner.
\newblock Dense correspondences across scenes and scales.
\newblock {\em ArXiv preprint:1406.6323}, 2014.

\bibitem{taylor}
G.~W. Taylor, R.~Fergus, Y.~LeCun, and C.~Bregler.
\newblock Convolutional learning of spatio-temporal features.
\newblock In {\em Proc. of the European Conference on Computer Vision (ECCV)}, pages 140--153, Springer, 2010.

\bibitem{vlfeat}
A.~Vedaldi and B.~Fulkerson.
\newblock Vlfeat: An open and portable library of computer vision algorithms.
\newblock In {\em Proc. of the International Conference on Multimedia}, pages 1469--1472, ACM, 2010.

\bibitem{vondrick2013hog}
C.~Vondrick, A.~Khosla, T.~Malisiewicz, and A.~Torralba.
\newblock Hog-gles: Visualizing object detection features.
\newblock In {\em Proc. of the International Conference on Computer Vision (ICCV)}, IEEE, 2013.

\bibitem{winder2007learning}
S.~Winder and M.~Brown.
\newblock Learning local image descriptors.
\newblock In {\em Proc. of the Conference on Computer Vision and Pattern Recognition (CVPR)}, pages 1--8, IEEE, 2007.

\end{thebibliography}

\appendix
\onecolumn

\section{Relation to Sampling Theory} 
\label{sect-relation}

This first section summarizes the background needed for the derivation, reported in the next section. 

\subsection{Sampling and aliasing}
\label{sect-sampling}

In this section we refer to a general scalar signal $f:\real \rightarrow \real; x \mapsto f(x)$, for instance the projection of the albedo of the scene onto a scanline. We define a {\em detector} to be a mechanism to select samples $x_i$, and a {\em descriptor} $\phi_i$ to be a statistic computed from the signal of interest and associated with the sample $i$. In the simplest case, $x$ is regularly sampled, so the detector does not depend on the signal, and the descriptor is simply the value of the function at the sample $\phi_{i} = f(x_i)$. Other examples include:

\subsubsection{Regular sampling (Shannon '49)}

The detector is trivial: $\{x_i\} = \Lambda$ is a lattice, independent of $f$. The descriptor  is a weighted average of $f$ in a neighborhood of fixed size $\sigma$ (possibly unbounded) around $x_i$: $\phi_i = \phi(\{f(x), \ x \in {\cal B}_{\sigma}(x_i)\})$. Neither the detector nor the descriptor function $\phi$ depend on $f$ (although the {\em value} of the latter, of course, does). 

If the signal was band-limited, Shannon's sampling theory would offer guarantees on the exact reconstruction $\hat f$ of $f(x), x \in \real$ from its sampled representation $\{x_i, \phi_i\}$.  Unfortunately, the signals of interest are not band-limited (images are discontinuous), and therefore the reconstruction $\hat f$ can only approximate $f$. Typically, the approximation include ``alien structures,'' \ie spurious extrema and discontinuities in $\hat f$ that do not exist in $f$. This phenomenon is known as {\em aliasing.} To reduce its effects, one can {\em replace} the original data $f$ with another $\tilde f$ that is (closer to) band-limited and yet close to $f$, so that the samples can encode $\hat f = \tilde f$ free of aliasing artifacts. The conflicting requirements of faithful approximation of $f$ and restriction on bandwidth trade off discriminative power (reconstruction error) with complexity, which is one of the goals of communications engineering. This tradeoff can be optimized by choice of {\em anti-aliasing operator}, that is the function that produces $\tilde f$ from $f$, usually via convolution with a low-pass filter. In our context, we seek for a tradeoff between discriminative power and {\em sensitivity to nuisance factors}. This will come naturally when anti-aliasing is performed with respect to the action of nuisance transformations.

\begin{figure}[h]
\begin{center}
\subfigure{\includegraphics[width=.16\columnwidth]{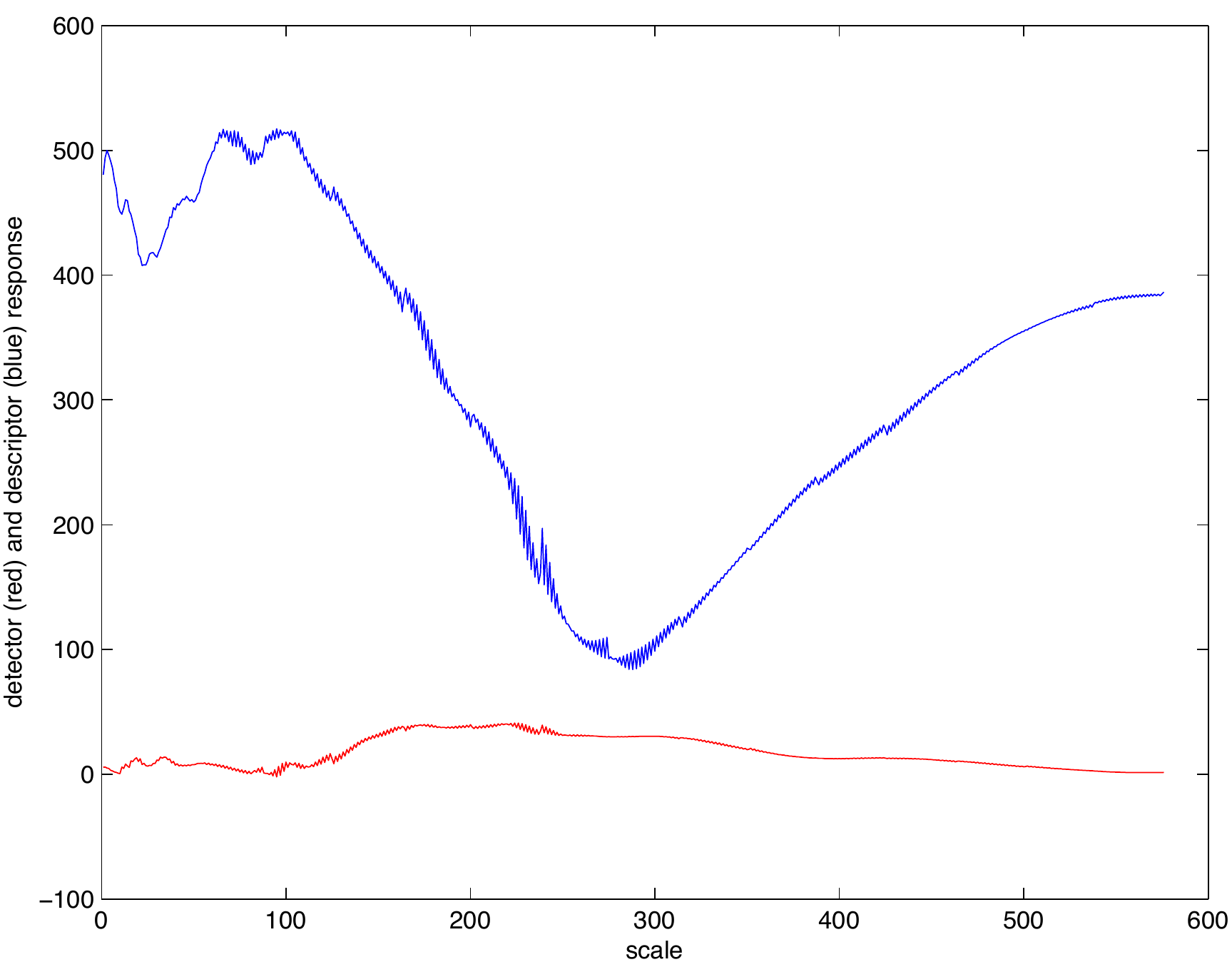}}
\subfigure{\includegraphics[width=.16\columnwidth]{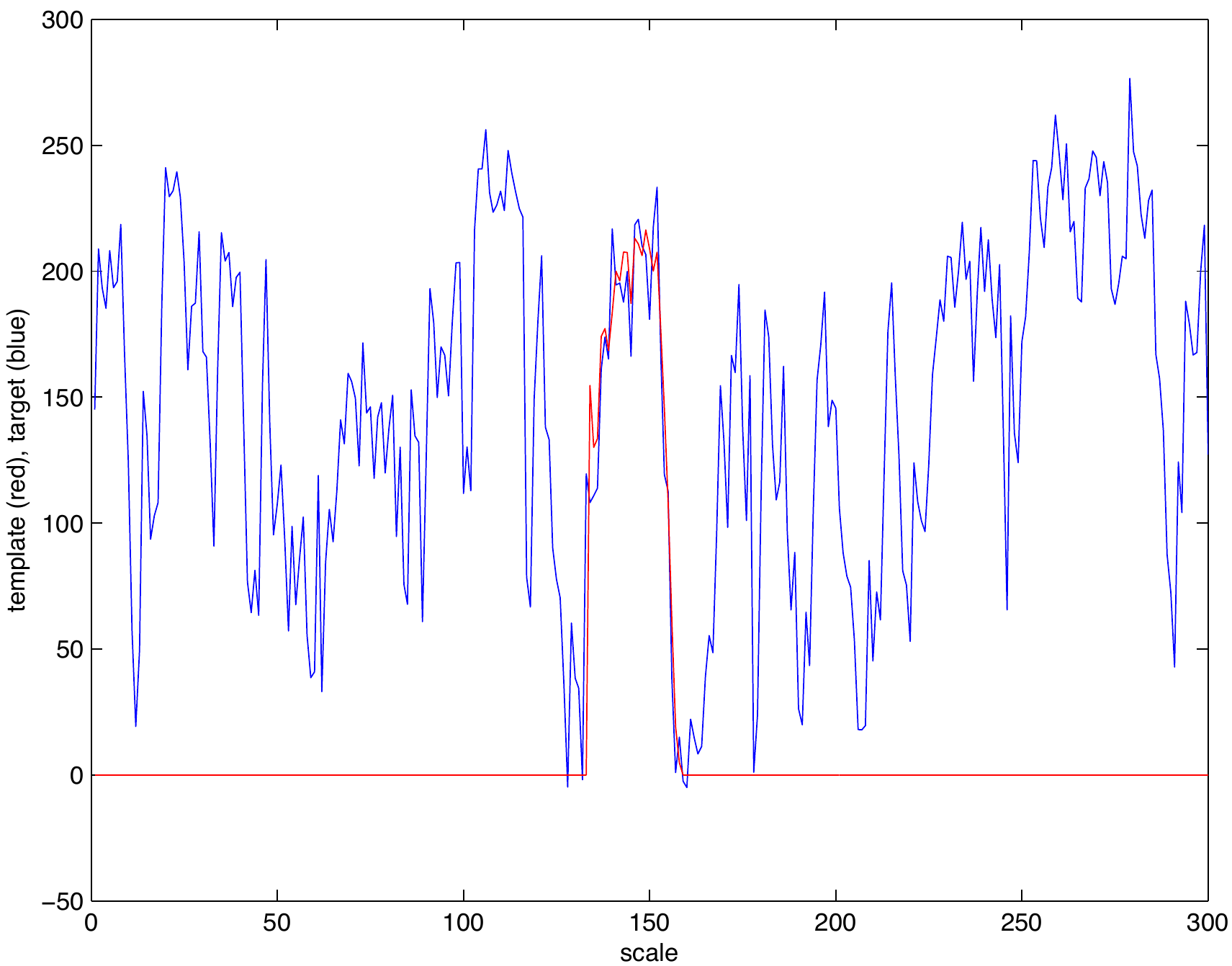}}
\subfigure{\includegraphics[width=.16\columnwidth]{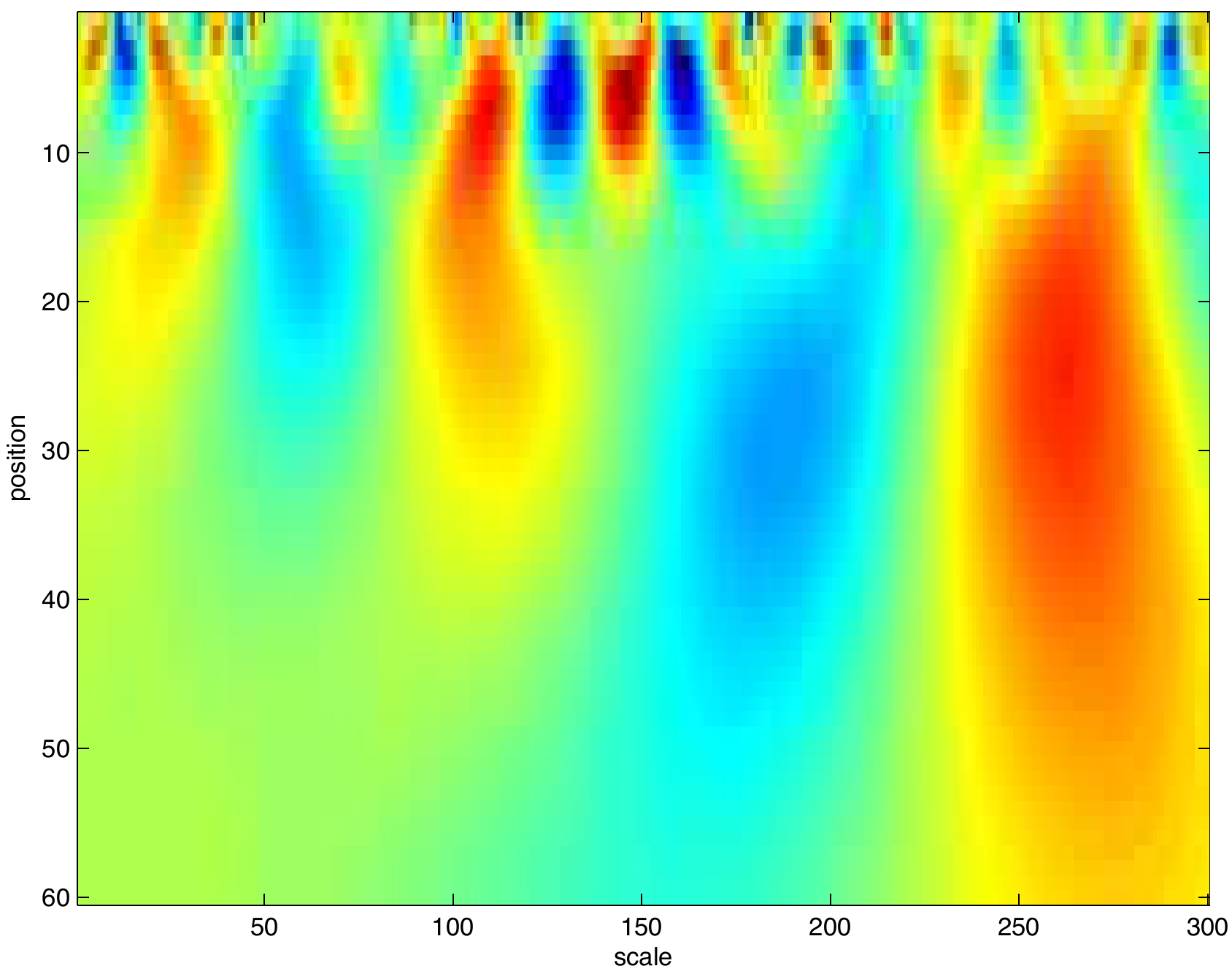}}\vspace{-.2cm}\\
\subfigure{\includegraphics[width=.16\columnwidth]{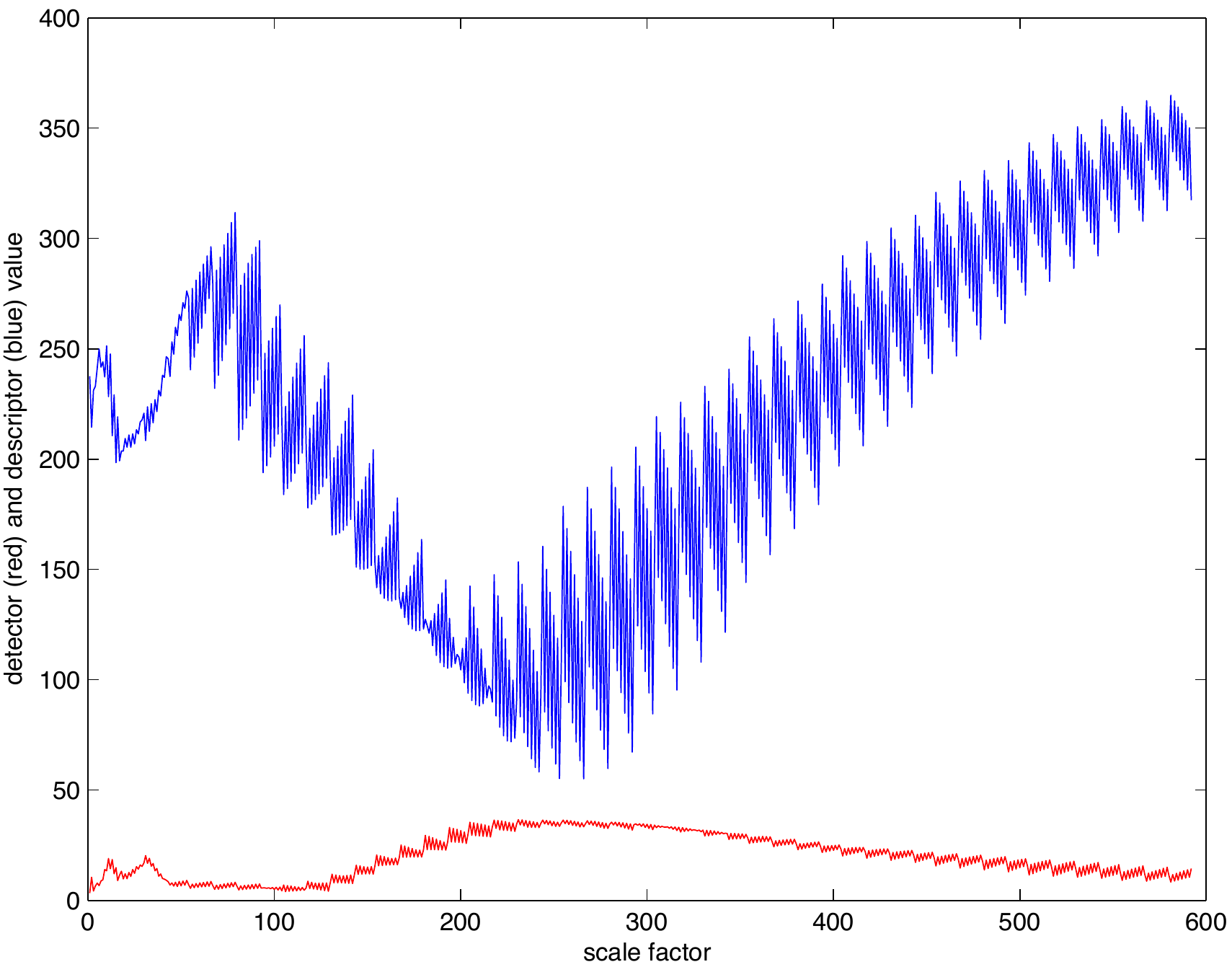}}
\subfigure{\includegraphics[width=.16\columnwidth]{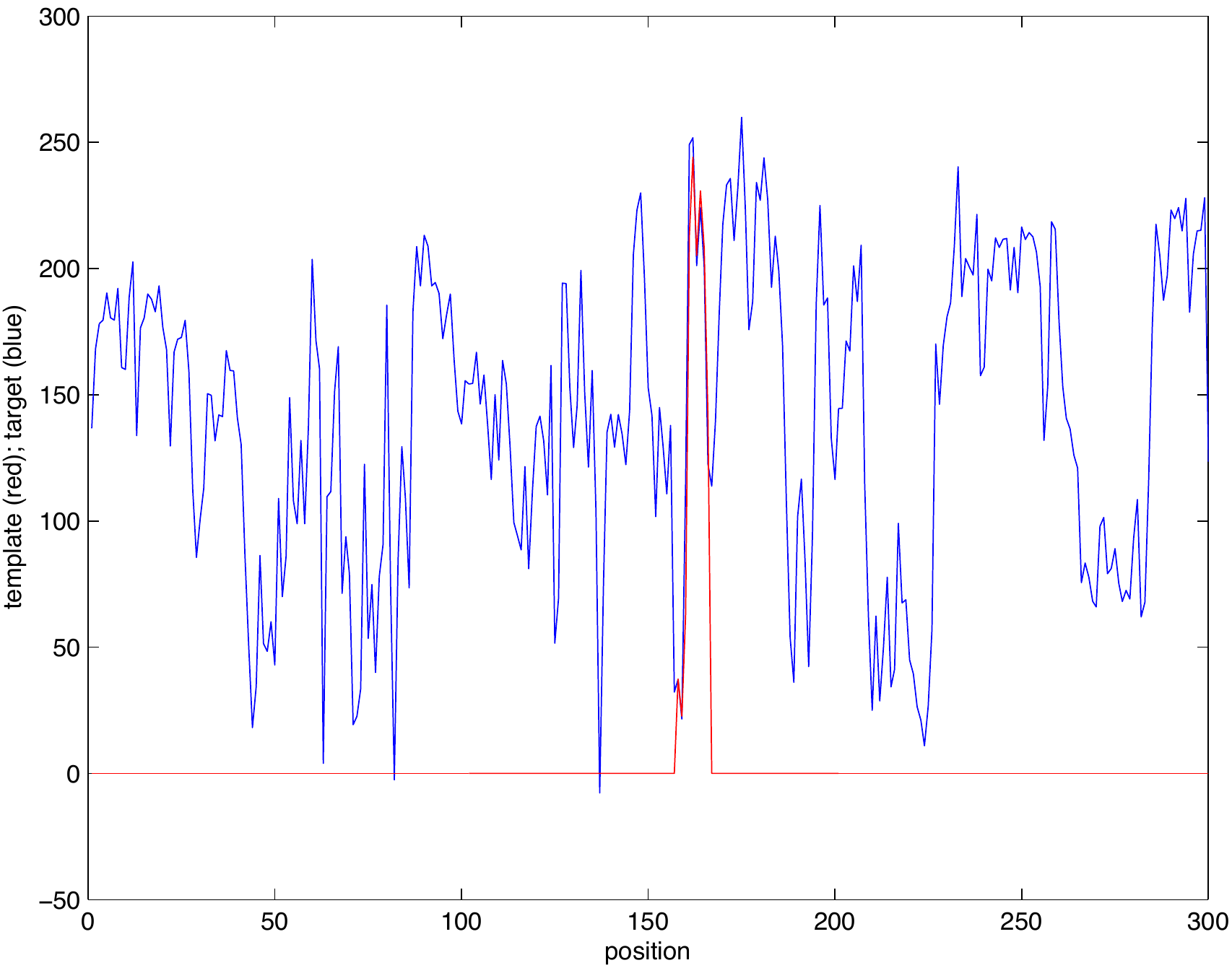}}
\subfigure{\includegraphics[width=.16\columnwidth]{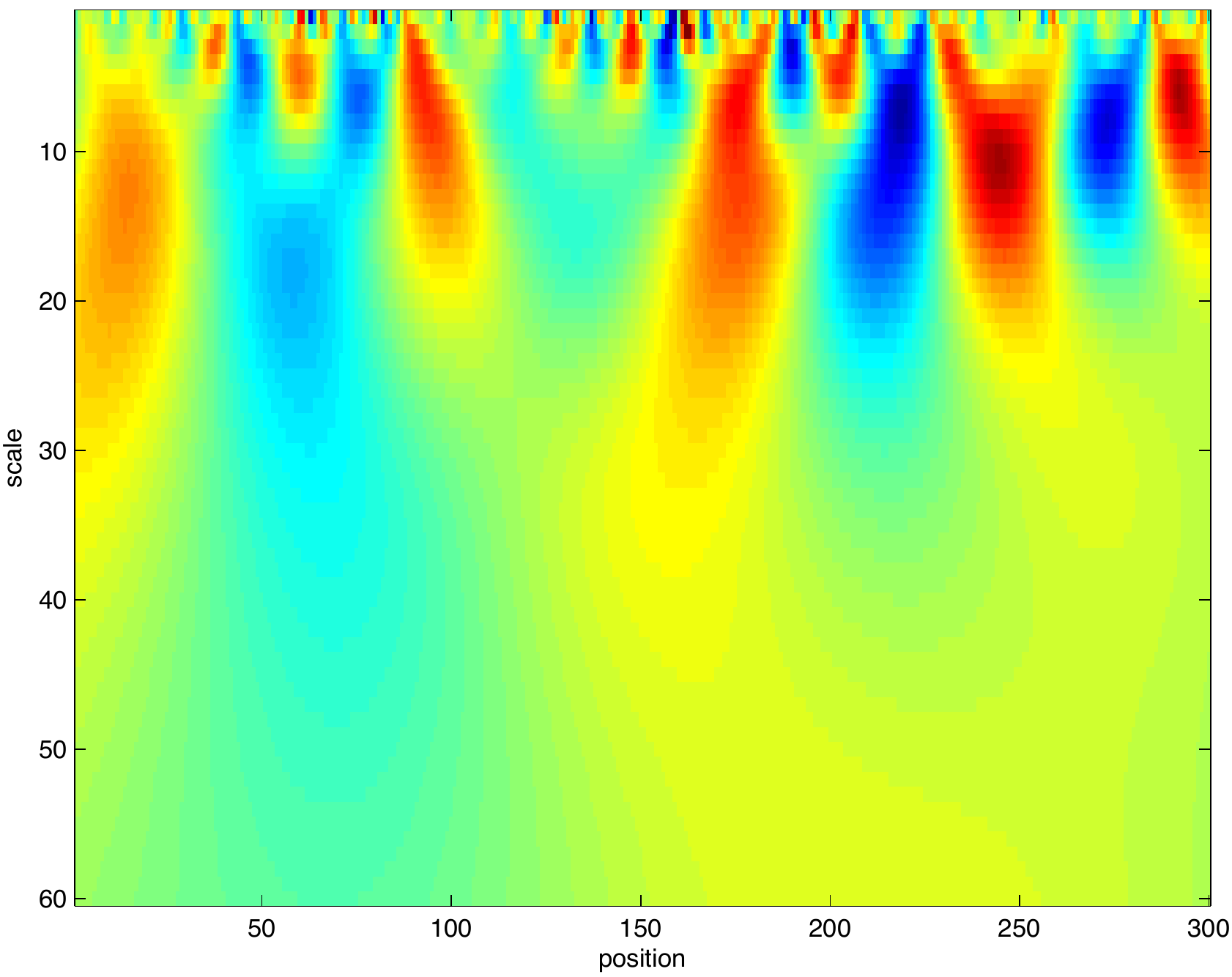}}\vspace{-.2cm}\\
\subfigure{\includegraphics[width=.16\columnwidth]{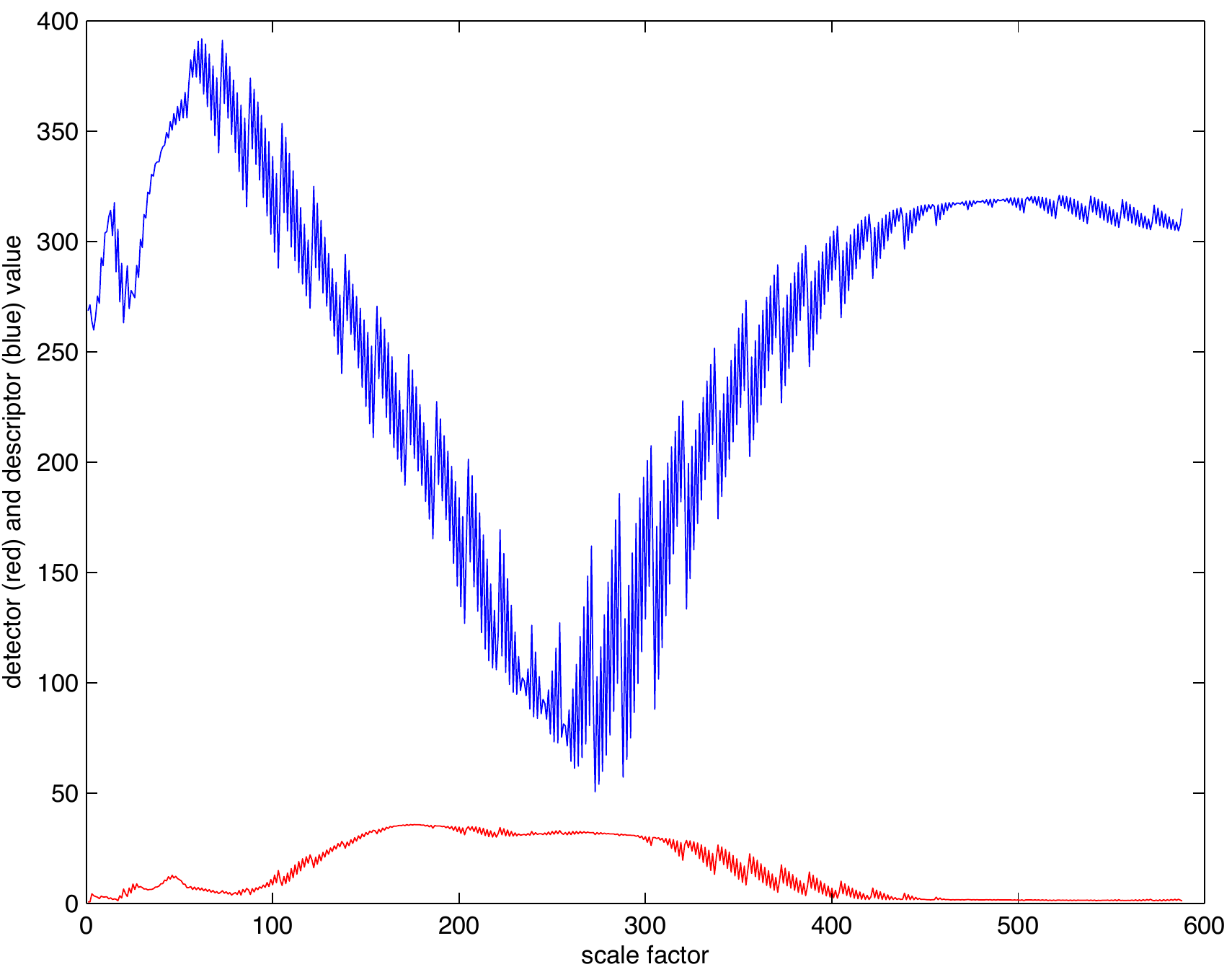}}
\subfigure{\includegraphics[width=.16\columnwidth]{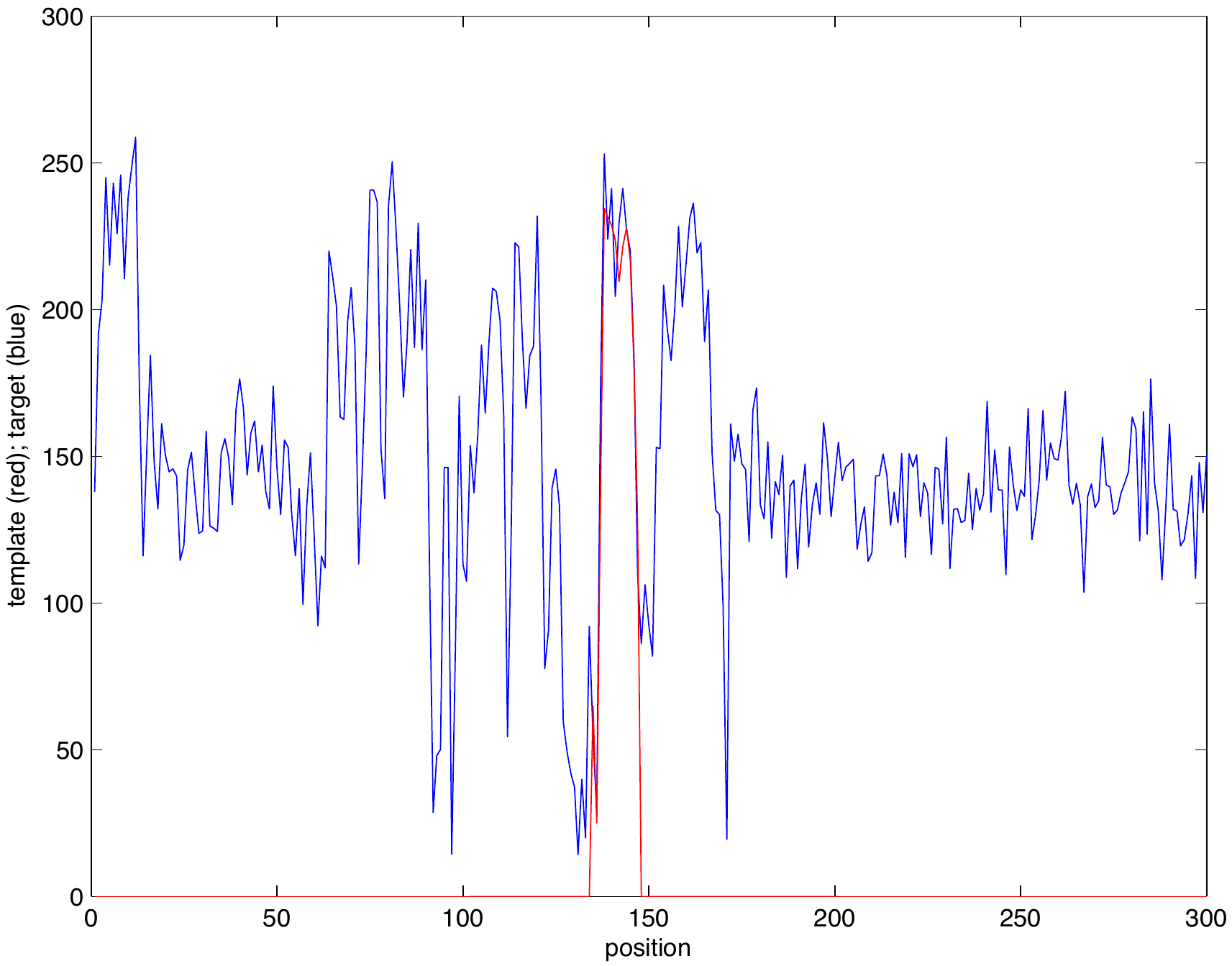}}
\subfigure{\includegraphics[width=.16\columnwidth]{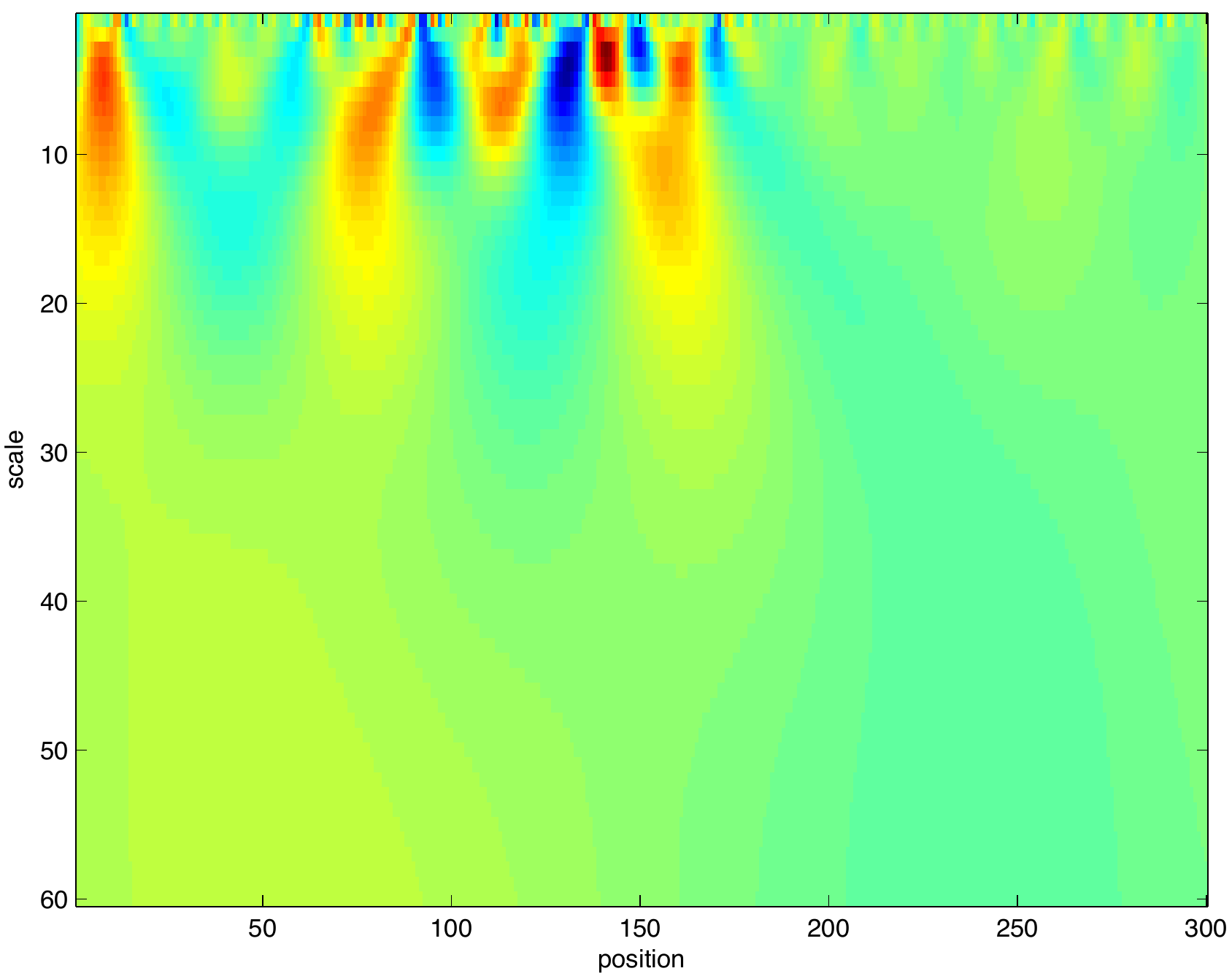}}\vspace{-.2cm}
\end{center}
\caption{{\sl Detector specificity vs. descriptor sensitivity.}  (Left) Change of detector response (red) as a function of scale, computed around the optimal location and scale (here corresponding to a value of $245$), and corresponding change of descriptor value (blue). An ideal detector would have high specificity (sharp maximum around the true scale) and an ideal descriptor would have low sensitivity (broad minimum around the same). The opposite is true. This means that it is difficult to precisely select scale, and selection error results in large changes in the descriptor. Experiments are for the DoG detector and identity descriptor. Referring to the notation in Appendix (see details therein), (middle) template $\rho$ (red) and target $f$ (blue). (Right) corresponding scale-space $[f]$. Note that the maximum detector response may even not correspond to the true location. The jaggedness of the response is an aliasing artifact. 
}
\vspace{-.2cm}
\label{fig-specificity-main}
\label{fig-specificity}
\end{figure}

\subsubsection{Adaptive sampling (Landau '67)}

The detector could be ``adapted'' to $f$ by designing a functional $\psi$ that selects samples $\{x_i\} = \psi(f)$. Typically, spatial frequencies of $f$ modulate the length of the interval $\delta x_i \doteq x_{i+1}-x_i$.
A special case of adaptive sampling that does not requires stationarity assumptions is described next. The descriptor may also depend on $\psi$, \eg by making the statistic depend on a neighborhood of variable size $\sigma_i$: $\phi_i = \phi(\{f(x), \ x \in {\cal B}_{\sigma_i}(x_i)\})$.

\subsubsection{Tailored sampling (Logan '77)}
\label{sect-tailored}

For signals that are neither stationary nor band-limited, we can leverage on the violations of these assumptions to design a detector. For instance, if $f$ contains discontinuities, the detector can place samples at discontinuous locations (``corners''). For band-limited signals, the detector can place samples at critical points (maxima, or ``blobs'', minima, saddles). A (location-scale) {\em co-variant detector} is a functional $\psi$ whose zero-level sets 
\be
\psi(f; s,t ) = 0
\ee
define isolated (but typically multiple) samples of scales $s_i>0$ and locations $t_i\in \real$ locally as a function of $f$ via the implicit function theorem \cite{guilleminP74}, in such a way that if $f$ is transformed, for instance via a linear operator depending on location $\tau$ and scale $\sigma$ parameters, $W(\sigma,\tau)f$, then so are the samples: $\psi(W(\sigma,\tau)f; s+\sigma,t+\tau ) = 0$.

The associated descriptor can then be any function of the image in the reference frame defined by the samples $t_i, s_i$, the most trivial being the restriction of the original function $f$ to the neighborhood ${\cal B}_{s_i}(t_i)$. This, however, does not reduce the dimensionality of the representation. Other descriptors can compute statistics of the signal in the neighborhood, or on the entire line. Note that descriptors $\phi_i$ could have different dimensions for each $i$. 

\begin{figure}[tb]
\begin{center}
\subfigure{\includegraphics[width=.25\columnwidth]{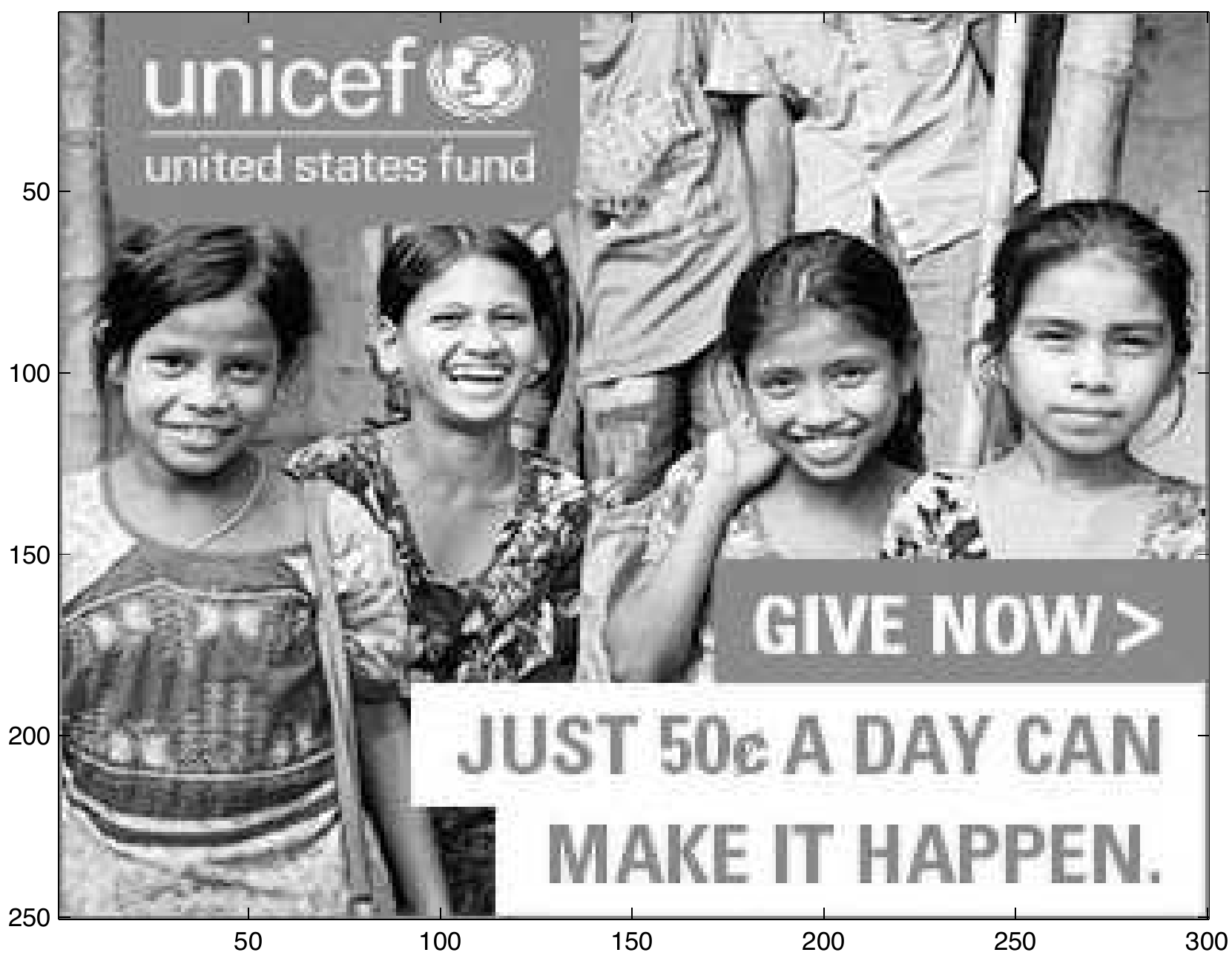}}
\subfigure{\includegraphics[width=.25\columnwidth]{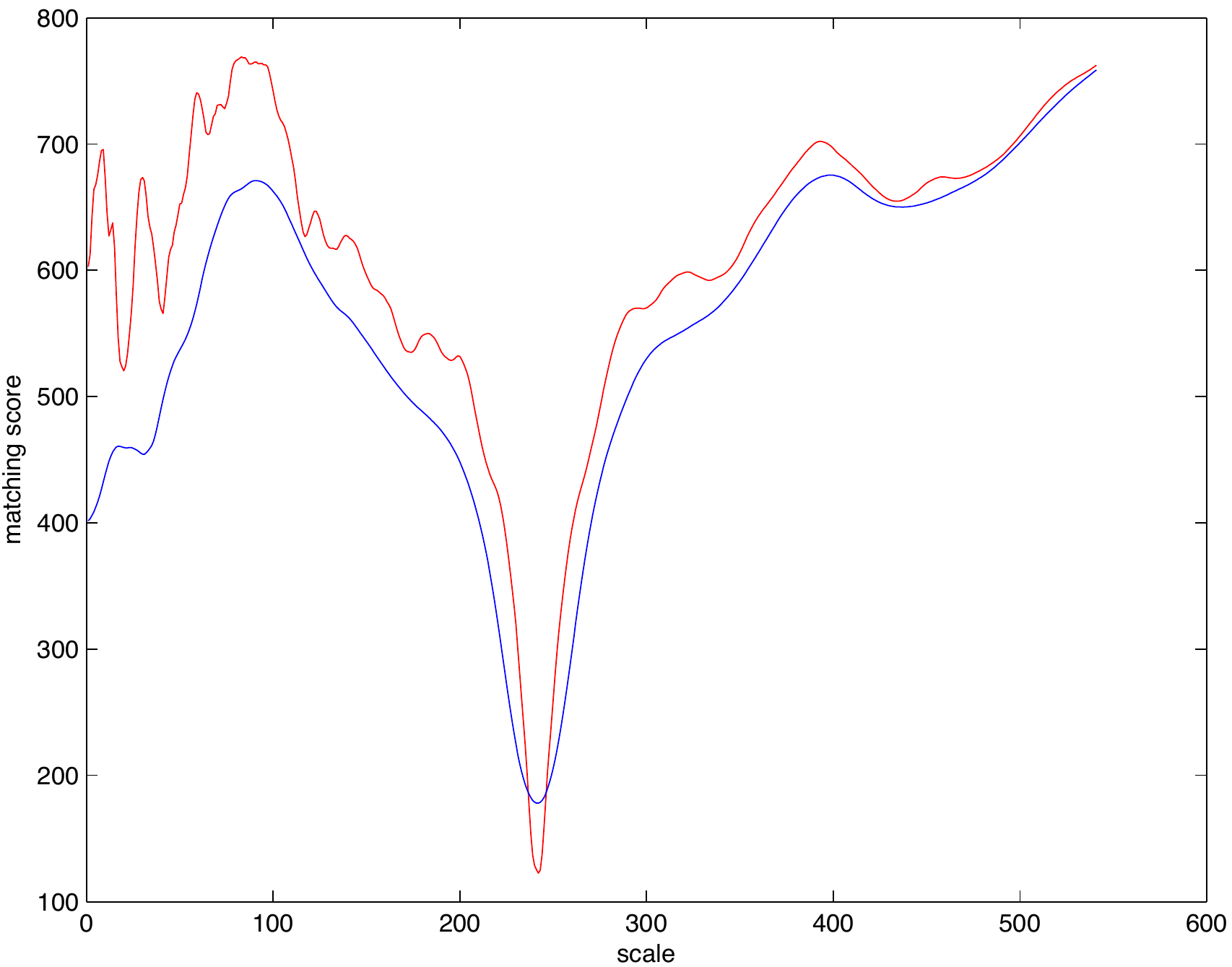}}\vspace{-.2cm} \\
\subfigure{\includegraphics[width=.25\columnwidth]{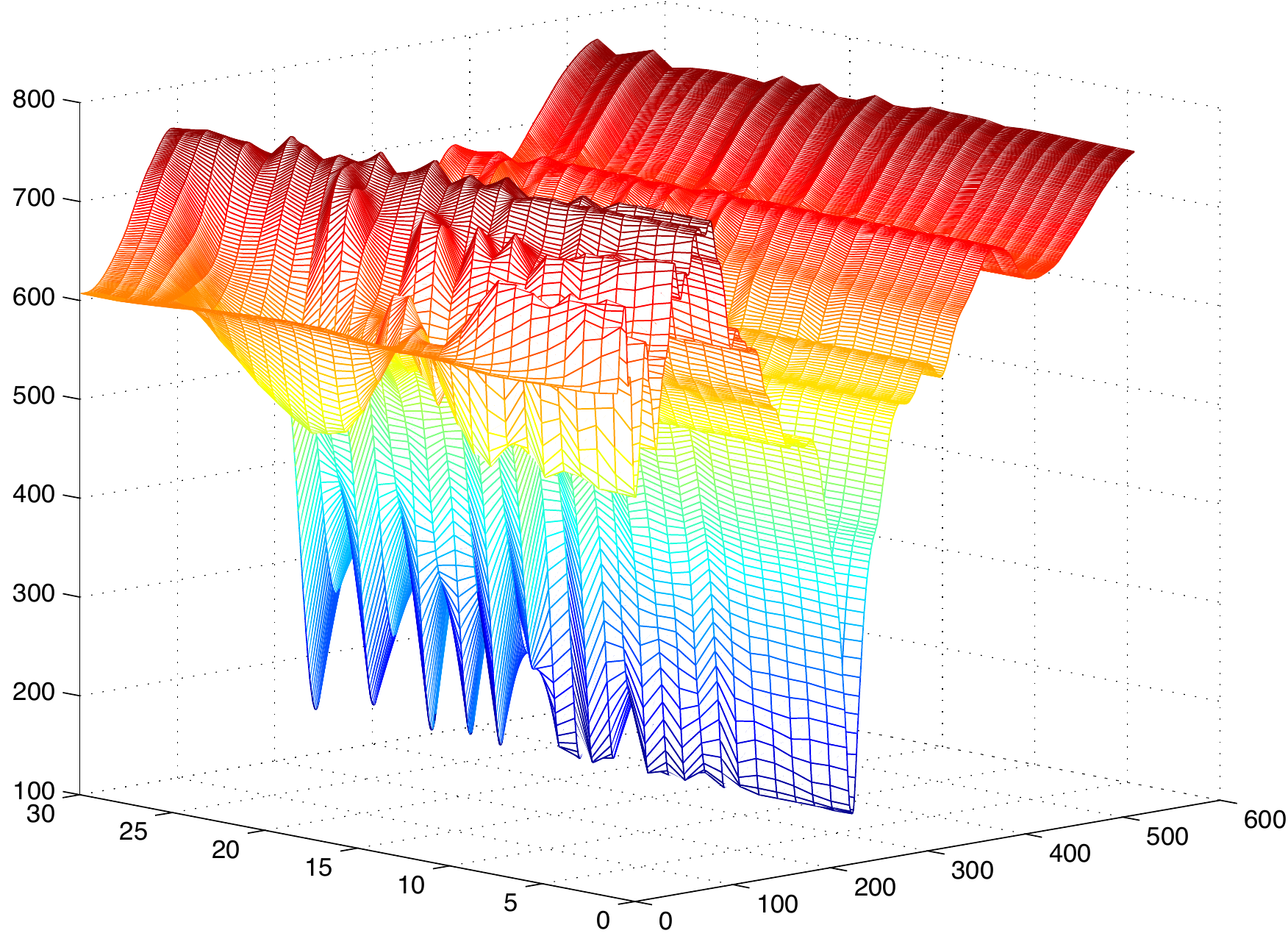}}
\subfigure{\includegraphics[width=.25\columnwidth]{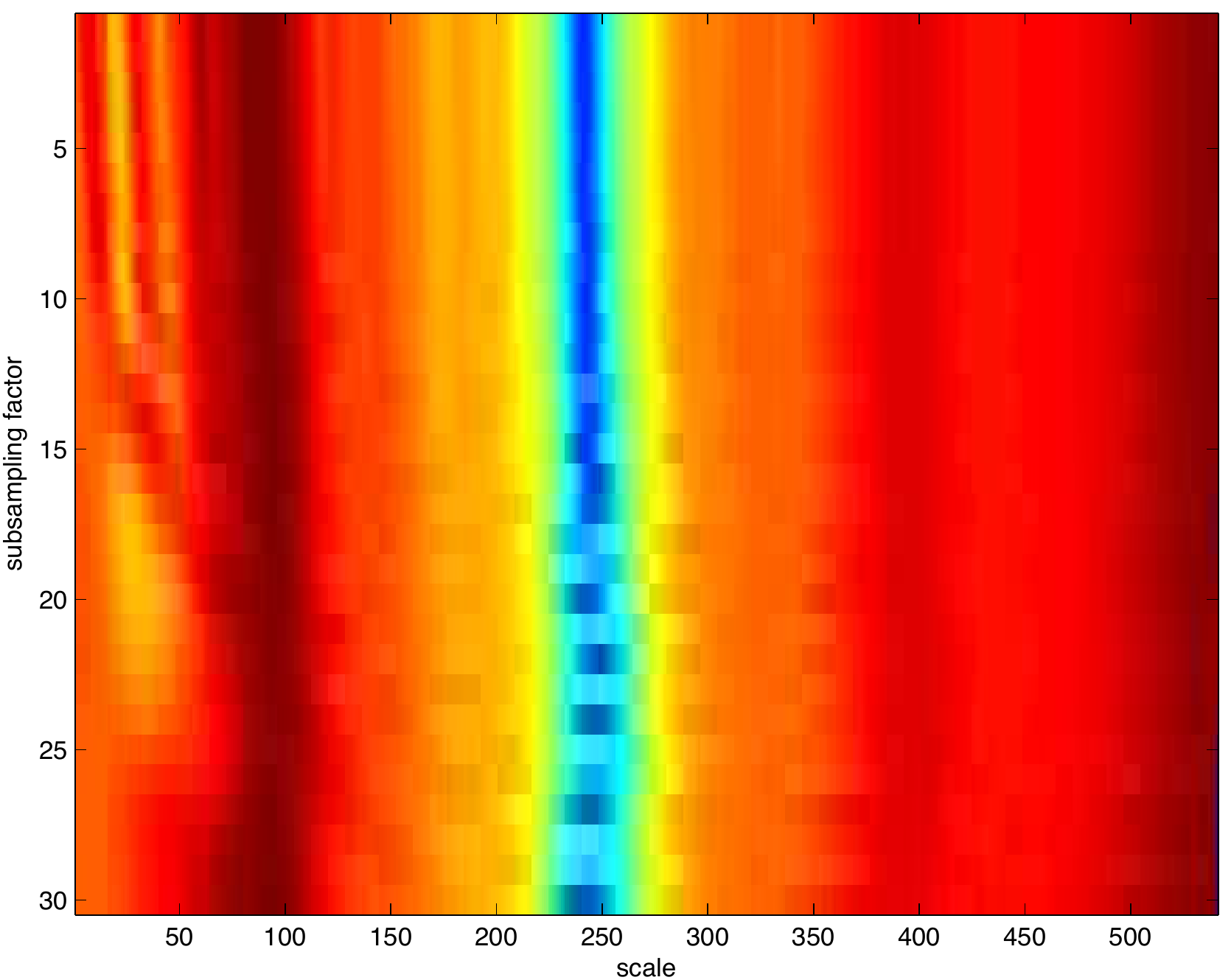}}\vspace{-.2cm}\\
\subfigure{\includegraphics[width=.25\columnwidth]{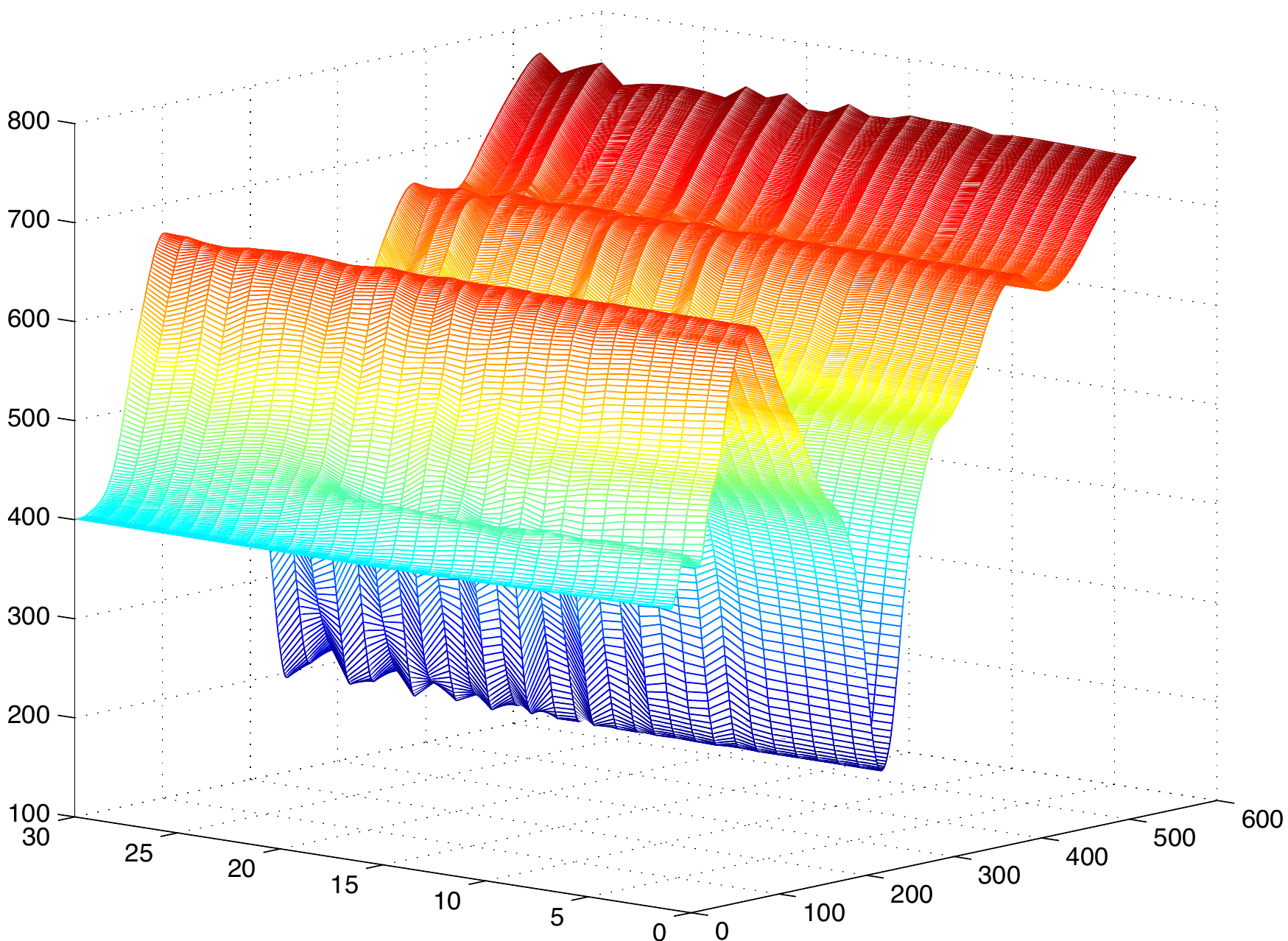}}
\subfigure{\includegraphics[width=.25\columnwidth]{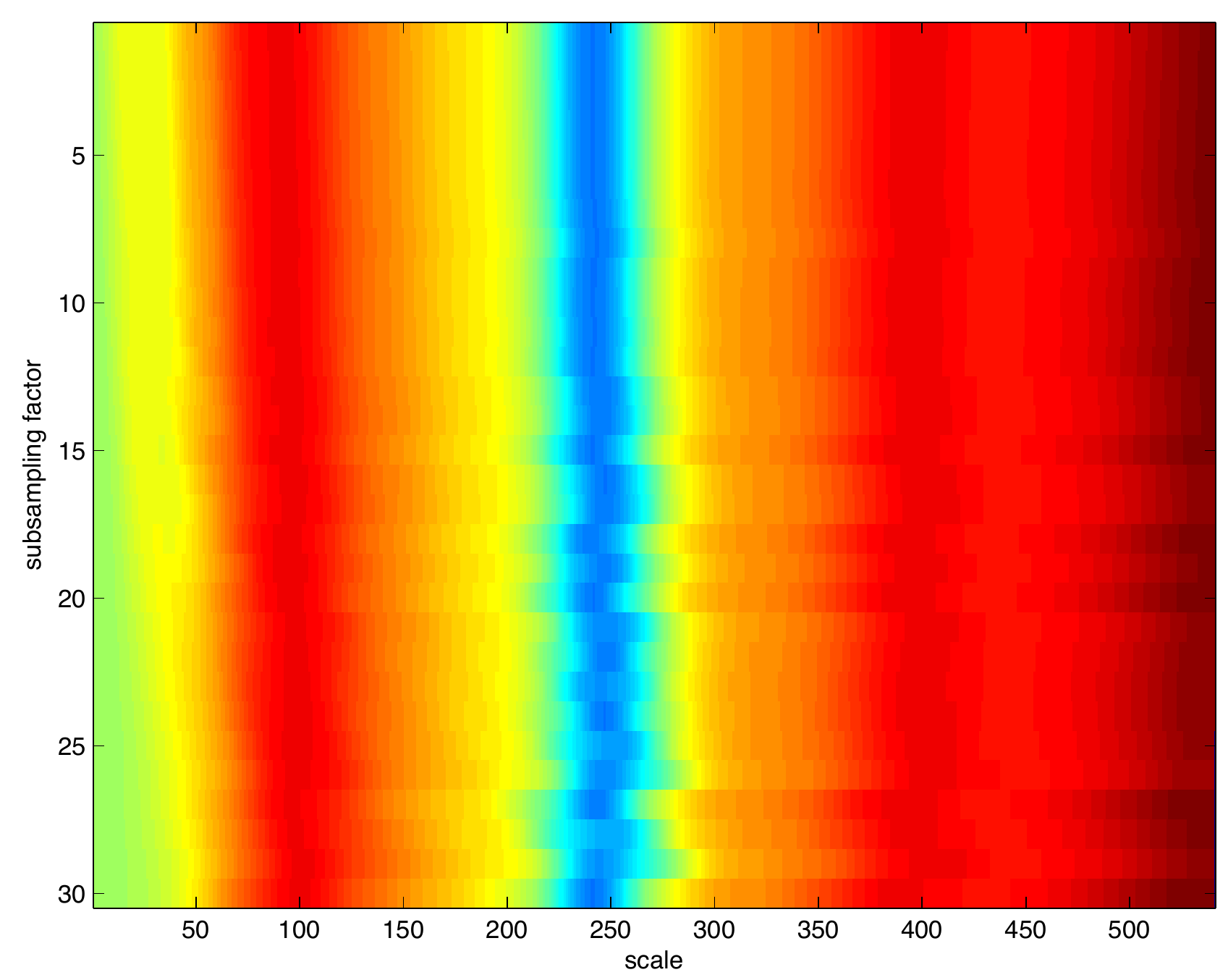}}\vspace{-.2cm}
\end{center}
\caption{{\sl Aliasing:} (Top left) A random row is selected as the target $f$ and re-scaled to yield the orbit $[f]$; a subset of $f$, cropped, re-scaled, and perturbed with noise, is chosen as the template $\rho$. The distance $E$ between $\rho$ and $[f]$ is shown in red (right) as a function of scale. The same exercise is repeated for different sub-sampling of $[f]$, and rescaled for display either as a mesh (middle left) or heat map (right) that clearly show aliasing artifacts along the optimal ridge. {\sl Anti-aliasing scale} (bottom) produces a cleaner ridge (left, right). The net effect of anti-aliasing has been to smooth the matching score $E$ (top-right, in blue) but without computing it on a fine grid. Note that the valley of the minimum is broader, denoting decreased sensitivity to scale, and the value is somewhat higher, denoting a decreased discriminative power and risk of aliasing if the value raises above that of other local minima.  
}
\vspace{-.2cm}
\label{fig-aliasing-main}
\label{fig-aliasing}
\end{figure}

\subsubsection{Anti-aliasing and ``pooling''}

In classical sampling theory, anti-aliasing refers to low-pass filtering or smoothing that {\em typically}\footnote{This central tenet of scale-space theory only holds for scalar signals. Nevertheless, genetic effects have been shown to be {\em rare} in two-dimensional Gaussian scale-space \cite{chen2011diffusion}.} does not cause {\em genetic phenomena} (spurious extrema, or {\em aliases}, appearing in the reconstruction of the smoothed signal.) Of course, anti-aliasing typically has {\em destructive} effects, in the sense of eliminating extrema that are instead present in the original signal. 

A side-effect of anti-aliasing, which has implications when the goal is {\em not} to reconstruct, but to detect or localize a signal, is to reduce the sensitivity of the relevant variable (descriptor) to variations  of the samples (detector). If we sample translations, $x_i = x + t_i$, and just store $f_i = f(x_i)$, an arbitrarily small translation of the sample $dx$ can cause an arbitrarily large variation in the representation $\delta f(x_i) = f(x_i + dx) - f_i$, when $x_i$ is a discontinuity. So, the sensitivity $S(f) = \frac{\delta f}{d x}  = \infty$. An anti-aliasing operator $\phi(f)$ should reduce sensitivity to translation: $\frac{\delta \phi(f)}{d x} \ll \frac{\delta f}{d x}$. Of course, this could be trivially achieved by choosing $\phi(f) = 0$ for any $f$. The goal is to trade off sensitivity with {\em discriminative power}. For the case of translation, this tradeoff has been described in \cite{mallatB11}. However, similar considerations holds for scale and domain-size sampling.

\section{Derivation of DSP-SIFT}
\label{sect-derivation}

The derivation of DSP-SIFT and its extensions follows a series of steps summarized as follows: 
\begin{itemize}
\item We start from the correspondence, or matching, task: Classify a given datum $f$ (test image, or target) as coming from one of $M$ model classes, each represented by an image $\rho_j$ (training images, or templates), with $j = 1, \dots, M$. 

\item Both training and testing data are affected by nuisance variability due to changes of (i) illumination (ii) vantage point and (iii) partial occlusion. The former is approximated by local contrast transformations (monotonic continuous changes of intensity values), a maximal invariant to which is the gradient orientation. Vantage point changes are decomposed as a translation parallel to the image plane, approximated by a planar translation of the image, and a translation orthogonal to it, approximated by a scaling of the image. Partial occlusions determine the shape of corresponding regions in training and test images, which are approximated by a given shape (say a circle, or square) of unknown size (scale). These are very crude approximations but nevertheless implicit to most local descriptors. In particular, camera rotations are not addressed in this work, although others have done so \cite{dongKDHBS15}. 

\item Solving the (local) correspondence problem amounts to an $M+1$-hypothesis testing problem, including the background class. Nuisance (i) is eliminated at the outset by considering gradient orientation instead of image intensity. Dealing with nuisances (ii)--(iii) requires searching across all (continuous) translations, scales, and domain sizes. 

\item The resulting matching function must be discretized for implementation purposes. Since the matching cost is quadratic in the number of samples, sampling should be reduced to a minimum, which in general introduces artifacts (``aliasing'').

\item Anti-aliasing operators can be used to reduce the effects of aliasing artifacts. For the case of (approximations of) the likelihood function, such as SIFT, anti-aliasing corresponds to marginalizing residual nuisance transformations, which in turn corresponds to pooling gradient orientations across different locations, scales and domain sizes. 
\item The samples can be thought of as a special case of ``deformation hypercolumns'' \cite{soatto10} (samples with respect to the orientation group) with the addition of the size-space semi-group (Fig. \ref{fig-scale-size}). Most importantly, the samples along the group are anti-aliased, to reduce the effects of structural perturbations. 
\end{itemize}

\subsection{Formalization} 
\label{sect-formalization}

For simplicity, we formalize the matching problem for a scalar image (a scanline), and neglect contrast changes for now, focusing on the location-scale group and domain size instead.

Let $\rho_j: \real \rightarrow \real$, with $j = 1, \dots, M$ possible models (templates, or ideal training images). The data (test image) is $f: [0, \dots, N] \rightarrow \real$ with each sample $f(x_i)$ obtained from one of the $\rho_j$ via translation by $\tau \in \real$, scaling by $\sigma > 0$, and sampling with interval $\epsilon$, {\em if $x_i$ is in the visible domain} $[a, \ b]$. Otherwise, the scene $\rho_j$ is occluded and $f(x_i)$ has nothing to do with it.

The forward model that, given $\rho$ and all nuisance factors $\sigma, \tau, a, b$, generates the data, is indicated as follows: If $x_i \in [a, \ b]$, then 
\be
f(x_i) = W_\epsilon(x_i; \sigma, \tau)\rho_j + n_{ij}
\ee
where $n_i$ is a sample of a white, zero-mean Gaussian random variable with variance $\kappa$. Otherwise, $x_i \notin [a, \ b]$, and $f(x_i) = \beta(x_i)$ is a realization of a process independent of $\rho_j$ (the ``background''). The operator $W_\epsilon$ is linear\footnote{$W: {\mathbb L}^2(\real) \ \rightarrow \real^N$ can be written as an integral on the real line using the characteristic function $\chi_{{}_{{\cal B}_\epsilon}}(x-x_i)$ or a more general sampling kernel $k_\epsilon(x-x_i)$, for instance a Gaussian with zero-mean and standard deviation $\epsilon$. Then we have 
\begin{multline}
\int_{{\cal B}_\epsilon(x_i)} \rho\left(\frac{x-\tau}{\sigma}\right) dx  = 
 \int k_\epsilon(x-x_i) \rho\left(\frac{x-\tau}{\sigma}\right) dx = \iint
 \delta\left(y - \frac{x-\tau}{\sigma}\right) k_\epsilon(x-x_i) \rho\left(y \right) dx dy= \\
\iint  \delta\left(y + \frac{\tau}{\sigma} - \frac{x}{\sigma} \right)  k_\epsilon(x-x_i) dx \rho\left(y \right) dy = \iint  \delta\left(y + \frac{\tau}{\sigma} - \bar x \right)  k_\epsilon(\sigma \bar x-x_i) \sigma  d\bar x \rho\left(y \right) dy = \sigma \int  k_\epsilon(\sigma y + \tau - x_i) \rho\left(y \right) dy
\end{multline}
} and given by 
\be
W_\epsilon(x_i; \sigma, \tau)\rho \doteq \int_{{\cal B}_\epsilon(x_i)} \rho\left(\frac{x-\tau}{\sigma}\right) dx 
\ee
where ${\cal B}_\epsilon(x_i)$ is a region corresponding to a pixel centered at $x_i$. Matching then amount to a hypothesis testing problem on 
 whether a given measured $f = \{f(x_i)\}_{i = 1}^N$ is generated by any of the $\rho_j$ -- under suitable choice of nuisance parameters -- or otherwise is just labeled as background:
\be
H_0: \ \exists \ j, a, b, \sigma, \tau \ | \ p(f(x_i) | \rho_j, a, b, \sigma, \tau) = p_{\beta}(\{f(x_k), \ x_k \notin [a, \ b]\}) \prod_{x_i \in [a, \ b]} {\cal N}(f(x_i) - W_\epsilon(x_i; \sigma, \tau)\rho_j) , \kappa)
\ee
and the alternate hypothesis is simply $p_\beta(\{f(x_i)\}_{i = 1}^N)$. If the background density $p_\beta$ is unknown, the likelihood ratio test reduces to the comparison of the product on the right-hand side to a threshold, typically tuned to the ratio with the second-best match (although some recent work using extreme-value theory improves this \cite{fragoso2013evsac}). In any case, the log-likelihood for points in the interval $ x_i \in [a, \ b]$ can be written as
\be
r_{ij}(a, b, \sigma, \tau) = \frac{1}{|b-a|}\sum_{x_i \in [a, b]} | f(x_i) - W_\epsilon(x_i; \sigma, \tau)\rho_j|
\ee
which will have to be minimized for all pixels $i = 1, \dots, N$ and templates $j = 1, \dots, M$, of which there is a finite number. However, it also has to be minimized over the continuous variables $a, b, \sigma, \tau$. Since $r$ is in general neither convex nor smooth as a function of these parameters, analytical solutions are not possible. Discretizing these variables is necessary,\footnote{Coarse-to-fine, homotopy-based methods or jump-diffusion processes can alleviate, but not remove, this burden.} and since the minimization amounts to a search in $2+4$ dimensions, we seek for methods to {\em reduce the number of samples with respect to the arguments $a, b, \sigma, \tau$} as much as possible. 

There are many ways to sample, some described in Sect. \ref{sect-sampling}, so several questions are in order: (a) How should each variable be sampled? Regularly or adaptively? (b) If sampled regularly, when do aliasing phenomena occur? Can anti-aliasing be performed to reduce their effects? (c) The search is jointly over $a, b$ and $\sigma, \tau$, and given one pair, it is easy to optimize over the other. Can these two be ``separated''? 
(d) Is it possible to quantify and optimize the tradeoff between the number of samples and classification performance? Or for a given number of samples develop the ``best'' anti-aliasing (``descriptor'')? (e) For a histogram descriptor, how is ``anti-aliasing'' accomplished? 

\subsection{Common approaches and their rationale} 
\label{sect-common}

Concerning question (a) above, most approaches in the literature perform {\em tailored sampling} (Sect. \ref{sect-tailored}) of both $\tau$ and $\sigma$, by deploying a location-scale covariant detector \cite{lowe04distinctive}. When time is not a factor, it is common to forgo the detector and compute descriptors ``densely'' (a misnomer) by regularly subsampling the image lattice, or possibly undersampling by a fixed ``stride.'' Sometimes, scale is also regularly sampled, typically at far coarser granularity than the scale-space used for scale selection, for obvious computational reasons. In general, regular sampling requires assumptions on band limits. The function $W\rho$ is {\em not} band-limited as a function of $\tau$. Therefore, {\em tailored sampling} (detector/descriptor) is best suited for the translation group.\footnote{Purported superiority of ``dense SIFT'' (regularly sampled at thousands of location) compared to ordinary SIFT (at tens or hundreds of detected location), as reported in few empirical studies, is misleading as comparison has to be performed for a comparable number of samples.} We will therefore assume that $\tau$ has been tailor-sampled (detected, or canonized), but only up to a localization error. Without loss of generality we assume the sample is centered at zero, and the residual translation $\tau$ is in the neighborhood of the origin. 
In Fig. \ref{fig-specificity} we show that the sensitivity to scale of a common detector (DoG), which should be high, and is instead lower than the sensitivity of the resulting descriptor, which should be low. 
Therefore, small changes in scale cause large changes in scale sample localization, which in turn cause large changes in the value of the descriptor. Therefore, we forgo scale selection, and instead finely sample scale. This causes complexity issues, which prompt the need to sub-sample, and correspondingly to anti-alias or aggregate across scale samples. Alternatively, as done in Sect. \ref{sect-expm}, we can have a coarse adaptive or tailored sampling of scales, and then perform fine-scale sampling and anti-aliasing around the (multiple) selected scales. 

Concerning (b), anti-aliasing phenomena appear as soon as Nyquist's conditions are violated, which is almost always the case for scale and domain-size (Fig. \ref{fig-aliasing}). While most practitioners are reluctant to down-sample spatially, leaving millions of locations to test, it is rare for anyone to employ more than a few tens of scales, corresponding to a wild down-sampling of scale-space. This is true {\em a fortiori} for domain-size, where the domain size is often fixed, say to $69\times 69$ or $91\times 91$ locations \cite{fischer2014descriptor}. And yet, spatial anti-aliasing is routinely performed in most descriptors, whereas none -- to the best of our knowledge -- perform scale or domain-size anti-aliasing. Anti-aliasing should ideally decrease the sensitivity of the descriptor, without excessive loss of discriminative power. This is illustrated in Fig. \ref{fig-aliasing}.
  
For (c), we make the choice of fixing the domain size in the target (test) image, and regularly sampling scale and domain-size, re-mapping each to the domain size of the target (Fig. \ref{fig-visualize}). For comparison with \cite{fischer2014descriptor}, we choose this to be $69\times 69$. While the choice of fixing one of the two domains entails a loss, it can be justified as follows: Clearly, the hypothesis cannot be tested independently on each datum $f(x_i)$. However, testing on any {\em subset} of the ``true inlier set'' $[a, \ b]$ reduces the {\em power}, but not the validity, of the test. Vice-versa, using a ``superset'' that includes outliers invalidates the test. However, a small percentage of outliers can be managed by considering a robust (Huber) norm $\|f - W \rho\|_{\mathcal H}$ instead of the ${\mathbb L}^2$ norm. Therefore, one could consider the sequential hypothesis testing problem, starting from each $x_i \in [a = b]$ as an hypothesis, then ``growing'' the region by one sample, and repeating the test. Note that the optimization has to be solved at each step.\footnote{In this interpretation, the test can be thought of as a setpoint change detection problem. Another interpretation is that of (binary) region-based segmentation, where one wishes to classify the {\em range} of a function $f - W \rho$ into two classes, with values coming from either $\rho$ or the background, but the thresholds is placed on the {\em domain} of the function $[a, \ b]$. Of course, the statistics used for the classification depend on $a, b$ so this has to be solved as an alternating minimization, but it is a convex one \cite{chanE05}.} As a first-order approximation, one can {\em fix} the interval $[a, \ b]$ and accept a less powerful test (if that is a subset of the actual domain) or a test corrupted by outliers (if it is a superset). This is, in fact, done in most local feature-based registration or correspondence methods, and even in region-based segmentation of textures, where statistics must be pooled in a region. 

While (d) is largely an open question, (e) follows directly from classical sampling considerations, as described in Sect. \ref{sect-sampling}.

\subsection{Anti-aliasing descriptors} 

In the case of matching images under nuisance variability, it has been shown \cite{dongKDHBS15} that the {\em ideal descriptor} computed at a location $x_i$ is not a vector, but a function that approximates the likelihood, where the nuisances are marginalized. In practice the descriptor is approximated with a regularized histogram, similar to SIFT \eqref{eq-sift}. In this case, anti-aliasing corresponds to a weighted average across different locations, scales and {\em domain sizes}. But the averaging in this case is simply accomplished by pooling the histogram across different locations {\em and} domain-sizes, as in \eqref{eq-new-sift}. The weight function can be design to optimize the tradeoff between sensitivity and discrimination, although in Sect. \ref{sect-expm} we use a simple uniform weight. 

To see how pooling can be interpreted as a form of generalized anti-aliasing, consider the function $f$ sampled on a discretized domain $f(x_i)$ and a neighborhood ${\cal B}_\sigma(x_i)$ (for instance the sampling interval). The pooled histogram is 
\be
p_{x_i}(y) = \frac{1}{\sigma} \sum_{x_j \in {\cal B}_\sigma(x_i)} \delta (y - f(x_j))
\ee
whereas the anti-aliased signal (for instance with respect to the pillbox kernel) is
\be
\phi(x_i) = \frac{1}{\sigma} \sum_{x_j \in {\cal B}_\sigma(x_i)} f(x_j)
\ee
The latter can be obtained as the mean of the former
\be
\phi(x_i) = \sum_y y p_{x_i}(y)
\ee
although former can be used for purposes other than computing the mean (which is the best estimate under Gaussian ($\ell^2$) uncertainty), for instance to compute the median (corresponding to the best estimate under uncertainty measured by the $\ell^1$ norm), or the mode: 
\be
\hat f(x_i) = \arg\max_y p_{x_i}(y).
\ee
The approximation is accurate only to the extent in which the underlying distribution $p_x(y) = p(f(x)=y)$ is stationary and ergodic (so the spatially pooled histogram approaches the density), but otherwise it is still a generalization of the weighted average or mean. 

This derivation also points the way to how a descriptor can be used to synthesize images: Simply by sampling the descriptor, thought of as a density for a given class \cite{dongKDHBS15,vondrick2013hog}. It also suggests how descriptors can be compared: Rather than computing descriptors in both training and test images, a test datum can just be fed to the descriptor, to yield the likelihood of a given model class  \cite{felzenswalb}, without computing the descriptor in the test image.

\cut{In reality, the ideal descriptor as described in \cite{dongKDHBS15} is the joint likelihood, whereas DSP-SIFT (as well as all other variants of SIFT) represent only a product of marginals: 
\be
p(f | \rho_j) = \int \prod_{x_i \in [a, \ b]} {\cal N}(f(x_i) - W_\epsilon(x_i; \sigma, \tau)\rho_j ; \kappa) dP(\sigma) dP(\tau)
\ee
by moving the product outside the integral. This clearly entails discriminative loss, but at the advantage of enabling pooling independently in each ``cell.'' 
}

\section{Effect of the detector on the descriptor}

A {\em detector} is a function of the data that returns an element of a chosen group of transformations, the most common being translation (\eg~FAST), translation-scale (\eg~SIFT), similarity (\eg~SIFT combined the the direction of maximum gradient), affine (\eg~Harris-affine). Once transformed by the (inverse of) the detected transformation, the data is, by construction, invariant to the chosen group. If that was the only nuisance affecting the data, there would be no need for a descriptor, in the sense that the data itself, in the reference frame determined by any co-variant detector, is a maximal invariant to the nuisance group.

However, often the chosen group only captures a small subset of the transformations undergone by the data. For instance, all the groups above are only coarse approximations of the deformations undergone by the domain of an image under a change of viewpoint \cite{sundaramoorthiPVS09}. Furthermore, there are transformations affecting the range of the data (image intensity) that are not captured by (most) co-variant detectors. The purpose of the {\em descriptor} is to reduce variability to transformations that are not captured by the detector, while retaining as much as possible of the discriminative power of the data.

In theory, so long as descriptors are compared using the same detector, the particular choice of detector should not affect the comparison. In practice, there are many second-order effect where quantization and unmodeled phenomena affect different descriptors in different manners. Moreover, the choice of detector could affect different descriptors in different ways. The important aspect of the detector, however, is to determine a co-variant reference frame where the descriptor should be computed.

In standard SIFT, image gradient orientations are aggregated in selected regions of scale-space. Each region is defined in the octave corresponding to the selected scale, centered at the selected pixel location, where the selection is determined by the SIFT detector. Although the size of the original image subtended by each region varies depending on the selected scale (from few to few hundred pixels), the histogram is aggregated in regions that have constant size across octaves (the sizes are slightly different within each octave to subtend a constant region of the image). These are design parameters. For instance, in VLFeat they are assigned by default to $30\times 30, \ 38\times 38, \ 48\times 48$. In a different scale-space implementation, one could have a single design parameter, which we can call ``base size'' $\sigma_0$ for simplicity. 

In comparing with a convolutional neural network, Fisher et al. \cite{fischer2014descriptor} chose patches of size $64\times 64$ and $91\times 91$ (which we call $\sigma^*$) in \newcomment{images of maximum dimension $<1000$}. This choice is made for convenience in order to enable using pre-trained networks. They use MSER to detect candidate regions for testing, rather than SIFT's detector. However, rather than using the size of the original MSER to determine the octave where SIFT should be computed, they pre-process all patches to size $\sigma^*$. As a result, all SIFT descriptors are computed at the same octave $\sigma^*/\sigma_0*1.6 = 4.8$, rather than at the scale determined by the detector. This short-changes SIFT, as some descriptors are computed in regions that are too small relative to their scale, and others too large. 

\section{Choice of domain for comparison with CNNs}
\label{sect-ds}

One way to correct this bias would be to use $\sigma^*$ as the base size. However, this would yield an even worse (dataset-induced) bias: A base size of $91\times91$ in \newcomment{images of maximum dimension $<1000$} means that any feature detected at higher octaves encompasses the entire image. While discriminative power increases with size, so does the probability of straddling an occlusion: The power of a local descriptor increases with size only up to a point, where occlusion phenomena become dominant \comment{(Fig.~\ref{fig-performance-vs-base-size}).} This phenomenon is evident even in \newcomment{Oxford} and Fisher's datasets despite them being free of any occlusion phenomena.  Note that while Fisher et al. allow regions of size smaller than $91\times 91$ to be detected (and scale them up to that size), in effect anything smaller than $91\times 91$ is considered at the native resolution, whereas using the SIFT detector would send anything larger than $\sigma_0$ to a higher octave.

A more sound way to correct the bias is to use the detector in its proper role, that is to determine a reference frame with respect to which the descriptor is computed. For the case of MSER, this consists of affine transformations. Therefore, the region where the descriptor is computed is centered, oriented, skewed and {\em scaled} depending on the area of the region detected. Rather than arbitrarily fixing the scale by choosing a size to which to re-scale all patches, regions of different size are selected, and then each assigned a scale which is equal its area divided by the base size $\sigma_0$. That would determine the octave where SIFT is computed. 

In any case, regardless of what detector is used, DSP-SIFT is all about {\em where} to compute the descriptor: Instead of being computed just at the selected size, however it is chosen, it should be computed for multiple domain sizes. But scales have to be selected and mapped to the corresponding location in scale-space. There, SIFT aggregates gradient orientation at a single scale, whereas DISP-SIFT aggregates at multiple scales. 

\begin{figure}[h]
\centering
\includegraphics[width=.5\textwidth]{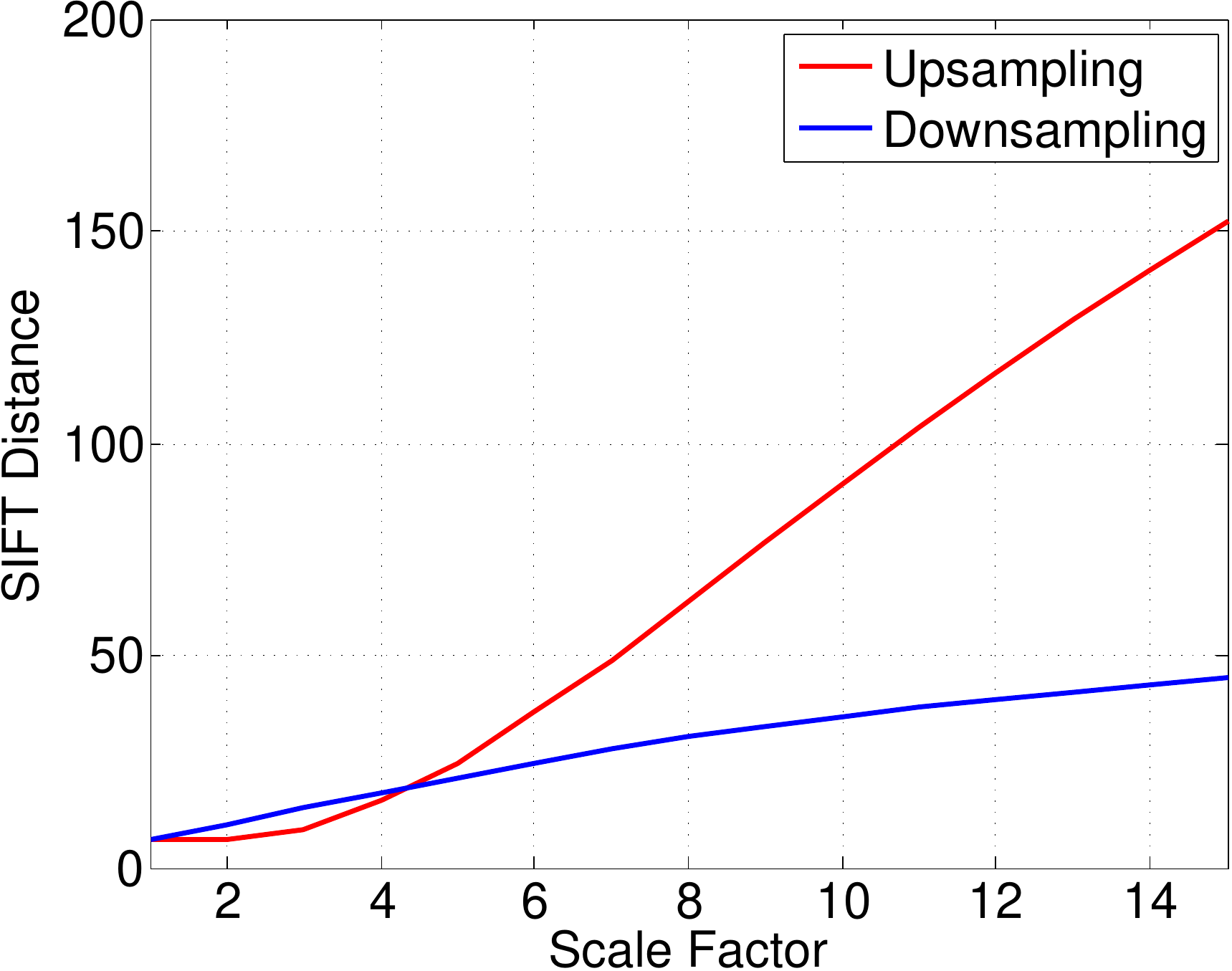}
\caption{{\sl Unidirectionality of mapping over scale.} Given two matching patches, one at high resolution, one at low resolution, comparison can be performed by mapping the high-resolution image to low-resolution by downsampling, or vice-versa mapping the low-resolution to high-resolution by upsampling and interpolation. Scale-space theory suggests that comparison should be performed at the lower resolution, since structures present at the high resolution cannot be re-created by upsampling and interpolation. The figure shows matching distance for matching high-to-low, and low-to-high (average for $2969$ random image patches in the Oxford dataset). This is why one should not choose a base region that is too large: That would cause all smaller regions to be upsampled and interpolated, to the detriment of matching scores. Note that computing descriptors at the native resolution, instead of the corresponding octave in scale-space, is equivalent to choosing a larger base region.
}
\end{figure}

\begin{figure}[h]
\centering
\includegraphics[width=.5\textwidth]{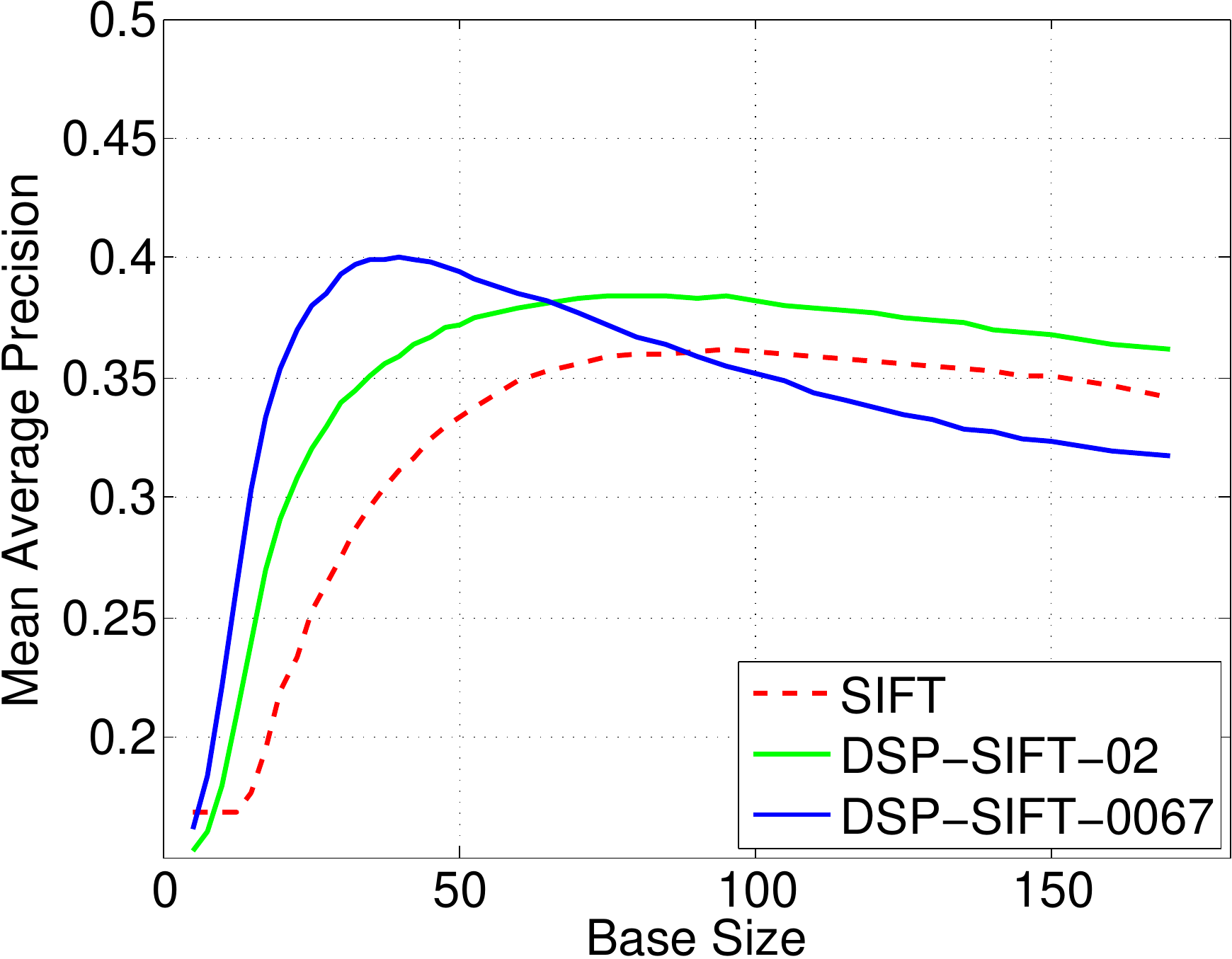}
\caption{{\sl Performance for varying choice of base size.} The base size determines the direction in which comparison over scale is performed: Larger regions are mapped down-scale, correctly. Smaller regions are mapped up-scale, to the detriment of the matching score. In theory, the larger the base size the better, up to the point where it impinges on occlusion phenomena. This explains the diminishing return behavior shown above. \newcomment{Different base size also affects what normalization threshold should be. We observe that a smaller threshold gives better performance with the most widely used base size ($\sim 30\times30$) default in VLFeat \cite{vlfeat}. }
}
\end{figure}

\end{document}